\documentclass[sigconf,natbib=true]{acmart}
\usepackage{enumitem} %
\usepackage{multirow}
\usepackage{booktabs}
\usepackage{breqn}
\usepackage{subcaption} %
\usepackage{graphicx}
\usepackage{amsmath}
\usepackage[most]{tcolorbox} %
\usepackage{float} %
\usepackage{listings} %
\usepackage{xcolor} %

\newtcolorbox{promptbox}[1][]{
    enhanced,
    title=#1,
    colback=white,
    colframe=black!50,
    boxrule=0.5pt,
    sharp corners,
    fonttitle=\bfseries,
    coltitle=black,
    attach boxed title to top left={yshift=-2mm, xshift=2mm},
    boxed title style={sharp corners, colback=white}
}

\definecolor{keycolor}{RGB}{19,110,194}      %
\definecolor{valuecolor}{RGB}{148,0,211}     %
\definecolor{numbcolor}{RGB}{128,64,0}       %
\definecolor{bracketcolor}{RGB}{102,102,102} %

\lstdefinelanguage{json}{ %
    basicstyle=\ttfamily\small,
    showstringspaces=false,
    breaklines=true,
    frame=single,
    morestring=[b]",
    stringstyle=\color{valuecolor},
    morecomment=[l]{:},
    commentstyle=\color{keycolor}\bfseries,
    literate=
    {:}{{{\color{keycolor}{:}}}}{1}
    {,}{{{\color{bracketcolor}{,}}}}{1}
    {\{}{{{\color{bracketcolor}{\{}}}}{1}
    {\}}{{{\color{bracketcolor}{\}}}}}{1}
    {[}{{{\color{bracketcolor}{[}}}}{1}
    {]}{{{\color{bracketcolor}{]}}}}{1}
    {0}{{{\color{numbcolor}{0}}}}{1}
    {1}{{{\color{numbcolor}{1}}}}{1}
    {2}{{{\color{numbcolor}{2}}}}{1}
    {3}{{{\color{numbcolor}{3}}}}{1}
    {4}{{{\color{numbcolor}{4}}}}{1}
    {5}{{{\color{numbcolor}{5}}}}{1}
    {6}{{{\color{numbcolor}{6}}}}{1}
    {7}{{{\color{numbcolor}{7}}}}{1}
    {8}{{{\color{numbcolor}{8}}}}{1}
    {9}{{{\color{numbcolor}{9}}}}{1}
}

\usepackage[capitalise]{cleveref} %

\AtBeginDocument{%
  }

\setcopyright{acmlicensed}
\copyrightyear{2025}
\acmYear{2025}
\acmConference[LLM4Eval@SIGIR '25]{Proceedings of the Third Workshop on Large Language Models for Evaluation in Information Retrieval}{July 13-17, 2025}{Padua, Italy}
\acmBooktitle{Proceedings of the Third Workshop on Large Language Models for Evaluation in Information Retrieval (LLM4Eval@SIGIR '25), July 13-17, 2025, Padua, Italy}

\begin{document}

\title{CCRS: A Zero-Shot LLM-as-a-Judge Framework for Comprehensive RAG Evaluation}

\author{Aashiq Muhamed}
\affiliation{%
  \institution{Carnegie Mellon University}
  \city{Pittsburgh, Pennsylvania}
  \country{USA}}
\email{amuhamed@andrew.cmu.edu}
\renewcommand{\shortauthors}{Aashiq Muhamed} %

\begin{abstract}
Retrieval-Augmented Generation (RAG) systems enhance Large Language Models (LLMs) by incorporating external knowledge, which is crucial for domains that demand factual accuracy and up-to-date information. However, evaluating the multifaceted quality of RAG outputs — spanning aspects such as contextual coherence, query relevance, factual correctness, and informational completeness—poses significant challenges. Existing evaluation methods often rely on simple lexical overlap metrics, which are inadequate for capturing these nuances, or involve complex multi-stage pipelines with intermediate steps like claim extraction or require finetuning specialized judge models, hindering practical efficiency. To address these limitations, we propose CCRS (Contextual Coherence and Relevance Score), a novel suite of five metrics that utilizes a single, powerful, pretrained LLM (Llama 70B) as a zero-shot, end-to-end judge. CCRS evaluates: Contextual Coherence (CC), Question Relevance (QR), Information Density (ID), Answer Correctness (AC), and Information Recall (IR). We apply CCRS to evaluate six diverse RAG system configurations (varying retrievers and readers) on the challenging BioASQ biomedical question-answering dataset. Our analysis demonstrates that CCRS effectively discriminates between system performances, confirming, for instance, that the Mistral-7B reader outperforms Llama variants and that the E5 neural retriever enhances QR and IR for Llama models in this task. We provide a detailed analysis of CCRS metric properties, including score distributions, convergent/discriminant validity, tie rates, population statistics, and discriminative power, finding QR highly discriminative overall, while observing a strong AC-IR correlation. Compared to the complex RAGChecker framework, CCRS offers comparable or superior discriminative power for key aspects like recall and faithfulness, while being significantly more computationally efficient. CCRS thus provides a practical, comprehensive, and efficient framework for evaluating and iteratively improving RAG systems.
\end{abstract}



\maketitle

\section{Introduction}
\label{sec:introduction}
RAG systems \citep{gao2023retrieval, asai2024reliable} represent a significant advancement in Large Language Model (LLM) technology by integrating external knowledge retrieval with generative capabilities. This synergy aims to produce responses that are more factual, current, and contextually appropriate than those generated by LLMs relying solely on their internal parametric knowledge. The ability to ground generated content in external evidence is particularly vital in specialized domains such as medicine, finance, and law, where precision, reliability, and grounding in specific, often dynamic, knowledge bases are paramount \citep{ziems2024can}. However, the effectiveness of RAG systems hinges on a complex interplay between the quality of the retrieved information (relevance, accuracy, completeness) and the LLM's ability to comprehend, synthesize, and faithfully generate responses based on that information. This inherent complexity poses significant evaluation challenges.

Failures in RAG systems can manifest in various ways, hindering their trustworthiness and utility. These failures include the retrieval of irrelevant or contradictory documents, the LLM's inability to discern correct information amidst noisy or conflicting context, unresolved conflicts between retrieved information and the LLM's internal parametric knowledge, or the generation of fluent text that nonetheless misrepresents, omits crucial details from, or hallucinates information beyond the provided context \citep{jin2024tug, kortukov2024studying, ye2023large}. Consequently, evaluating RAG systems requires a multifaceted approach that moves beyond assessing simple output quality. It must consider multiple critical dimensions, including faithfulness to the provided context, relevance to the user's query, factual accuracy against ground truth, completeness of information, logical coherence, and conciseness \citep{es2023ragas, saadfalcon2023ares}.

Traditional NLP metrics, such as BLEU \citep{papineni2002bleu}, ROUGE \citep{lin2004rouge}, and even more advanced semantic similarity measures like BERTScore \citep{zhang2019bertscore}, primarily rely on comparing the generated output to one or more static reference texts. These metrics are often insufficient for RAG evaluation because they fail to adequately capture critical aspects like whether the generated text is factually supported by the \emph{retrieved} context—a core requirement for trustworthy RAG systems \citep{es2023ragas, saadfalcon2023ares}. They cannot reliably penalize plausible-sounding but unfaithful or factually inaccurate generation, nor can they effectively assess the quality of the retrieval process itself.

The advent of increasingly powerful LLMs has spurred interest in using them as evaluators, often referred to as the \emph{LLM-as-a-judge} paradigm \citep{zheng2023judging, chiang-lee-2023-large}. This approach offers a potential solution for assessing the nuanced quality dimensions required for RAG evaluation. Several RAG-specific evaluation frameworks leveraging this paradigm have been proposed. RAGAS \citep{es2023ragas} introduces metrics like faithfulness and answer relevance, but its faithfulness calculation involves a multi-step pipeline: first decomposing the generated answer into individual statements and then performing Natural Language Inference (NLI) checks for each statement against the retrieved context. ARES \citep{saadfalcon2023ares} utilizes finetuned smaller LLMs as judges to predict scores for context relevance, answer faithfulness, and answer relevance. While potentially accurate due to supervised training, this approach demands considerable effort in generating large synthetic training datasets and obtaining human validation data for calibration, often using complex statistical techniques like Prediction-Powered Inference (PPI). CRUD-RAG \citep{lyu2024crud} focuses on completeness and accuracy by using an LLM to first generate questions from the ground truth and then assessing the RAG system's response based on its ability to answer these generated questions. RAGChecker \citep{ru2024ragchecker} offers highly fine-grained analysis by extracting claims from the response, ground truth, and context, followed by pairwise entailment checking. This allows for detailed metrics like precision, recall, faithfulness, hallucination rate, and noise sensitivity, providing comprehensive insights but at the cost of significant computational complexity and sensitivity to potential errors in the intermediate claim extraction and entailment steps.

While these frameworks represent significant progress, their reliance on complex multi-stage pipelines (RAGAS, RAGChecker, CRUD-RAG) or extensive setup requirements involving data generation and fine-tuning (ARES) can limit their practical applicability, especially for rapid iteration during development or for large-scale evaluations across many system variants. There remains a need for a robust, comprehensive, yet efficient RAG evaluation method.

To address this gap, we introduce CCRS (Contextual Coherence and Relevance Score), a novel suite of five evaluation metrics. CCRS employs a single, powerful, pretrained LLM (specifically, Llama 70B-Instruct \citep{touvron2023Llama}) as a zero-shot judge. It directly evaluates the end-to-end RAG output across five key dimensions without requiring intermediate processing steps (like claim extraction or question generation) or specialized fine-tuning. The CCRS metrics are designed to capture critical aspects of RAG quality:
\begin{enumerate}[noitemsep, topsep=2pt, partopsep=1pt, parsep=1pt, leftmargin=*]
    \item \textbf{Contextual Coherence (CC):} Assesses the logical consistency and flow of the response with respect to the provided context.
    \item \textbf{Question Relevance (QR):} Evaluates the appropriateness and directness of the response in addressing the user's query.
    \item \textbf{Information Density (ID):} Measures the balance between conciseness and informativeness in the response.
    \item \textbf{Answer Correctness (AC):} Determines the factual accuracy of the response compared to a ground truth answer, considering the context.
    \item \textbf{Information Recall (IR):} Assesses the completeness of the response in capturing essential information present in the ground truth answer.
\end{enumerate}
\looseness=-1
This framework aims to harness the nuanced understanding capabilities of large LLMs in an efficient, end-to-end manner, providing a practical tool for RAG evaluation.
Our main contributions are:
\begin{enumerate}
    \item We propose and define CCRS, a zero-shot, end-to-end LLM-as-a-judge framework with five distinct metrics designed for practical and comprehensive RAG evaluation (\autoref{sec:ccrs_framework}).
    \item We conduct a comprehensive empirical evaluation using the challenging BioASQ biomedical question-answering dataset, comparing six diverse RAG configurations (varying retrievers and readers) and rigorously testing hypotheses about the impact of system components (\autoref{sec:hypothesis_testing_results}).
    \item We perform a detailed analysis of the properties of the CCRS metrics, including their distributions, validity (convergent and discriminant), tie rates, population statistics, and discriminative power. We also compare their effectiveness and computational efficiency against the complex, state-of-the-art RAGChecker framework (\autoref{sec:metric_properties}).
\end{enumerate}
This work contributes a practical evaluation methodology and valuable empirical insights into RAG system performance. 

\section{Related Work and Framework}
\label{sec:related_work_and_framework}

\subsection{Limitations of Traditional Metrics}
\label{sec:related_work_traditional}

\looseness=-1
Evaluating the output quality of generative systems like RAG requires moving beyond metrics designed for tasks like machine translation or summarization, which primarily assess surface-level similarity to reference texts. Traditional metrics such as BLEU \citep{papineni2002bleu}, which measures n-gram precision against references (often calculated as $\text{BLEU} = \text{BP} \cdot \exp(\sum_{n=1}^N w_n \log p_n)$, where BP is a brevity penalty), and ROUGE \citep{lin2004rouge}, particularly ROUGE-L which uses the Longest Common Subsequence (LCS) ($F_L = \frac{(1+\beta^2)R_L P_L}{\beta^2 R_L + P_L}$), focus on lexical overlap. While useful for assessing fluency and lexical similarity, they fail to capture deeper semantic meaning or factual correctness.

Metrics like BERTScore \citep{zhang2019bertscore} represent an improvement by leveraging contextual embeddings 
($\sum_{x_i \in gen} \max_{y_j \in ref} \text{sim}(\mathbf{x}_i, \mathbf{y}_j)$) to compute semantic similarity between generated and reference texts. However, even these advanced semantic metrics have significant limitations in the context of RAG evaluation \citep{liu2023geval, es2023ragas}. They cannot directly evaluate the faithfulness of the generated response to the \emph{retrieved context}, nor can they assess factual accuracy independently of lexical or semantic similarity to a potentially incomplete or imperfect reference answer. They also do not penalize plausible but unfaithful hallucinations or factually incorrect statements that might still share some semantic similarity with the reference.

\subsection{LLM-based Evaluation Frameworks}
\label{sec:related_work_llm}
Recognizing the limitations of traditional metrics, recent research has increasingly focused on leveraging the sophisticated language understanding capabilities of LLMs themselves to perform evaluation, often termed \emph{LLM-as-a-judge} \citep{zheng2023judging, chiang-lee-2023-large}. Several frameworks specifically designed for RAG evaluation have emerged from this line of work:

\begin{itemize}[noitemsep, topsep=2pt, partopsep=1pt, parsep=1pt, leftmargin=*]
    \item \textbf{TruLens} \citep{ferrara2024rag}: Proposes a conceptual \emph{RAG Triad} consisting of groundedness (faithfulness to context), answer relevance (to the query), and context relevance (of retrieved documents to the query), which are assessed using LLM prompts. While conceptually aligned with key RAG quality dimensions, the framework remains somewhat underspecified in its published form and lacks extensive validation studies demonstrating its reliability and effectiveness compared to other methods.
    \item \textbf{RAGAS} \citep{es2023ragas}: \looseness=-1 Introduces several metrics, including \textit{faithfulness}, \textit{answer relevancy}, and \textit{context relevancy}. Its faithfulness calculation requires a multi-step pipeline: an LLM first extracts discrete claims from the generated response ($r$), and then another LLM (or the same one) performs Natural Language Inference (NLI) checks for each claim against the retrieved context ($C$). The final score is the proportion of verified claims ($\text{Faithfulness} = \frac{|\{s_i \in \text{claims}(r) \mid \text{verify}(s_i, C)\}|}{|\text{claims}(r)|}$). This reliance on intermediate claim extraction and verification steps can make the process computationally demanding and potentially brittle if the intermediate steps fail.
    \item \textbf{ARES} \citep{saadfalcon2023ares}: Employs finetuned smaller LLMs (e.g., DeBERTa variants) as specialized judges to predict scores for context relevance, answer faithfulness, and answer relevance. Its potential strength lies in achieving high accuracy through supervised training. However, this comes at the cost of significant upfront investment: creating large-scale synthetic datasets (often using powerful generator models like FLAN-T5 XXL) to train the judges, and obtaining human validation data for calibration using sophisticated statistical techniques like Prediction-Powered Inference (PPI). This setup complexity can be a barrier to adoption.
    \item \textbf{CRUD-RAG} \citep{lyu2024crud}: \looseness=-1 Focuses specifically on evaluating the completeness and accuracy of the RAG response ($r$) relative to a ground truth answer ($g$). It uses an LLM to first generate a set of questions $Q(g)$ based on the ground truth. It then assesses the RAG response by checking if it can answer these generated questions ($\text{Recall} = |\{q \in Q(g) \mid A_r(q) \neq \emptyset\}| / |Q(g)|$) and measures the accuracy of the answers extracted from $r$ compared to those derived from $g$ ($\text{Precision} \approx \text{avg}(Sim(A_r(q), A_g(q)))$). This approach introduces dependencies on the quality and coverage of the intermediate question generation and answer extraction steps.
    \item \textbf{RAGChecker} \citep{ru2024ragchecker}: Provides arguably the most fine-grained analysis through claim-level checking. It involves extracting claims from the model's response ($m$), the ground truth ($gt$), and the retrieved context chunks ($\{chunk_j\}$). It then performs pairwise entailment checks between these sets of claims. This enables the computation of detailed metrics such as Precision ($\frac{|\{c^{(m)}_i \mid c^{(m)}_i \in gt\}|}{|\{c^{(m)}_i\}|}$), Recall ($\frac{|\{c^{(gt)}_i \mid c^{(gt)}_i \in m \}|}{|\{c^{(gt)}_i\}|}$), Faithfulness \\($\frac{|\{c^{(m)}_i \mid c^{(m)}_i \in \{chunk_j\}\}|}{|\{c^{(m)}_i\}|}$), Hallucination Rate, Noise Sensitivity, and Context Utilization. While extremely comprehensive, the multi-step process involving claim extraction and numerous entailment checks makes RAGChecker computationally intensive and potentially sensitive to errors propagating from the intermediate steps.
\end{itemize}

\subsection{The CCRS Framework}
\label{sec:ccrs_framework}
CCRS (Contextual Coherence and Relevance Score) is designed to provide a comprehensive yet efficient evaluation of RAG systems, balancing the depth of multi-dimensional assessment with the practicality of a streamlined workflow. Unlike the related works discussed, CCRS seeks an effective balance between the simplicity of direct LLM prompting (akin to aspects of TruLens or general reference-free LLM evaluation) and the multi-dimensional assessment capability offered by more complex frameworks like RAGAS, ARES, CRUD-RAG, and RAGChecker. The core hypothesis underlying CCRS is that a single, sufficiently powerful, pretrained LLM (in our case, Llama 70B-Instruct) can perform nuanced judgments across multiple quality dimensions in a zero-shot, end-to-end manner, directly evaluating the final RAG output without needing intermediate processing steps or specialized training.

By avoiding the pipeline complexity inherent in claim extraction/verification (RAGAS, RAGChecker) or question generation/answering (CRUD-RAG), and bypassing the extensive training data generation and calibration overhead required by supervised judge models (ARES), CCRS aims to offer a practical and computationally efficient solution suitable for rapid development cycles and large-scale comparative evaluations.

\subsubsection{Task Definition \& Constructs}
\looseness=-1
We focus on evaluating RAG systems designed for Question Answering (QA). The input to the RAG system is $x = (q, D)$, where $q$ is the user's question and $D$ is the document collection. The output is $y = (C, r)$, where $C \subseteq D$ is the set of retrieved context passages and $r$ is the generated response. The evaluation task is to assess the quality of the generated response $r$, given the input question $q$, the retrieved context $C$, and a ground truth answer $g$.

Our primary goal is to develop metrics that serve as effective proxies for \textbf{user satisfaction} with the RAG system's output. We operationalize this goal by measuring five key quality constructs, defined in \autoref{tab:constructs_condensed_framework_detail_main_v4}. These constructs were chosen to cover essential aspects determining whether a RAG response is helpful, reliable, and addresses the user's information need effectively.

\begin{table}[htbp]
\centering
\caption{CCRS Construct Definitions and Motivations}
\label{tab:constructs_condensed_framework_detail_main_v4}
\small \vspace{-0.5em}
\begin{tabular}{@{}lp{0.6\columnwidth}@{}}
\toprule
Construct & Definition \& Motivation \\
\midrule
Contextual Coherence (CC) & Logical consistency and flow between the response ($r$) and the retrieved context ($C$). (\textit{Motivation:} Ensures the response is well-supported by the evidence and internally consistent with it, crucial for groundedness.) \\
Question Relevance (QR) & Appropriateness and directness of the response ($r$) to the user's query ($q$). (\textit{Motivation:} Ensures the response actually addresses the user's specific information need, fundamental to usefulness.) \\
Information Density (ID) & Optimal balance between informativeness and conciseness in the response ($r$). (\textit{Motivation:} Avoids overly verbose answers that overwhelm the user or overly brief answers that lack necessary detail, impacting user experience.) \\
Answer Correctness (AC) & Factual accuracy of claims in the response ($r$) compared to the ground truth answer ($g$). (\textit{Motivation:} Critical for trust and reliability, especially in high-stakes domains like medicine or finance.) \\
Information Recall (IR) & Coverage of essential information from the ground truth answer ($g$) within the response ($r$). (\textit{Motivation:} Ensures the answer is sufficiently complete and provides necessary details, complementing correctness.) \\
\bottomrule
\end{tabular}
\vspace{-1em}
\end{table}

\subsubsection{CCRS Components and Metric Calculation}
\label{sec:ccrs_components}
CCRS utilizes \texttt{Llama-70B-Instruct} as the $LLM_{\text{Judge}}$ to perform zero-shot evaluations for each of the five constructs. Specific prompts, detailed in \autoref{app:prompts}, are used to guide the LLM judge. The judge is instructed to output a score between 0 and 100 for each dimension, which we then normalize to a [0, 1] range for consistency.

\textbf{Contextual Coherence (CC):} Calculated as  $CC(r, C) =  \\ LLM_{\text{Judge}}(r, C, \text{prompt}_{\text{CC}}) / 100$.
This metric directly assesses whether the generated text $r$ logically follows from and avoids contradicting the provided context $C$. A high CC score indicates that the response is well-grounded in the evidence and integrates information coherently. The corresponding prompt (\autoref{app:prompt_cc}) explicitly asks the judge to evaluate logical consistency and coherence.

\textbf{Question Relevance (QR):} Calculated as  $QR(r, q) = \\ LLM_{\text{Judge}}(r, q, \text{prompt}_{\text{QR}}) / 100$.
This metric directly evaluates if the response $r$ effectively answers the specific question $q$ posed by the user. High relevance is fundamental to the perceived usefulness of the response. The prompt (\autoref{app:prompt_qr}) asks the judge \emph{how well the response addresses the user's query}.

\textbf{Information Density (ID):} Calculated as $ID(r, C, q) = \\ LLM_{\text{Judge}}(r, C, q, \text{prompt}_{\text{ID}}) / 100$.
This assesses the communication efficiency of the response $r$. It's important for user experience, aiming to avoid information overload from excessive verbosity or frustration from underspecification. The prompt (\autoref{app:prompt_id}) asks the judge to assess the \emph{balance of conciseness and informativeness}.

\textbf{Answer Correctness (AC):} Calculated as $AC(r, g, C) = (\lambda \cdot \text{EM}(r, g) + (1 - \lambda) \cdot LLM_{\text{Judge}}(r, g, C, \text{prompt}_{\text{AC}}) / 100)$.
Measures the factual alignment of the response $r$ with the ground truth answer $g$, which is crucial for reliability. We combine a strict exact match (EM) check ($\text{EM}(r, g) = 1$ if $r=g$, else 0) with the LLM's judgment of semantic correctness, allowing for paraphrasing while considering the retrieved context $C$. We use a weight $\lambda=0.7$ to emphasize exact matches, reflecting the high precision often required in domains like biomedical QA, while still incorporating semantic evaluation. The prompt (\autoref{app:prompt_ac}) asks for \emph{factual accuracy... compared to the ground truth answer, considering the context}.

\textbf{Information Recall (IR):} Calculated as  $IR(r, g, C) = \\ LLM_{\text{Judge}}(r, g, C, \text{prompt}_{\text{IR}}) / 100$.
Measures the completeness of the response $r$ in covering the essential information present in the ground truth answer $g$. This complements AC by ensuring that key details required for a full answer are not omitted. The prompt (\autoref{app:prompt_ir}) asks \emph{how much of the essential information from the ground truth is captured}.

By combining these five metrics, CCRS provides a multi-faceted evaluation of RAG system output quality in an efficient, zero-shot manner.

\section{Experimental Setup}
\label{sec:setup}

We designed our experiments to rigorously evaluate the CCRS framework and compare the performance of different RAG system configurations on a challenging task.

\subsection{Dataset: BioASQ}
We utilized the BioASQ dataset \citep{tsatsaronis2015overview}, a widely recognized benchmark for biomedical question answering. We used a publicly available version containing N=4,719 expert-curated question-answer pairs along with associated PubMed passages\footnote{\url{https://huggingface.co/datasets/rag-datasets/rag-mini-bioasq}}. BioASQ questions are formulated by domain experts and cover a range of types (factoid, list, summary, yes/no) and complexities, requiring both accurate retrieval from biomedical literature and precise generation. Its domain specificity and complexity make it a suitable and challenging testbed for evaluating RAG systems and the metrics designed to assess them. Example data points can be found in \autoref{app:dataset_examples}.

\subsection{Systems Compared}
To assess CCRS's ability to differentiate between systems and to investigate the impact of RAG components, we evaluated six distinct RAG system configurations. These configurations were created by combining two different retrieval methods with three different reader LLMs:

\noindent \textbf{Retrievers:}
\begin{itemize}[noitemsep, topsep=0pt, partopsep=0pt, parsep=0pt, leftmargin=*]
    \item \textbf{BM25:} A traditional sparse retrieval algorithm based on lexical matching \citep{robertson2009probabilistic}, implemented using OpenSearch.
    \item \textbf{E5-Mistral:} A state-of-the-art neural dense retriever \citep{wang2023improving}, implemented using OpenSearch's approximate k-NN search capabilities.
\end{itemize}

\noindent  \textbf{Readers (Generators):}
\begin{itemize}[noitemsep, topsep=0pt, partopsep=0pt, parsep=0pt, leftmargin=*]
    \item \textbf{Mistral-7B:} Mistral-7B-Instruct-v0.3 \citep{jiang2023mistral}.
    \item \textbf{Llama3-8B:} Meta-Llama-3-8B-Instruct \citep{AI2024Llama}.
    \item \textbf{Llama3.2-3B:} Meta-Llama-3.2-3B-Instruct \citep{dubey2024Llama}.
\end{itemize}

\medskip
\noindent This resulted in the following six system configurations, labeled A through F:
\begin{itemize}[noitemsep, topsep=0pt, partopsep=0pt, parsep=0pt, leftmargin=*]
    \item \textbf{System A:} BM25 Retriever + Mistral-7B Reader
    \item \textbf{System B:} E5-Mistral Retriever + Mistral-7B Reader
    \item \textbf{System C:} BM25 Retriever + Llama3-8B Reader
    \item \textbf{System D:} E5-Mistral Retriever + Llama3-8B Reader
    \item \textbf{System E:} BM25 Retriever + Llama3.2-3B Reader
    \item \textbf{System F:} E5-Mistral Retriever + Llama3.2-3B Reader
\end{itemize}

\medskip \noindent
\textbf{CCRS Judge Model:} For all CCRS metric calculations, we used Meta-Llama-3-70B-Instruct as the $LLM_{\text{Judge}}$.

\subsection{Implementation Details}
All experiments were executed on institutional compute clusters, utilizing nodes equipped with 8x A100 80GB GPUs for parallel model inference.

\textbf{Document Processing and Retrieval:} Documents from the BioASQ corpus were chunked into segments of 300 tokens with an overlap of 20\% (60 tokens). We consistently used the tokenizer associated with the E5-Mistral model for chunking across both BM25 and E5 experiments to ensure comparability. For each query, the top-k=20 chunks as ranked by the respective retriever (BM25 or E5-Mistral) were retrieved and concatenated to form the context provided to the reader LLM.

\textbf{Response Generation:} The reader LLMs generated responses based on the input query and the retrieved context (k=20 chunks). To ensure deterministic and comparable outputs, we used a temperature setting of 0.0 for generation. The maximum length for generated responses was set to 2048 tokens. The prompt template used for instructing the reader models is shown in \autoref{fig:prompt_main_setup_impl_detail_v4}.

\begin{figure}[htbp]
\fbox{
    \begin{minipage}{\linewidth}
    \small
    Please answer the given question based on the context.
    \texttt{<context>}
    \texttt{<content>}{chunk\_1}\texttt{</content>}
    \texttt{<content>}{chunk\_2}\texttt{</content>}
    ...
    \texttt{<content>}{chunk\_k}\texttt{</content>}
    \texttt{</context>}

    Question: \texttt{\{question\}}

    Please answer the question and tag your answer with \texttt{<answer></answer>}.
    \end{minipage}
}
\vspace{-0.1in}
\caption{Default prompt template used for RAG response generation by the reader models (Mistral-7B, Llama3-8B, Llama3.2-3B).}
\label{fig:prompt_main_setup_impl_detail_v4}
\vspace{-1em}
\end{figure}

\textbf{Evaluation:} CCRS metrics were computed for each generated response using the \texttt{Llama-70B-Instruct} judge model. For the comparison with RAGChecker (\autoref{sec:ragchecker_comp}), we also used \texttt{Llama-70B-Instruct} for the claim extraction step to maintain consistency in the LLM used, although the original RAGChecker paper used GPT-4o. Statistical analyses were performed using custom Python scripts leveraging libraries like NumPy and SciPy, with the specific implementation detailed in \autoref{app:code}.

\section{Results and Analysis}
\label{sec:results}
In this section, we present the results of our experiments. We first analyze the properties of the CCRS metrics themselves, including their distributions, validity, tie rates, and discriminative power, along with a comparison to the RAGChecker framework. We then evaluate the performance of the six RAG systems based on the CCRS metrics and test our predefined hypotheses.

\subsection{CCRS Metric Properties}
\label{sec:metric_properties}

Understanding the behavior and characteristics of the CCRS metrics is crucial for interpreting the RAG system evaluation results.

\subsubsection{Score Distributions and Boundedness}
\label{sec:dist_bound}
We analyzed the distributions of scores for each CCRS metric across all 4,719 queries and 6 systems. The histograms of the mean scores per query (averaged over the 6 systems) are shown in \autoref{fig:mean_dist_results_detail_v4}. Detailed per-system histograms are available in Appendix Figures \ref{fig:cc_dist}-\ref{fig:ir_dist}.

\begin{figure*}[htbp!]
\centering
    \begin{subfigure}[b]{0.32\textwidth}
        \centering
        \includegraphics[width=\linewidth]{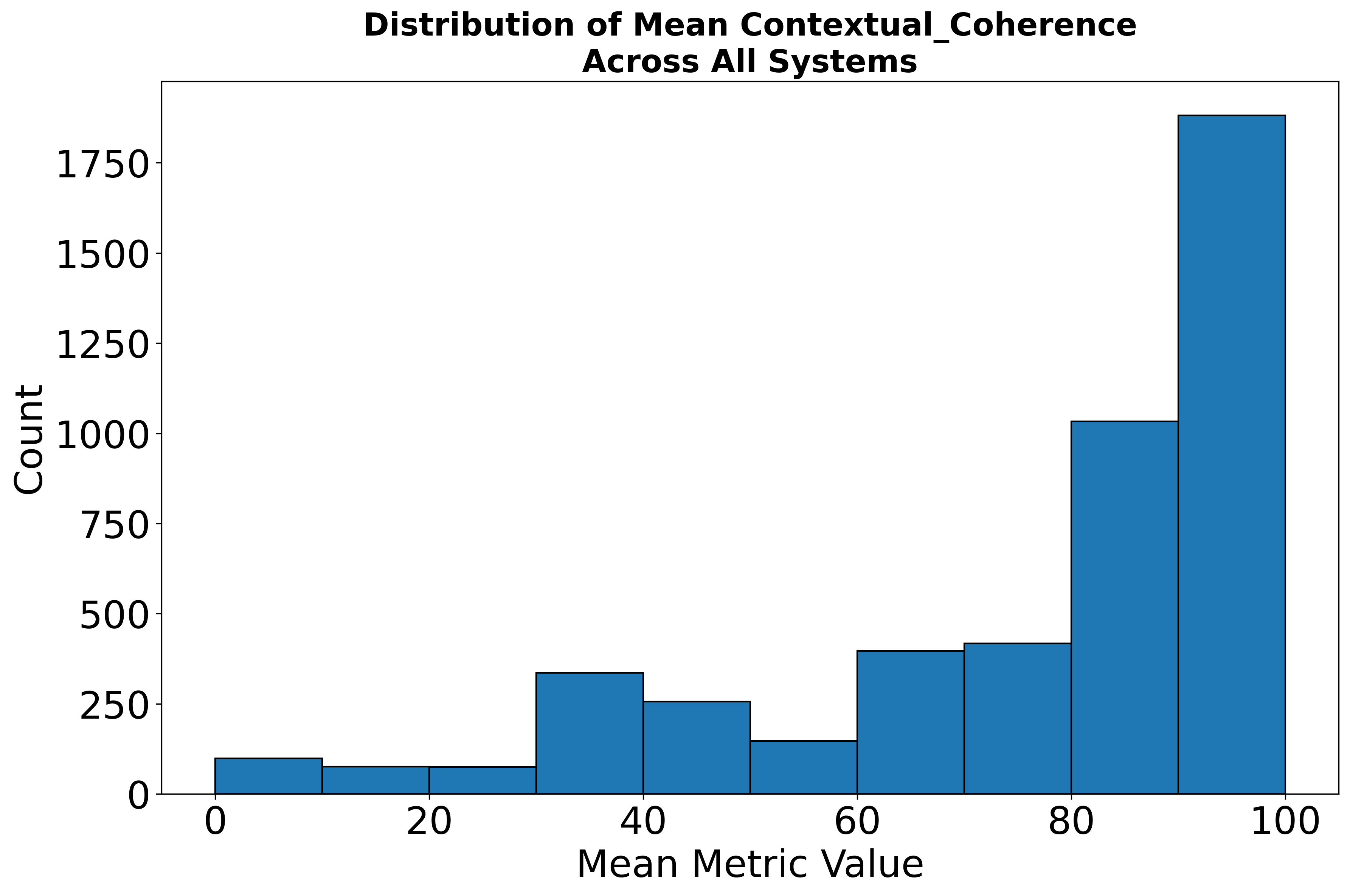}
        \caption{Contextual Coherence (CC)}
    \end{subfigure} \hfill
    \begin{subfigure}[b]{0.32\textwidth}
        \centering
        \includegraphics[width=\linewidth]{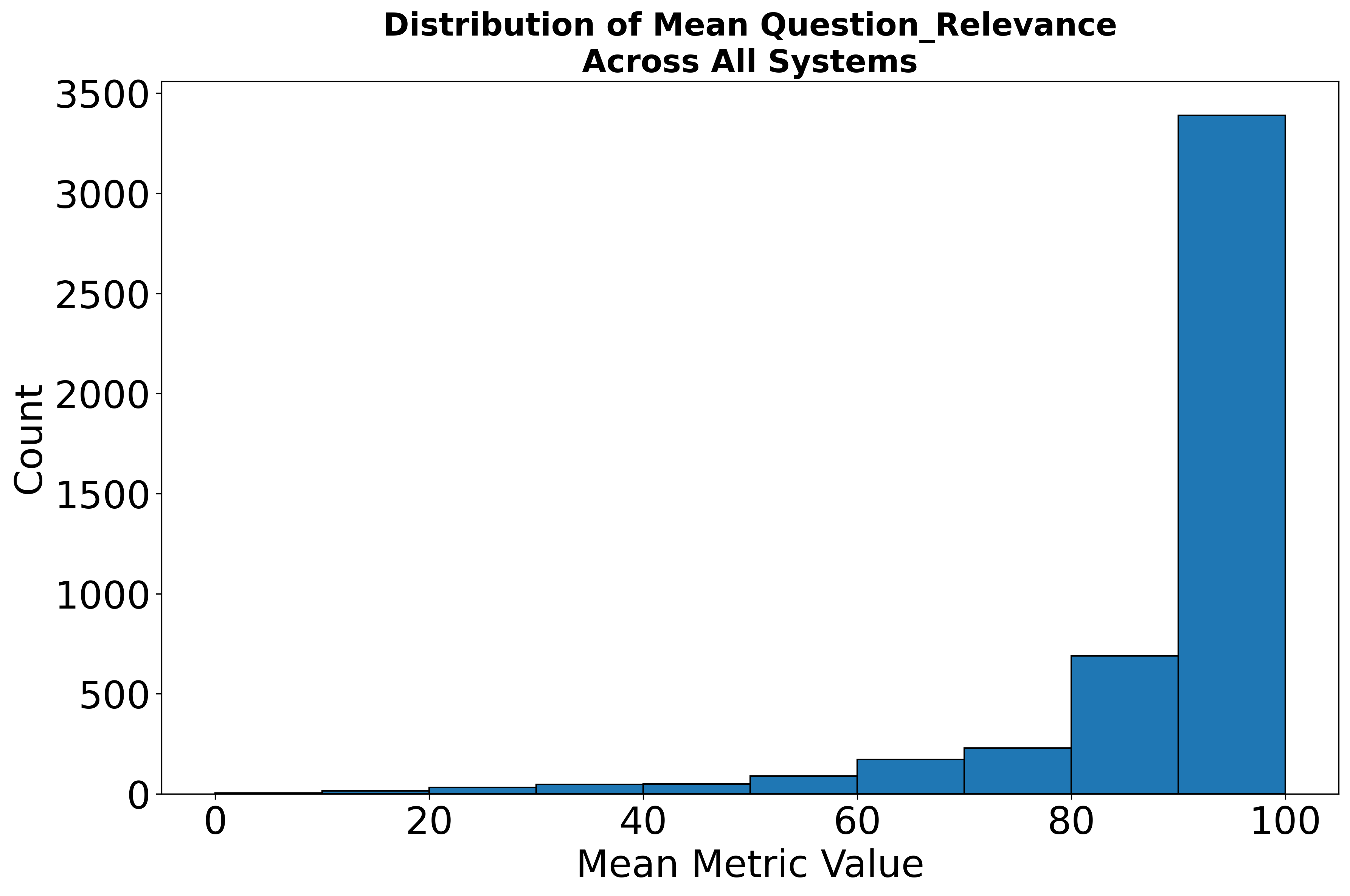}
        \caption{Question Relevance (QR)}
     \end{subfigure} \hfill
     \begin{subfigure}[b]{0.32\textwidth}
         \centering
         \includegraphics[width=\linewidth]{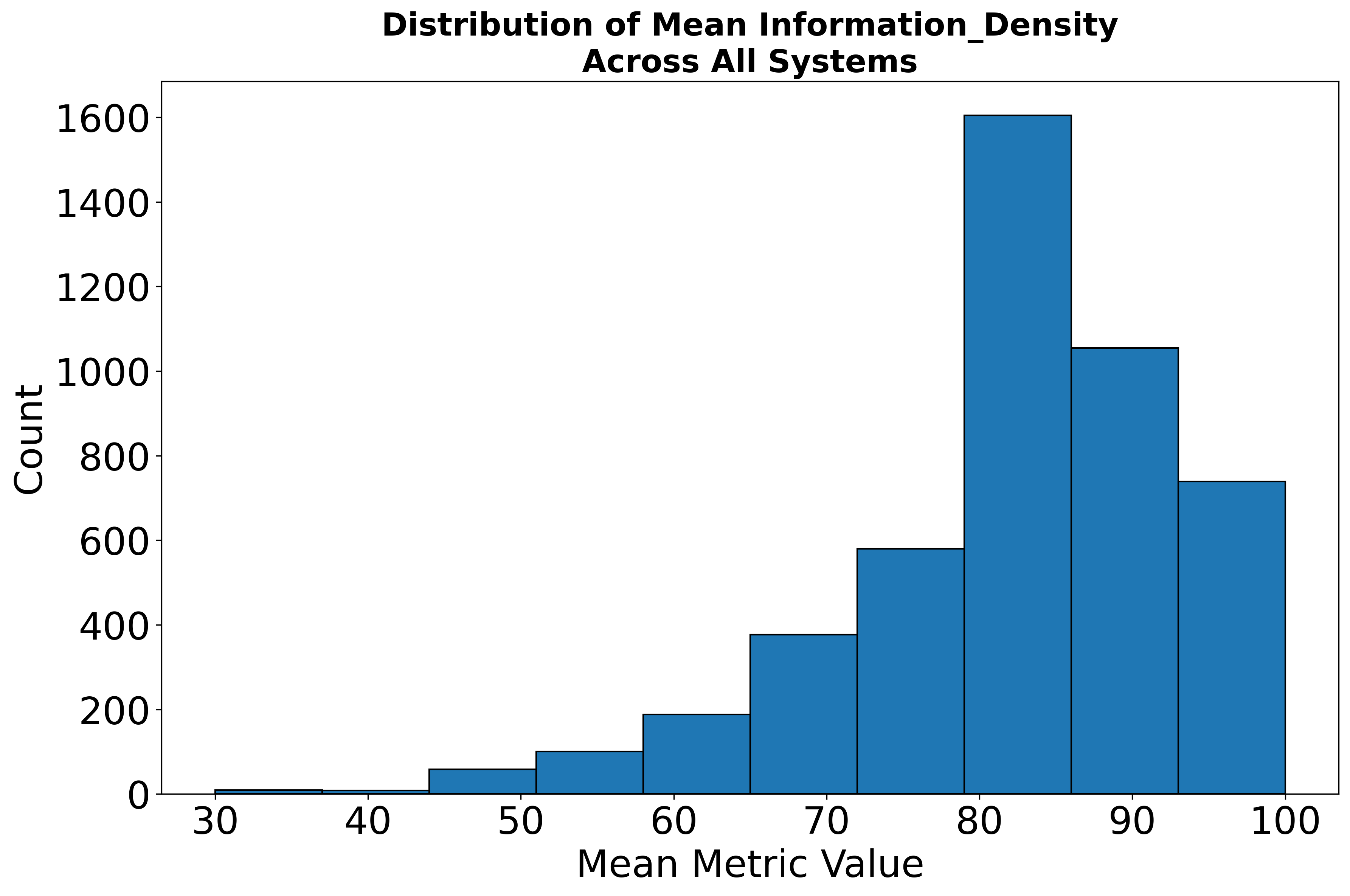}
         \caption{Information Density (ID)}
     \end{subfigure} \\ \vspace{1em} %
     \begin{subfigure}[b]{0.32\textwidth}
         \centering
         \includegraphics[width=\linewidth]{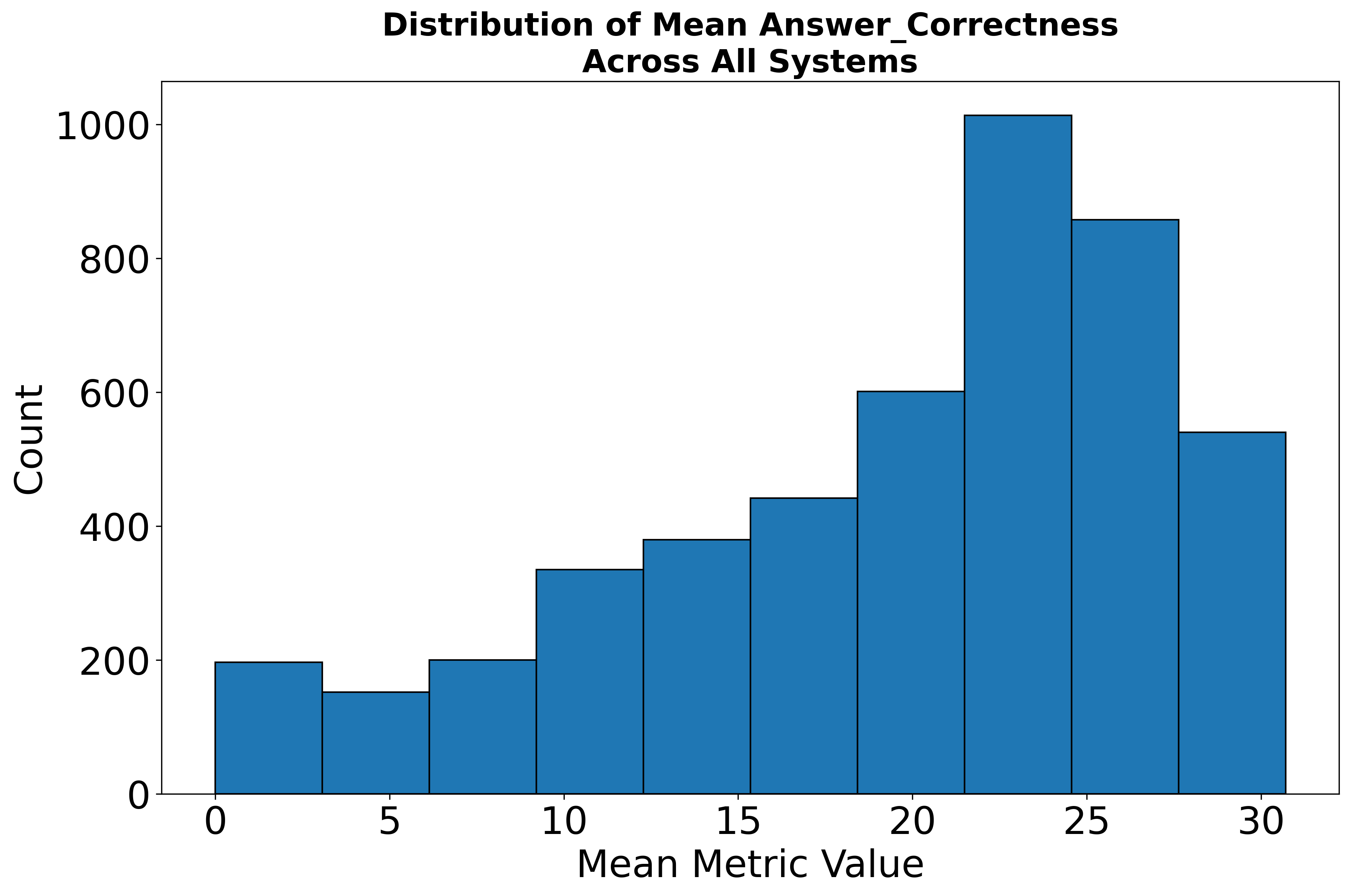}
         \caption{Answer Correctness (AC)}
     \end{subfigure} \hfill
     \begin{subfigure}[b]{0.32\textwidth}
         \centering
         \includegraphics[width=\linewidth]{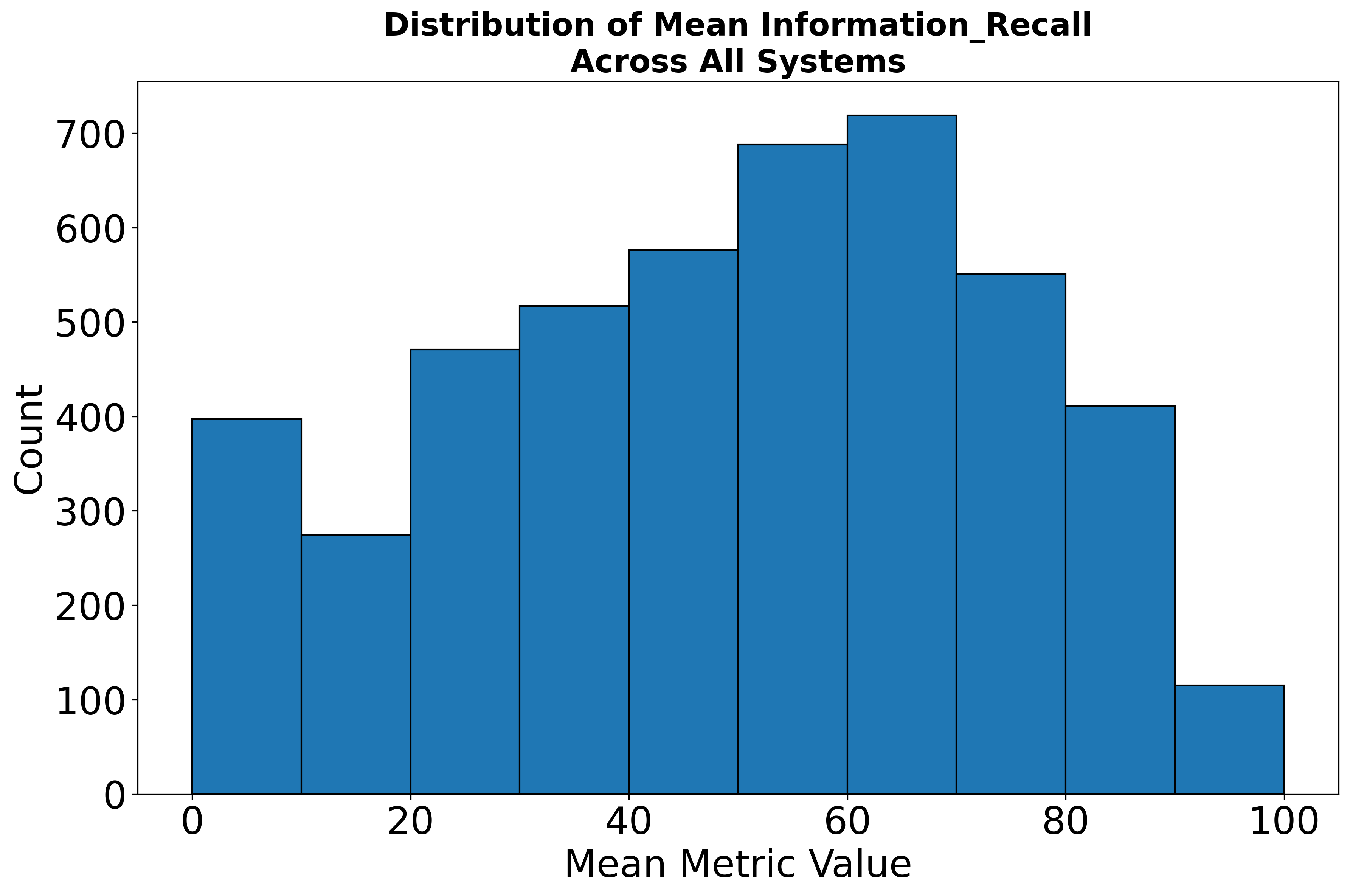}
         \caption{Information Recall (IR)}
     \end{subfigure} \hfill
     \begin{subfigure}[b]{0.32\textwidth} %
         \centering
         \phantom{\includegraphics[width=\linewidth]{figures/histogram_mean_Information_Recall.png}} %
     \end{subfigure}
    \caption{Distribution of Mean CCRS Metric Values Across All Systems (N=4719 queries). These histograms show the frequency (Y-axis) of mean scores (X-axis, range 0-1, averaged across the 6 systems for each query) for each of the five CCRS metrics (a-e), illustrating overall scoring trends.}
    \label{fig:mean_dist_results_detail_v4}
\end{figure*}

Most metrics exhibit a negative skew (left tail), with scores tending towards the upper end of the scale (1.0). This suggests that, on average, the evaluated RAG systems perform reasonably well across these dimensions on the BioASQ dataset.

\textbf{Question Relevance (QR)} shows a particularly pronounced negative skew and a strong ceiling effect. As detailed in \autoref{table:observed_bounds_part1_app}, between 65\% and 85\% of responses received a perfect score of 1.0 for QR, depending on the system. This indicates that QR is effective at identifying poorly relevant answers (assigning low scores) but offers limited granularity for differentiating among highly relevant responses.

\textbf{Answer Correctness (AC)} scores are notably low, with a mean around 0.2 across systems (\autoref{tab:means_main_results_detail_v4}). Crucially, AC never reached the maximum score of 1.0 in our observations (\autoref{table:observed_bounds_part2_app}). This is partly due to the $\lambda=0.7$ weight given to the strict Exact Match component (meaning perfect semantic match without exact match yields max 0.3) and partly reflects the inherent difficulty of achieving perfect factual accuracy in the complex biomedical domain.

\textbf{Information Recall (IR)} displays the widest and most symmetric distribution among the metrics (\autoref{fig:mean_dist_results_detail_v4}(e)). This suggests that IR effectively captures a broad range of completeness levels in the generated answers, making it potentially useful for identifying systems that excel or struggle with providing comprehensive information.

\textbf{Contextual Coherence (CC)} shows distributions whose shapes are highly dependent on the reader model (\autoref{fig:cc_dist} in Appendix). Systems using Llama readers (C, D, E, F) exhibit a higher proportion of scores at the lower bound (0.0) compared to Mistral-based systems (A, B), suggesting more frequent coherence failures (hallucinations or contradictions relative to context) in the Llama models tested. This highlights CC's ability to capture differences in generation quality related to grounding.

\looseness=-1 The observed bounds (\autoref{table:observed_bounds_part1_app}, \autoref{table:observed_bounds_part2_app} in Appendix) confirm these trends, showing high frequencies of perfect scores for QR and CC (especially for Mistral), zero perfect scores for AC, and a wider spread for IR.

\subsubsection{Validity (Convergent \& Discriminant)}
\label{sec:validity}
\looseness=-1
To assess whether the CCRS metrics measure distinct constructs (discriminant validity),
and whether related metrics correlate as expected (convergent validity), we computed Pearson, Spearman, and Kendall correlations between all pairs of metrics. The averaged 
Pearson correlations (using Fisher's Z-transform) across all six systems are shown in \autoref{fig:metric_properties_main_results_detail_v4}(a). Detailed per-system correlation matrices are in \autoref{app:validity_details}.

We observe strong \textbf{convergent validity} between Answer Correctness (AC) and Information Recall (IR), with an average Pearson correlation of r=0.756. This strong positive relationship is expected, as responses that are more complete (higher IR) are often also more likely to contain the correct facts (higher AC), and vice-versa.

We find evidence for \textbf{discriminant validity}, particularly for Contextual Coherence (CC). CC shows only weak to moderate correlations with the other metrics (average Pearson r ranging from 0.201 with AC to 0.452 with IR). This suggests that CC captures a distinct aspect of quality—logical consistency with the context—that is not strongly captured by metrics focusing on relevance (QR), density (ID), accuracy (AC), or completeness (IR).

\looseness=-1
Question Relevance (QR) and Information Density (ID) show a moderate positive correlation (r=0.494), suggesting some relationship: highly relevant answers might also tend to be appropriately dense, or judges might implicitly link these qualities. However, the correlation is not strong enough to suggest they measure the same construct entirely.

Per-system analyses (\autoref{app:validity_details}) reveal some architectural influences. For instance, the correlation between CC and IR is stronger for Llama-based systems (e.g., r=0.528 for System C) than for Mistral-based systems (e.g., r=0.377 for System B), suggesting coherence and recall might be more tightly coupled in Llama architectures.

\subsubsection{Discriminative Power (DP) and Ties}
\label{sec:dp_ties}
\begin{figure}[htbp]
    \centering
    \begin{subfigure}[b]{0.49\linewidth}
        \centering
        \includegraphics[width=\linewidth]{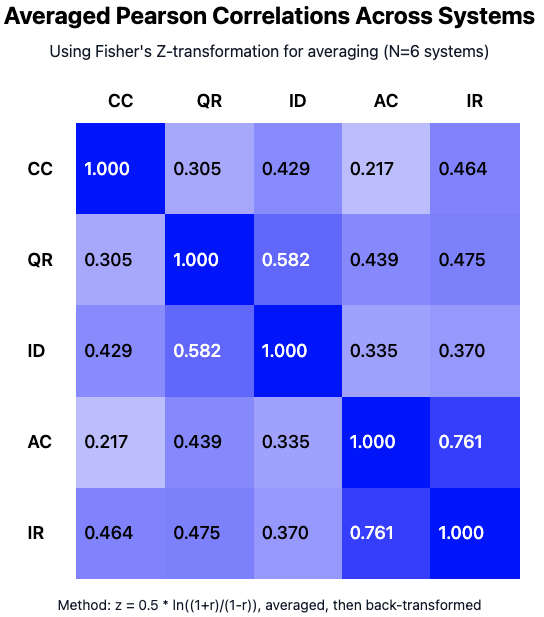}
        \caption{Avg. Pearson Correlation}
        \label{fig:avg_pearson_corr_main_results_detail_v4}
    \end{subfigure}
    \hfill %
    \begin{subfigure}[b]{0.49\linewidth}
        \centering
        \includegraphics[width=\linewidth]{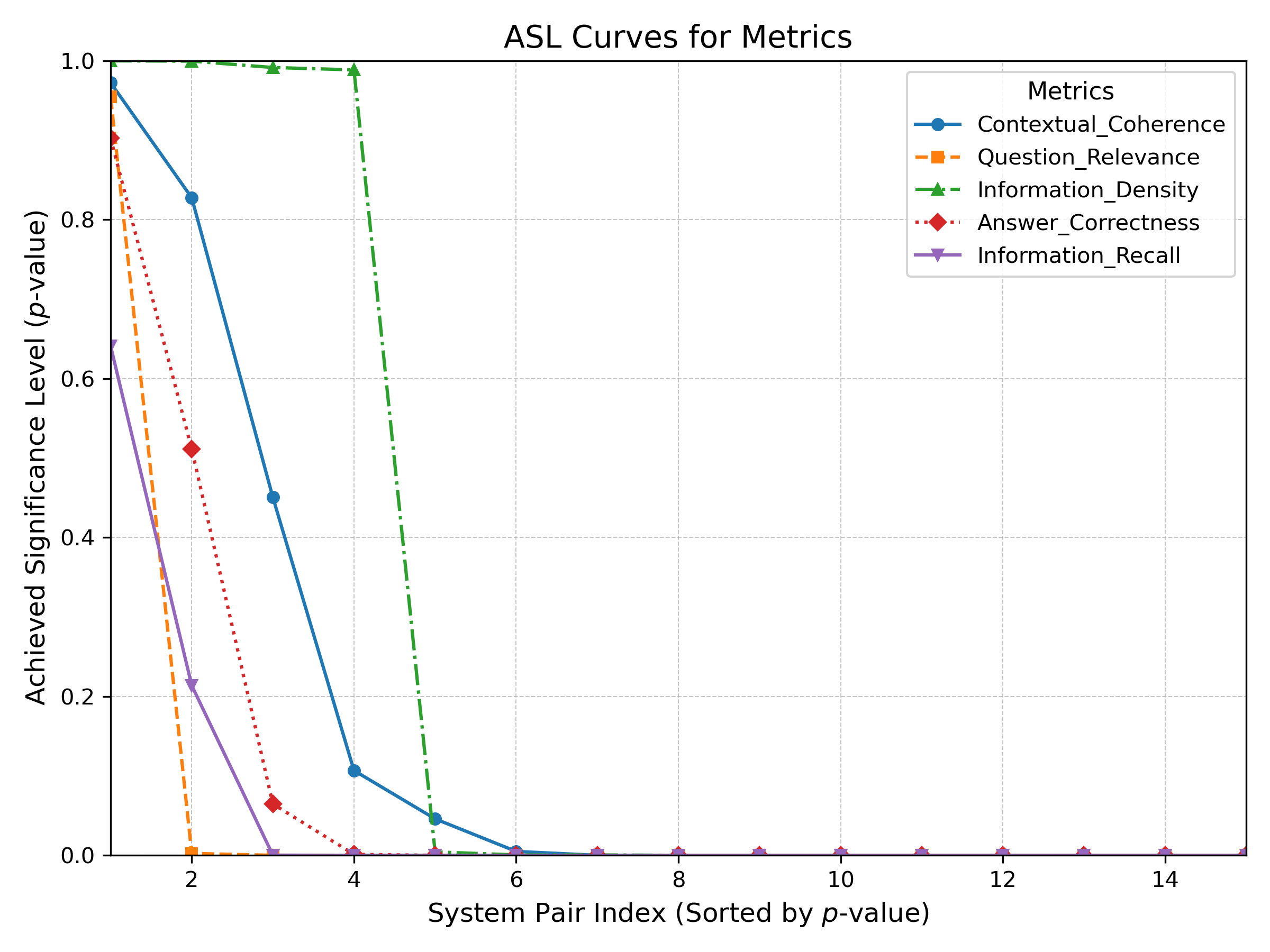}
        \caption{ASL Curves (DP)}
        \label{fig:asl_curves_main_results_detail_v4}
    \end{subfigure}
    \caption{CCRS Metric Validity (left) assessed via averaged Pearson correlations across 6 systems, and Discriminative Power (right) visualized with Achieved Significance Level (ASL) curves.}
    \label{fig:metric_properties_main_results_detail_v4}
    \vspace{-1em}
\end{figure}

An important property of an evaluation metric is its ability to reliably distinguish between systems that perform differently (discriminative power). We assessed DP using pairwise statistical tests (Tukey's HSD with randomization, B=10,000, $\alpha=0.05$) between all $\binom{6}{2}=15$ pairs of systems for each metric. DP is calculated as the fraction of pairs for which a statistically significant difference was found. Results are summarized in \autoref{tab:dp_summary_main_results_detail_v4}, and visualized using Achieved Significance Level (ASL) curves in \autoref{fig:metric_properties_main_results_detail_v4}(b).

\textbf{Question Relevance (QR)} exhibits the highest DP (0.9333), detecting significant differences in 14 out of 15 system pairs. This indicates high sensitivity to performance variations in relevance. \textbf{Information Recall (IR)} follows closely with DP=0.8667 (13/15 pairs), and \textbf{Answer Correctness (AC)} also shows good DP=0.8000 (12/15 pairs). \textbf{Contextual Coherence (CC)} and \textbf{Information Density (ID)} have lower DP (both 0.7333, 11/15 pairs), suggesting they are slightly less sensitive in distinguishing between the systems evaluated here.

However, DP must be considered alongside the rate of ties (\autoref{tab:empirical-tie-probs_app}). QR, despite its high DP, suffers from a very high empirical tie rate (ranging from 47\% to 74\% depending on the system) due to the ceiling effect discussed earlier. This means that while QR effectively separates systems with different average relevance, it offers poor granularity for distinguishing between individual responses that are all deemed highly relevant.

Conversely, AC and IR have the lowest tie rates (AC: 15-18\%, IR: 15-18\%). Their combination of good DP and low ties suggests they offer better potential for fine-grained differentiation between RAG outputs, capturing variations in accuracy and completeness more granularly than QR captures variations in relevance at the top end of the scale.

\begin{table}[htbp]
\centering
\caption{Discriminative Power (DP) of CCRS Metrics based on pairwise Tukey's HSD tests ($\alpha=0.05$) across 6 systems (15 pairs).}
\label{tab:dp_summary_main_results_detail_v4}
\small \vspace{-0.5em}
\begin{tabular}{@{}lcc@{}}
\toprule
Metric & Significant Pairs & DP \\
\midrule
Question Relevance (QR) & 14 / 15 & 0.9333 \\
Information Recall (IR) & 13 / 15 & 0.8667 \\
Answer Correctness (AC) & 12 / 15 & 0.8000 \\
Contextual Coherence (CC) & 11 / 15 & 0.7333 \\
Information Density (ID) & 11 / 15 & 0.7333 \\
\bottomrule
\end{tabular}
\vspace{-1em}
\end{table}

\subsubsection{Comparison with RAGChecker}
\label{sec:ragchecker_comp}
To benchmark CCRS against a state-of-the-art but complex RAG evaluation framework, we compared it to RAGChecker \citep{ru2024ragchecker}. We focused on RAGChecker's core end-to-end metrics: Precision (P), Recall (R), and Faithfulness (F), which rely on claim extraction and entailment checking. We used \texttt{Llama-70B-Instruct} for claim extraction in RAGChecker to maintain consistency with the CCRS judge model. Full details, including distributions, bounds, ties, and population stats for RAGChecker metrics, are in \autoref{app:ragchecker_details}.

\textbf{Distributions/Bounds/Ties:} RAGChecker Faithfulness (F) exhibited a strong ceiling effect, similar to CCRS QR, with high mean scores, 64-74\% of responses achieving a perfect score of 1.0 (\autoref{table:observed_bounds_ragchecker_app}), and high tie rates (44-54\%, \autoref{tab:empirical-tie-probs-all_app}). In contrast, RAGChecker Precision (P) and Recall (R) showed more balanced distributions (\autoref{tab:pop_stats_ragchecker_app}) and low tie rates (P: 13-20\%, R: 9-10\%), behaving similarly to CCRS AC and IR.

\begin{figure}[htbp]
\centering
\includegraphics[width=0.9\linewidth]{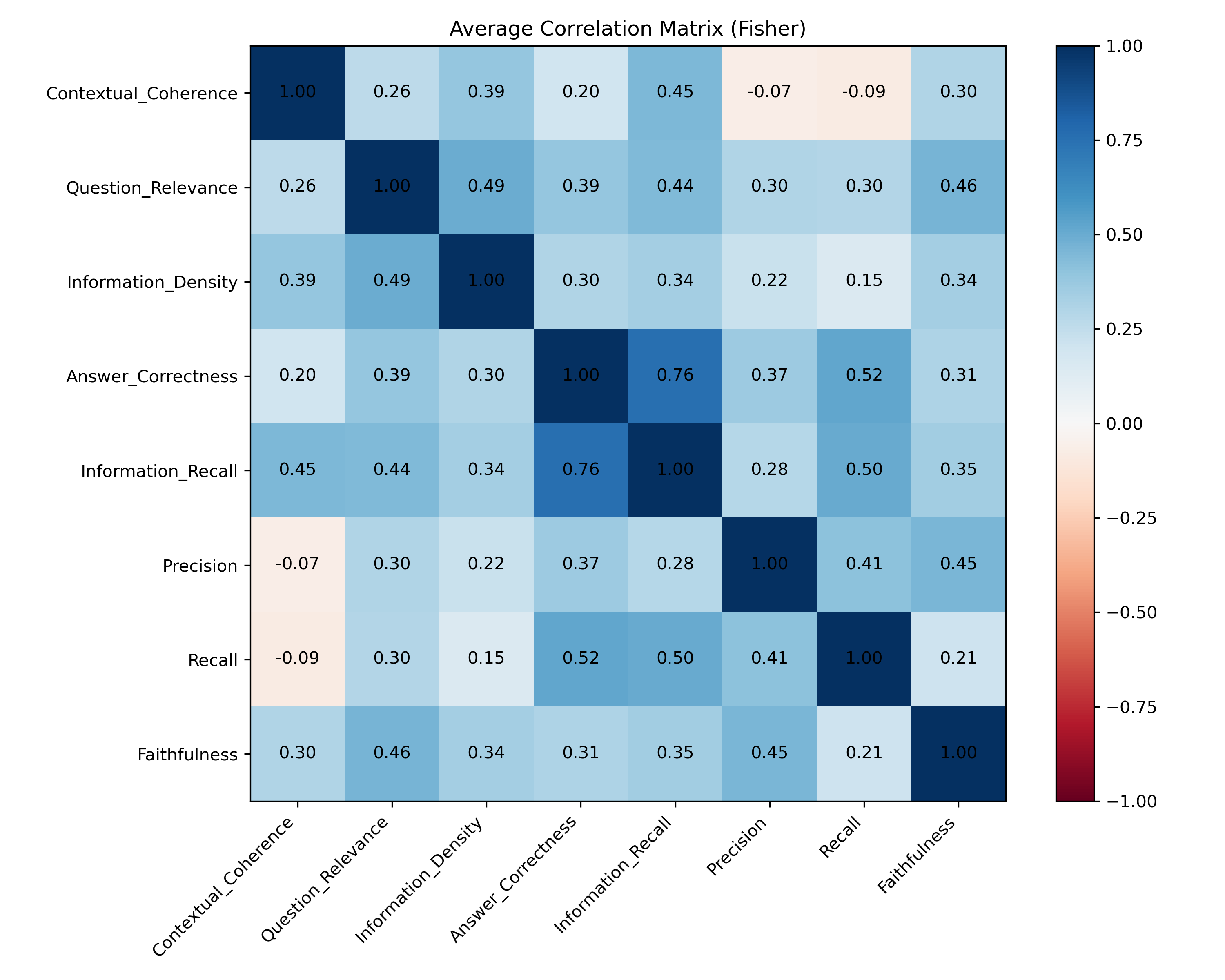}
\caption{Averaged Pearson Correlation Matrix comparing CCRS metrics (CC, QR, ID, AC, IR) and RAGChecker metrics (P, R, F) across 6 systems.}
\label{fig:full_correlation_matrix_main_results_detail_v4}
\vspace{-1em}
\end{figure}

\textbf{Validity:} We examined the correlation between CCRS and \\ RAGChecker metrics~(\autoref{fig:full_correlation_matrix_main_results_detail_v4}). RAGChecker Recall (R) showed moderate-to-strong correlations with CCRS AC (r=0.518) and CCRS IR (r=0.504), confirming they measure related aspects of correctness and completeness. RAGChecker Faithfulness (F) correlated most strongly with CCRS QR (r=0.463) and moderately with CCRS CC (r=0.304), suggesting overlap in assessing relevance and coherence or grounding. RAGChecker Precision (P) showed moderate correlation with CCRS AC (r=0.366) but weaker links to other CCRS metrics.

\textbf{Discriminative Power (DP):} We compared the DP of key metrics (\autoref{tab:dp_comparison_main_results_detail_v4}). CCRS QR (DP=0.933) and IR (DP=0.867) significantly outperformed RAGChecker's Recall (R, DP=0.800) and Faithfulness (F, DP=0.800). CCRS AC (DP=0.800) matched the DP of RAGChecker R and F. RAGChecker Precision (P) showed very poor discriminative power (DP=0.200) in our experiments.

\begin{table}[htbp]
\centering
\caption{Discriminative Power (DP) Comparison: CCRS vs. RAGChecker (Core Metrics).}
\label{tab:dp_comparison_main_results_detail_v4}
\small \vspace{-0.5em}
\begin{tabular}{@{}lc|lc@{}}
\toprule
CCRS Metric & DP & RAGChecker Metric & DP \\
\midrule
QR & \textbf{0.9333} & Faithfulness (F) & 0.8000 \\
IR & \textbf{0.8667} & Recall (R) & 0.8000 \\
AC & 0.8000 & Precision (P) & 0.2000 \\
CC & 0.7333 & & \\
ID & 0.7333 & & \\
\bottomrule
\end{tabular}
\vspace{-1em}
\end{table}

\textbf{Efficiency:} A significant advantage of CCRS is its computational efficiency. In our setup, running the five zero-shot CCRS evaluations using \texttt{Llama-70B} was approximately \textbf{5 times faster} than executing the RAGChecker pipeline (claim extraction using \texttt{Llama-70B} + entailment checks) for P, R, and F. This difference stems from CCRS avoiding the computationally intensive claim extraction and multiple pairwise entailment steps required by RAGChecker.

\textbf{Conclusion vs. RAGChecker:} CCRS offers comparable or superior discriminative power for key aspects of RAG evaluation (especially relevance via QR, completeness via IR, and correctness via AC) compared to RAGChecker's core metrics (F, R, P), while being significantly more computationally efficient and simpler to implement due to its zero-shot, end-to-end nature.

\subsection{RAG System Performance and Hypothesis Testing}
\label{sec:hypothesis_testing_results}
\begin{figure*}[tb]
    \centering
    \begin{subfigure}[b]{0.32\textwidth}
        \centering
        \includegraphics[width=\linewidth]{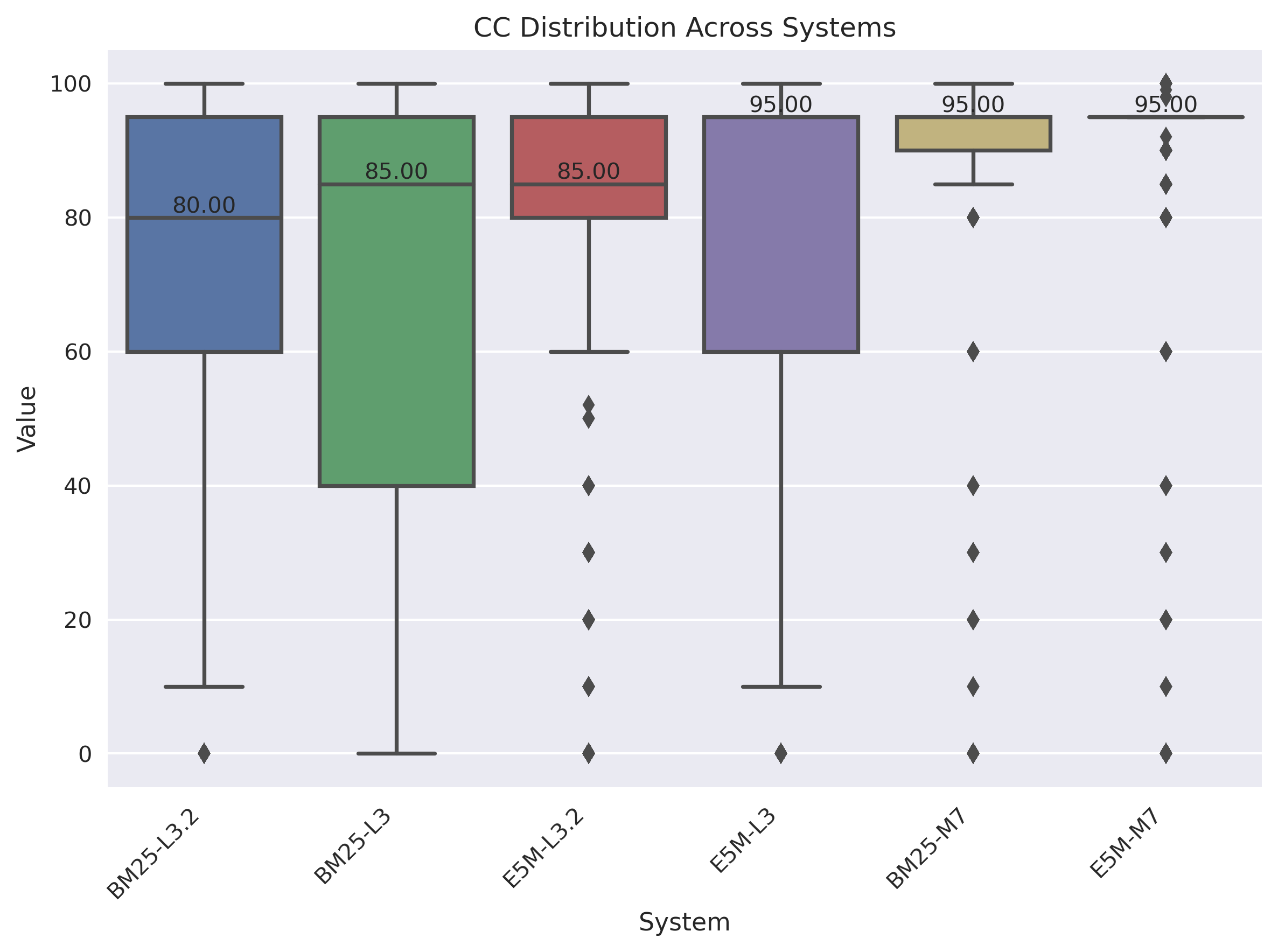}
        \caption{Contextual Coherence (CC)}
        \label{fig:boxplot_cc_main}
    \end{subfigure}
    \hfill %
    \begin{subfigure}[b]{0.32\textwidth}
        \centering
        \includegraphics[width=\linewidth]{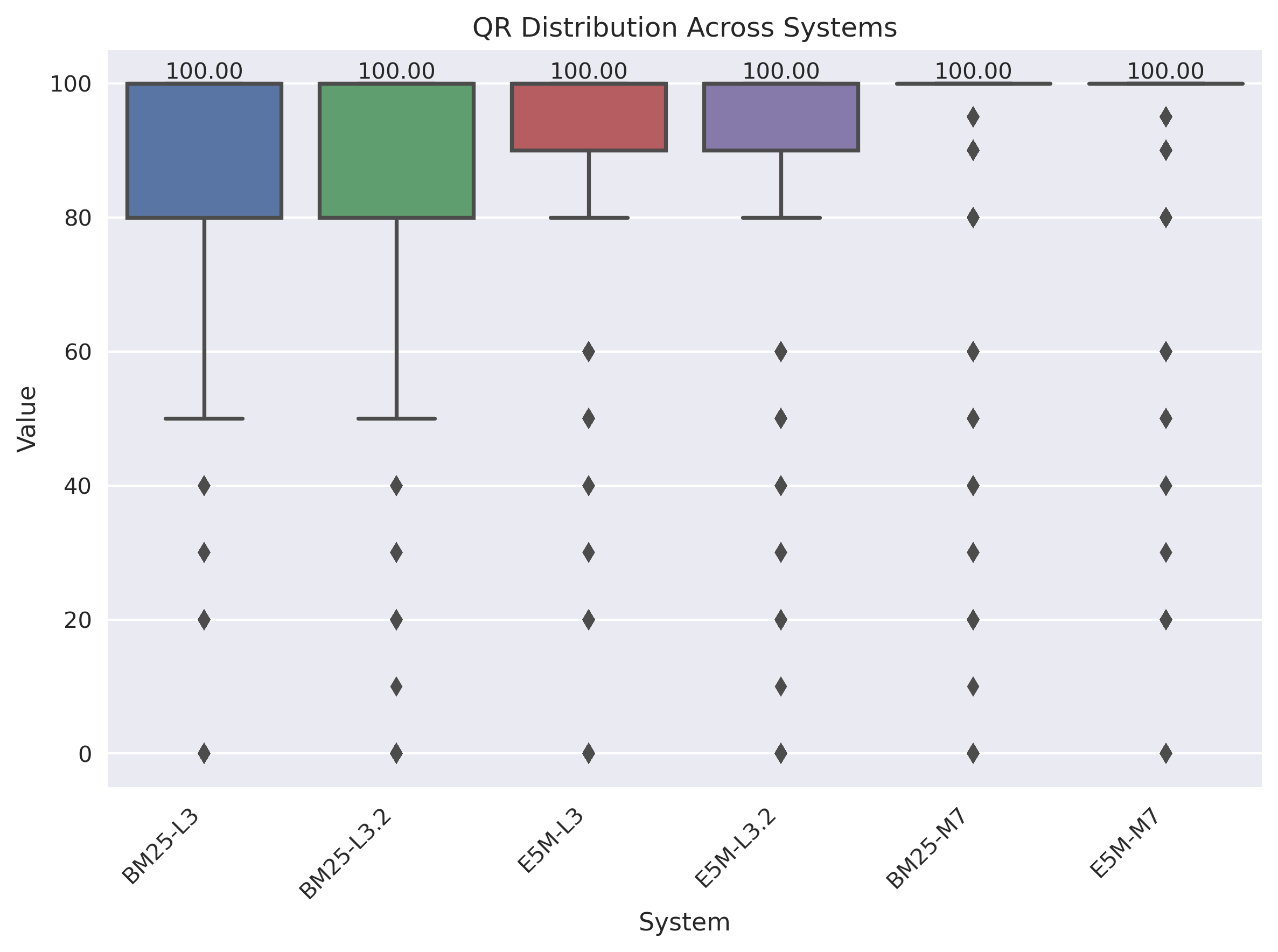}
        \caption{Question Relevance (QR)}
        \label{fig:boxplot_qr_main}
    \end{subfigure}
    \hfill %
    \begin{subfigure}[b]{0.32\textwidth}
        \centering
        \includegraphics[width=\linewidth]{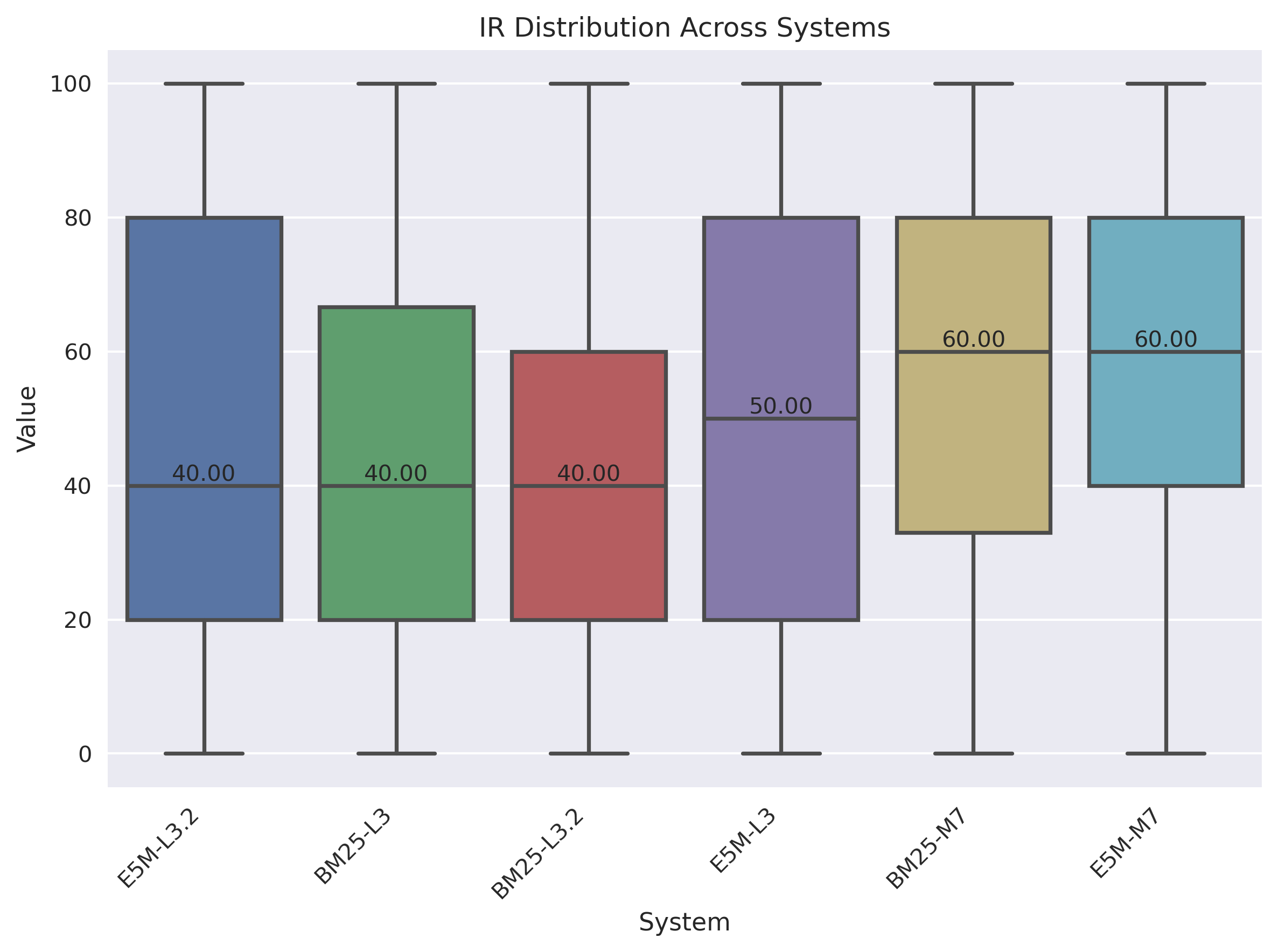}
        \caption{Information Recall (IR)}
        \label{fig:boxplot_ir_main}
    \end{subfigure}
    \vspace{1em}
    \begin{subfigure}[b]{0.32\textwidth}
        \centering
        \includegraphics[width=\linewidth]{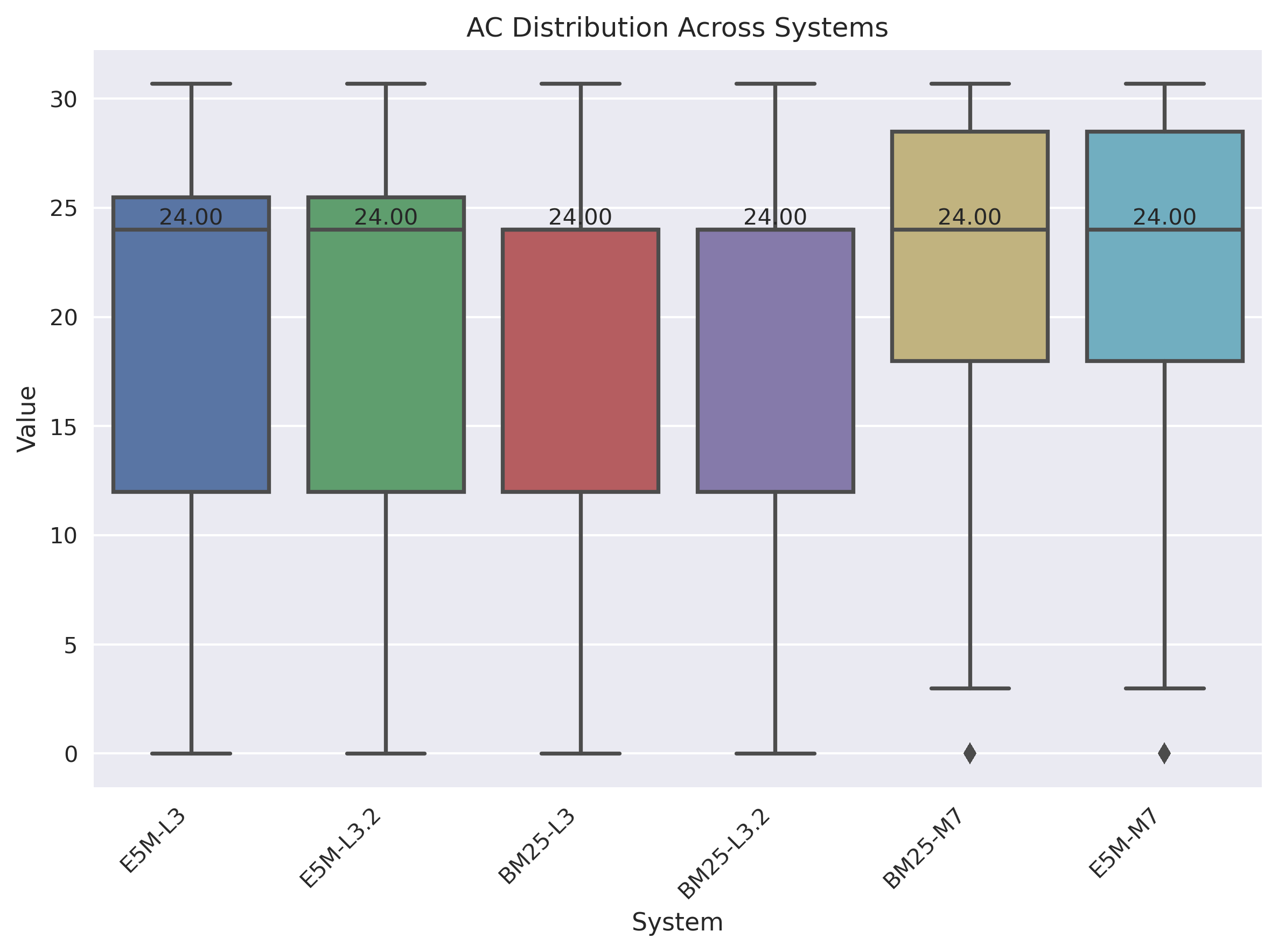}
        \caption{Answer Correctness (AC)}
        \label{fig:boxplot_ac_main}
    \end{subfigure}
    \hfill %
    \begin{subfigure}[b]{0.32\textwidth}
        \centering
        \includegraphics[width=\linewidth]{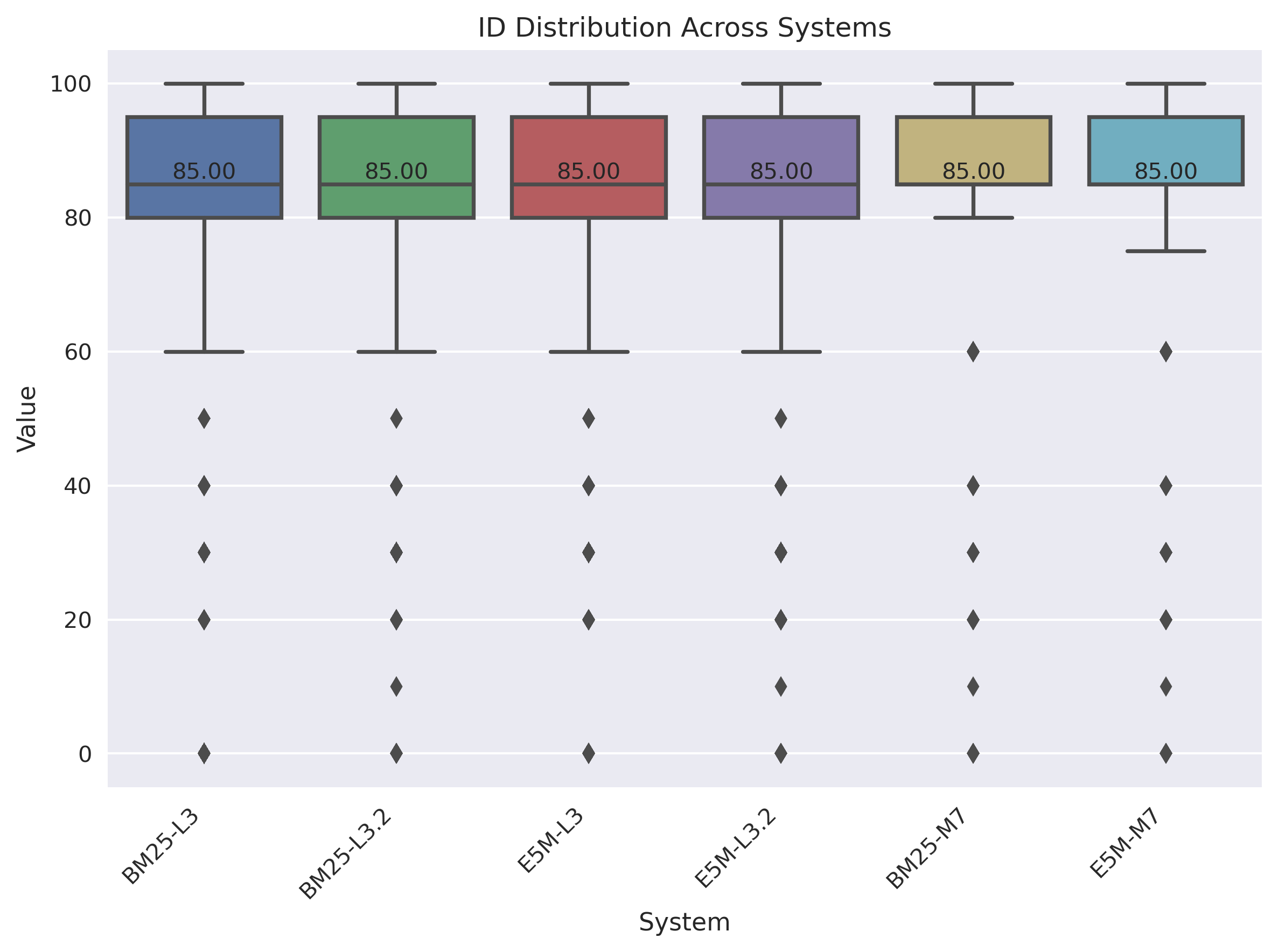}
        \caption{Information Density (ID)}
        \label{fig:boxplot_id_main}
    \end{subfigure}
    \hfill %
    \begin{subfigure}[b]{0.32\textwidth}
        \centering
        \includegraphics[width=\linewidth]{figures/asl_curves_camera_ready.png} %
        \caption{ASL Curves (DP Summary)}
        \label{fig:asl_curves_summary_main}
    \end{subfigure}
    \caption{Box plots illustrating performance distributions for all CCRS metrics across the six RAG systems (A-F), plus ASL curves summarizing discriminative power. Systems within each plot are sorted by median performance. System Key: A(M+BM25), B(M+E5), C(L8+BM25), D(L8+E5), E(L3.2+BM25), F(L3.2+E5).}
    \label{fig:boxplots_perf_main_results_detail_perf_v4}
\end{figure*}

We used the CCRS framework to evaluate the six RAG system configurations (A-F). Mean performance scores are presented in \autoref{tab:means_main_results_detail_v4}, and performance distributions are visualized using box plots in \autoref{fig:boxplots_perf_main_results_detail_perf_v4}. We tested three pre-defined hypotheses using paired Tukey's HSD tests (B=10,000 permutations) with Holm-Bonferroni correction across hypotheses ($\alpha=0.05$). Detailed statistical test results are available in \autoref{app:stats_details}. All three hypotheses were supported by the data.

\begin{table}[htb]
\small
\centering
\caption{Mean Performance Scores (\%) by System and CCRS Metric}
\label{tab:means_main_results_detail_v4} \vspace{-0.5em}
\begin{tabular}{@{}clrrrrr@{}}
\toprule
Sys. & Config. & CC & QR & ID & AC & IR \\
\midrule
A & Mistral+BM25 & 87.95 & 92.83 & 86.81 & 21.10 & 54.96 \\
B & Mistral+E5   & 88.61 & 95.40 & 86.90 & 21.81 & 58.28 \\
C & Llama8B+BM25 & 67.89 & 87.07 & 79.09 & 18.54 & 43.00 \\
D & Llama8B+E5   & 69.29 & 90.55 & 80.41 & 19.24 & 45.44 \\
E & Llama3.2+BM25& 68.88 & 88.28 & 80.23 & 18.41 & 42.04 \\
F & Llama3.2+E5  & 70.88 & 90.83 & 80.43 & 18.76 & 44.79 \\
\bottomrule
\end{tabular}
\vspace{-1em}
\end{table}

\textbf{H1: Mistral-7B Reader Superiority (Adjusted p < 0.0001):} The results overwhelmingly support the hypothesis that Mistral-7B significantly outperforms both \texttt{Llama3-8B} and \texttt{Llama3.2-3B} readers across all five CCRS metrics, regardless of the retriever used. Comparing systems B (Mistral+E5) and D (Llama8B+E5), Mistral showed large advantages in CC (+19.3 points), QR (+4.9 points), ID (+6.5 points), AC (+2.6 points), and IR (+12.8 points). Similar significant advantages were observed when comparing A vs C, A vs E, and B vs F (see \autoref{tab:app_stats_cc_full_final} to \autoref{tab:app_stats_ir_full_final} for all pairwise p-values). The box plots in \autoref{fig:boxplots_perf_main_results_detail_perf_v4} visually confirm these findings, showing consistently higher median scores and often tighter interquartile ranges (IQRs) for Mistral-based systems (A, B) compared to Llama-based systems (C, D, E, F).

\textbf{H2: E5 Retriever Advantage for Llama Models (Adjusted p < 0.0001):} The hypothesis that the E5 neural retriever provides benefits over BM25 specifically for QR and IR when used with Llama readers was strongly supported. Comparing System D (Llama8B+E5) vs. System C (Llama8B+BM25), E5 yielded significant improvements in QR (+3.48 points, p<0.0001) and IR (+2.44 points, p<0.0001).

Similarly, comparing System F (Llama3.2+E5) vs.   System E \\ (Llama3.2+BM25), E5 led to significant gains in QR (+2.55 points, p<0.0001) and IR (+2.74 points, p<0.0001). These results highlight the targeted benefits of neural retrieval for enhancing relevance and completeness, particularly with Llama models in this task.

\textbf{H3: Performance Differences Between Llama Models (Final adj. p = 0.0070):} The hypothesis that there exists a significant performance difference between the \texttt{Llama3-8B} and \texttt{Llama3.2-3B} readers was supported. The differences varied across metrics. \\ \texttt{Llama3.2-3B} showed significant advantages in QR (System E > C, p=0.0037), ID (System E > C, p=0.0032), and CC (System F > D, p=0.0417). Conversely, \texttt{Llama3-8B} demonstrated a significant advantage in AC (System D > F, p=0.0014). No significant difference was found for IR (p=0.2171 for C vs E, p=0.6328 for D vs F). This indicates a nuanced trade-off, with the smaller Llama3.2 potentially benefiting from higher-quality training data for coherence and relevance, while the larger \texttt{Llama3-8B} maintains an edge in factual accuracy.

\section{Discussion}
\label{sec:discussion}
Our evaluation using the CCRS framework on the BioASQ dataset provides several important insights into RAG system performance and evaluation methodologies.

\textbf{Interpreting RAG System Performance:}
The results strongly underscore the critical role of the generator (reader) LLM in determining overall RAG performance. Mistral-7B's consistent and statistically significant superiority across all five CCRS dimensions (supporting H1) suggests that its architecture or training provides substantial advantages in synthesizing information, maintaining coherence, and extracting relevant facts from retrieved context compared to the Llama 3 and 3.2 models tested. The performance gaps were particularly striking for Contextual Coherence (approx. +19-20 points) and Information Recall (approx. +10-13 points), indicating that the choice of reader model can have a profound impact on the quality and completeness of the generated output, potentially outweighing differences stemming from the retrieval component.

The targeted benefits observed for the E5 neural retriever (supporting H2) highlight the value of semantic retrieval. E5 significantly improved Question Relevance and Information Recall specifically for the Llama-based systems. This suggests E5 successfully retrieved context that was more topically aligned with the query and contained more of the necessary factual elements present in the ground truth. However, a crucial finding is that these improvements in retrieval quality (QR, IR) did not translate into statistically significant gains in Answer Correctness (AC) for the same Llama systems (D vs C, F vs E). This points towards a potential bottleneck at the generation stage: providing more relevant or complete context does not automatically guarantee a more factually accurate generated response. The LLM's ability to faithfully extract, synthesize without distortion, and avoid introducing its own errors or hallucinations remains a distinct challenge, emphasizing the need for evaluation frameworks like CCRS that assess both faithfulness/coherence (CC) and factual accuracy (AC) independently.

The comparison between \texttt{Llama3-8B} and \texttt{Llama3.2-3B} (supporting H3) revealed a nuanced relationship between model scale, training data, and performance dimensions. The smaller \texttt{Llama3.2-3B}, potentially benefiting from higher-quality (possibly multimodal) training data, excelled in QR, ID, and CC. This suggests specialized training can enhance aspects related to query understanding, structured text generation, and conciseness, even in smaller models. Conversely, the larger \texttt{Llama3-8B} maintained an edge in AC, possibly due to better factual recall or more precise generation capabilities associated with its scale. This complex outcome suggests that optimizing RAG systems might involve trade-offs, requiring component selection based on which quality dimensions (e.g., coherence vs. factual precision) are most critical for the specific application.

\textbf{Evaluating the Task and Dataset:}
The BioASQ dataset served as an effective and challenging benchmark, capable of discriminating clearly between the performance levels of different RAG systems. The overall performance profile observed—generally high scores for QR and CC, but significantly lower scores for AC and IR—underscores the inherent difficulty of high-fidelity question answering in specialized domains like biomedicine. Generating fluent and relevant text appears considerably less challenging for current models than ensuring absolute factual correctness and comprehensive coverage of information according to an expert standard. This highlights the importance of including metrics like AC and IR alongside QR and CC for a holistic evaluation, particularly in domains where reliability and completeness are paramount.

\textbf{Reflecting on CCRS Metrics and Framework:}
CCRS proved to be a viable, efficient, and effective evaluation framework for RAG systems. Its multi-dimensional nature successfully captured distinct aspects of response quality, as validated by the correlation analysis (e.g., the strong AC-IR link indicating convergent validity, and CC's weak correlations with others suggesting discriminant validity). The discriminative power analysis identified QR and IR as highly sensitive metrics in our experimental setup, capable of detecting subtle system differences, while AC also performed well.

The comparison with the RAGChecker framework (\autoref{sec:ragchecker_comp}) showed that CCRS achieved comparable or superior discriminative power for core aspects of RAG quality (approximating recall via IR, correctness via AC, and aspects of faithfulness via CC/QR) compared to RAGChecker's key metrics (R, F, P). Crucially, CCRS achieved this with significantly reduced computational overhead (5x faster in our setup) and implementation complexity, as it avoids the need for intermediate claim extraction and entailment checking. This strongly supports the potential and practicality of using zero-shot, end-to-end LLM judgments for comprehensive RAG evaluation.

However, CCRS is not without limitations and challenges. The pronounced ceiling effect observed for QR, while indicating good performance on relevance, limits its granularity for differentiating between top-performing systems. The consistently low absolute scores and absence of perfect scores for AC warrant further investigation into its calibration, the influence of the EM component, and the fundamental difficulty of exact factual matching in complex domains. The lower DP observed for CC and ID might reflect limitations in the judge LLM's ability to consistently discern subtle differences in these more subjective qualities, or it might simply indicate smaller actual performance gaps between the systems on these dimensions. A significant limitation of the current study is the lack of direct human correlation data; the alignment between CCRS scores and human perception of quality remains to be formally established. Furthermore, interpretability remains a general challenge for all LLM-based metrics, including CCRS. Understanding \emph{why} the judge assigned a particular score can be difficult.

\textbf{Future Directions:} Based on our findings and limitations, priority future directions include: (1) Validating CCRS across diverse domains and datasets to assess its generalizability. (2) Conducting rigorous human correlation studies to establish the alignment between CCRS metrics and human judgments of RAG quality. (3) Analyzing the sensitivity of CCRS scores to the choice of judge LLM and prompt variations. (4) Developing methods to enhance the interpretability of CCRS scores, potentially by prompting the judge for justifications alongside scores. (5) Exploring principled approaches to combine the five CCRS metrics into a single composite score, possibly weighted based on user studies or task requirements.

\section{Conclusion}
\label{sec:conclusion}
Evaluating the multifaceted quality of RAG systems remains a significant challenge, requiring methods that go beyond traditional metrics. To address the need for comprehensive yet efficient evaluation, we introduced CCRS, a novel suite of five metrics (Contextual Coherence, Question Relevance, Information Density, Answer Correctness, Information Recall). CCRS leverages a single, powerful, pretrained LLM (Llama 70B) as a zero-shot, end-to-end judge, thereby avoiding the complexities of multi-stage pipelines or extensive fine-tuning required by previous frameworks.

\bibliographystyle{ACM-Reference-Format}
\bibliography{references}


\begin{thebibliography}{24}


\ifx \showCODEN    \undefined \def \showCODEN     #1{\unskip}     \fi
\ifx \showISBNx    \undefined \def \showISBNx     #1{\unskip}     \fi
\ifx \showISBNxiii \undefined \def \showISBNxiii  #1{\unskip}     \fi
\ifx \showISSN     \undefined \def \showISSN      #1{\unskip}     \fi
\ifx \showLCCN     \undefined \def \showLCCN      #1{\unskip}     \fi
\ifx \shownote     \undefined \def \shownote      #1{#1}          \fi
\ifx \showarticletitle \undefined \def \showarticletitle #1{#1}   \fi
\ifx \showURL      \undefined \def \showURL       {\relax}        \fi
\providecommand\bibfield[2]{#2}
\providecommand\bibinfo[2]{#2}
\providecommand\natexlab[1]{#1}
\providecommand\showeprint[2][]{arXiv:#2}

\bibitem[Asai et~al\mbox{.}(2024)]%
        {asai2024reliable}
\bibfield{author}{\bibinfo{person}{Akari Asai}, \bibinfo{person}{Zexuan Zhong}, \bibinfo{person}{Danqi Chen}, \bibinfo{person}{Pang~Wei Koh}, \bibinfo{person}{Luke Zettlemoyer}, \bibinfo{person}{Hannaneh Hajishirzi}, {and} \bibinfo{person}{Wen-tau Yih}.} \bibinfo{year}{2024}\natexlab{}.
\newblock \showarticletitle{Reliable, adaptable, and attributable language models with retrieval}.
\newblock \bibinfo{journal}{\emph{arXiv preprint arXiv:2403.03187}} (\bibinfo{year}{2024}).
\newblock


\bibitem[Chiang and Lee(2023)]%
        {chiang-lee-2023-large}
\bibfield{author}{\bibinfo{person}{Cheng-Han Chiang} {and} \bibinfo{person}{Hung-yi Lee}.} \bibinfo{year}{2023}\natexlab{}.
\newblock \showarticletitle{Can large language models be an alternative to human evaluations?}
\newblock \bibinfo{journal}{\emph{arXiv preprint arXiv:2305.01937}} (\bibinfo{year}{2023}).
\newblock


\bibitem[Dubey et~al\mbox{.}(2024)]%
        {dubey2024Llama}
\bibfield{author}{\bibinfo{person}{Abhimanyu Dubey}, \bibinfo{person}{Abhinav Jauhri}, \bibinfo{person}{Abhinav Pandey}, \bibinfo{person}{Abhishek Kadian}, \bibinfo{person}{Ahmad Al-Dahle}, \bibinfo{person}{Aiesha Letman}, \bibinfo{person}{Akhil Mathur}, \bibinfo{person}{Alan Schelten}, \bibinfo{person}{Amy Yang}, \bibinfo{person}{Angela Fan}, {et~al\mbox{.}}} \bibinfo{year}{2024}\natexlab{}.
\newblock \showarticletitle{The llama 3 herd of models}.
\newblock \bibinfo{journal}{\emph{arXiv preprint arXiv:2407.21783}} (\bibinfo{year}{2024}).
\newblock


\bibitem[Es et~al\mbox{.}(2023)]%
        {es2023ragas}
\bibfield{author}{\bibinfo{person}{Shahul Es}, \bibinfo{person}{Jithin James}, \bibinfo{person}{Luis Espinosa-Anke}, {and} \bibinfo{person}{Steven Schockaert}.} \bibinfo{year}{2023}\natexlab{}.
\newblock \showarticletitle{Ragas: Automated evaluation of retrieval augmented generation}.
\newblock \bibinfo{journal}{\emph{arXiv preprint arXiv:2309.15217}} (\bibinfo{year}{2023}).
\newblock


\bibitem[Ferrara et~al\mbox{.}(2024)]%
        {ferrara2024rag}
\bibfield{author}{\bibinfo{person}{Joe Ferrara}, \bibinfo{person}{Ethan-Tonic}, {and} \bibinfo{person}{Oguzhan~Mete Ozturk}.} \bibinfo{year}{2024}\natexlab{}.
\newblock \bibinfo{booktitle}{\emph{The RAG Triad}}.
\newblock
\newblock
\shownote{\url{https://www.trulens.org/trulens_eval/core_concepts_rag_triad/}}.


\bibitem[Gao et~al\mbox{.}(2023)]%
        {gao2023retrieval}
\bibfield{author}{\bibinfo{person}{Yunfan Gao}, \bibinfo{person}{Yun Xiong}, \bibinfo{person}{Xinyu Gao}, \bibinfo{person}{Kangxiang Jia}, \bibinfo{person}{Jinliu Pan}, \bibinfo{person}{Yuxi Bi}, \bibinfo{person}{Yi Dai}, \bibinfo{person}{Jiawei Sun}, {and} \bibinfo{person}{Haofen Wang}.} \bibinfo{year}{2023}\natexlab{}.
\newblock \showarticletitle{Retrieval-augmented generation for large language models: A survey}.
\newblock \bibinfo{journal}{\emph{arXiv preprint arXiv:2312.10997}} (\bibinfo{year}{2023}).
\newblock


\bibitem[Huang et~al\mbox{.}(2023)]%
        {ye2023large}
\bibfield{author}{\bibinfo{person}{Jie Huang}, \bibinfo{person}{Xinyun Chen}, \bibinfo{person}{Swaroop Mishra}, \bibinfo{person}{Huaixiu~Steven Zheng}, \bibinfo{person}{Adams~Wei Yu}, \bibinfo{person}{Xinying Song}, {and} \bibinfo{person}{Denny Zhou}.} \bibinfo{year}{2023}\natexlab{}.
\newblock \bibinfo{title}{Large Language Models Cannot Self-Correct Reasoning Yet}.
\newblock
\showeprint[arxiv]{2310.01798}~[cs.CL]


\bibitem[Jiang et~al\mbox{.}(2023)]%
        {jiang2023mistral}
\bibfield{author}{\bibinfo{person}{Albert~Q Jiang}, \bibinfo{person}{Alexandre Sablayrolles}, \bibinfo{person}{Arthur Mensch}, \bibinfo{person}{Chris Bamford}, \bibinfo{person}{Devendra~Singh Chaplot}, \bibinfo{person}{Diego de~las Casas}, \bibinfo{person}{Florian Bressand}, \bibinfo{person}{Gianna Lengyel}, \bibinfo{person}{Guillaume Lample}, \bibinfo{person}{Lucile Saulnier}, {et~al\mbox{.}}} \bibinfo{year}{2023}\natexlab{}.
\newblock \showarticletitle{Mistral 7B}.
\newblock \bibinfo{journal}{\emph{arXiv preprint arXiv:2310.06825}} (\bibinfo{year}{2023}).
\newblock


\bibitem[Jin et~al\mbox{.}(2024)]%
        {jin2024tug}
\bibfield{author}{\bibinfo{person}{Zhuoran Jin}, \bibinfo{person}{Pengfei Cao}, \bibinfo{person}{Yubo Chen}, \bibinfo{person}{Kang Liu}, \bibinfo{person}{Xiaojian Jiang}, \bibinfo{person}{Jiexin Xu}, \bibinfo{person}{Qiuxia Li}, {and} \bibinfo{person}{Jun Zhao}.} \bibinfo{year}{2024}\natexlab{}.
\newblock \showarticletitle{Tug-of-war between knowledge: Exploring and resolving knowledge conflicts in retrieval-augmented language models}.
\newblock \bibinfo{journal}{\emph{arXiv preprint arXiv:2402.14409}} (\bibinfo{year}{2024}).
\newblock


\bibitem[Kortukov et~al\mbox{.}(2024)]%
        {kortukov2024studying}
\bibfield{author}{\bibinfo{person}{Evgenii Kortukov}, \bibinfo{person}{Alexander Rubinstein}, \bibinfo{person}{Elisa Nguyen}, {and} \bibinfo{person}{Seong~Joon Oh}.} \bibinfo{year}{2024}\natexlab{}.
\newblock \showarticletitle{Studying Large Language Model Behaviors Under Realistic Knowledge Conflicts}.
\newblock \bibinfo{journal}{\emph{arXiv preprint arXiv:2404.16032}} (\bibinfo{year}{2024}).
\newblock


\bibitem[Lin(2004)]%
        {lin2004rouge}
\bibfield{author}{\bibinfo{person}{Chin-Yew Lin}.} \bibinfo{year}{2004}\natexlab{}.
\newblock \showarticletitle{ROUGE: A package for automatic evaluation of summaries}. In \bibinfo{booktitle}{\emph{Text summarization branches out}}. \bibinfo{pages}{74--81}.
\newblock


\bibitem[Liu et~al\mbox{.}(2023)]%
        {liu2023geval}
\bibfield{author}{\bibinfo{person}{Yang Liu}, \bibinfo{person}{Dan Iter}, \bibinfo{person}{Yichong Xu}, \bibinfo{person}{Shuohang Wang}, \bibinfo{person}{Ruochen Xu}, {and} \bibinfo{person}{Chenguang Zhu}.} \bibinfo{year}{2023}\natexlab{}.
\newblock \showarticletitle{{G-Eval: NLG Evaluation using GPT-4 with Better Human Alignment}}.
\newblock \bibinfo{journal}{\emph{arXiv preprint arXiv:2303.16634}} (\bibinfo{year}{2023}).
\newblock


\bibitem[Lyu et~al\mbox{.}(2024)]%
        {lyu2024crud}
\bibfield{author}{\bibinfo{person}{Yuanjie Lyu}, \bibinfo{person}{Zhiyu Li}, \bibinfo{person}{Simin Niu}, \bibinfo{person}{Feiyu Xiong}, \bibinfo{person}{Bo Tang}, \bibinfo{person}{Wenjin Wang}, \bibinfo{person}{Hao Wu}, \bibinfo{person}{Huanyong Liu}, \bibinfo{person}{Tong Xu}, {and} \bibinfo{person}{Enhong Chen}.} \bibinfo{year}{2024}\natexlab{}.
\newblock \showarticletitle{CRUD-RAG: A comprehensive chinese benchmark for retrieval-augmented generation of large language models}.
\newblock \bibinfo{journal}{\emph{arXiv preprint arXiv:2401.17043}} (\bibinfo{year}{2024}).
\newblock


\bibitem[Meta(2024)]%
        {AI2024Llama}
\bibfield{author}{\bibinfo{person}{Meta}.} \bibinfo{year}{2024}\natexlab{}.
\newblock \showarticletitle{Introducing Meta Llama 3: The most capable openly available LLM to date}.
\newblock  (\bibinfo{year}{2024}).
\newblock
\urldef\tempurl%
\url{https://ai.meta.com/blog/meta-llama-3/}
\showURL{%
\tempurl}


\bibitem[Papineni et~al\mbox{.}(2002)]%
        {papineni2002bleu}
\bibfield{author}{\bibinfo{person}{Kishore Papineni}, \bibinfo{person}{Salim Roukos}, \bibinfo{person}{Todd Ward}, {and} \bibinfo{person}{Wei-Jing Zhu}.} \bibinfo{year}{2002}\natexlab{}.
\newblock \showarticletitle{BLEU: a method for automatic evaluation of machine translation}. In \bibinfo{booktitle}{\emph{Proceedings of the 40th annual meeting of the Association for Computational Linguistics}}. \bibinfo{pages}{311--318}.
\newblock


\bibitem[Robertson et~al\mbox{.}(2009)]%
        {robertson2009probabilistic}
\bibfield{author}{\bibinfo{person}{Stephen Robertson}, \bibinfo{person}{Hugo Zaragoza}, {et~al\mbox{.}}} \bibinfo{year}{2009}\natexlab{}.
\newblock \showarticletitle{The probabilistic relevance framework: BM25 and beyond}.
\newblock \bibinfo{journal}{\emph{Foundations and Trends{\textregistered} in Information Retrieval}} \bibinfo{volume}{3}, \bibinfo{number}{4} (\bibinfo{year}{2009}), \bibinfo{pages}{333--389}.
\newblock


\bibitem[Ru et~al\mbox{.}(2024)]%
        {ru2024ragchecker}
\bibfield{author}{\bibinfo{person}{Dongyu Ru}, \bibinfo{person}{Lin Qiu}, \bibinfo{person}{Xiangkun Hu}, \bibinfo{person}{Tianhang Zhang}, \bibinfo{person}{Peng Shi}, \bibinfo{person}{Shuaichen Chang}, \bibinfo{person}{Jiayang Cheng}, \bibinfo{person}{Cunxiang Wang}, \bibinfo{person}{Shichao Sun}, \bibinfo{person}{Huanyu Li}, \bibinfo{person}{Zizhao Zhang}, \bibinfo{person}{Binjie Wang}, \bibinfo{person}{Jiarong Jiang}, \bibinfo{person}{Tong He}, \bibinfo{person}{Zhiguo Wang}, \bibinfo{person}{Pengfei Liu}, \bibinfo{person}{Yue Zhang}, {and} \bibinfo{person}{Zheng Zhang}.} \bibinfo{year}{2024}\natexlab{}.
\newblock \showarticletitle{RAGCHECKER: A Fine-grained Framework for Diagnosing Retrieval-Augmented Generation}.
\newblock \bibinfo{journal}{\emph{arXiv preprint arXiv:2408.08067}} (\bibinfo{year}{2024}).
\newblock


\bibitem[Saad-Falcon et~al\mbox{.}(2023)]%
        {saadfalcon2023ares}
\bibfield{author}{\bibinfo{person}{Jon Saad-Falcon}, \bibinfo{person}{Omar Khattab}, \bibinfo{person}{Christopher Potts}, {and} \bibinfo{person}{Matei Zaharia}.} \bibinfo{year}{2023}\natexlab{}.
\newblock \bibinfo{title}{ARES: An Automated Evaluation Framework for Retrieval-Augmented Generation Systems}.
\newblock
\showeprint[arxiv]{2311.09476}~[cs.CL]


\bibitem[Touvron et~al\mbox{.}(2023)]%
        {touvron2023Llama}
\bibfield{author}{\bibinfo{person}{Hugo Touvron}, \bibinfo{person}{Thibaut Lavril}, \bibinfo{person}{Gautier Izacard}, \bibinfo{person}{Xavier Martinet}, \bibinfo{person}{Marie-Anne Lachaux}, \bibinfo{person}{Timoth{\'e}e Lacroix}, \bibinfo{person}{Baptiste Rozi{\`e}re}, \bibinfo{person}{Naman Goyal}, \bibinfo{person}{Eric Hambro}, \bibinfo{person}{Faisal Azhar}, {et~al\mbox{.}}} \bibinfo{year}{2023}\natexlab{}.
\newblock \showarticletitle{Llama: Open and efficient foundation language models}.
\newblock \bibinfo{journal}{\emph{arXiv preprint arXiv:2302.13971}} (\bibinfo{year}{2023}).
\newblock


\bibitem[Tsatsaronis et~al\mbox{.}(2015)]%
        {tsatsaronis2015overview}
\bibfield{author}{\bibinfo{person}{George Tsatsaronis}, \bibinfo{person}{Georgios Balikas}, \bibinfo{person}{Prodromos Malakasiotis}, \bibinfo{person}{Ioannis Partalas}, \bibinfo{person}{Matthias Zschunke}, \bibinfo{person}{Michael~R Alvers}, \bibinfo{person}{Dirk Weissenborn}, \bibinfo{person}{Anastasia Krithara}, \bibinfo{person}{Sergios Petridis}, \bibinfo{person}{Dimitris Polychronopoulos}, {et~al\mbox{.}}} \bibinfo{year}{2015}\natexlab{}.
\newblock \showarticletitle{An overview of the BIOASQ large-scale biomedical semantic indexing and question answering competition}.
\newblock \bibinfo{journal}{\emph{BMC bioinformatics}}  \bibinfo{volume}{16} (\bibinfo{year}{2015}), \bibinfo{pages}{1--28}.
\newblock


\bibitem[Wang et~al\mbox{.}(2023)]%
        {wang2023improving}
\bibfield{author}{\bibinfo{person}{Liang Wang}, \bibinfo{person}{Nan Yang}, \bibinfo{person}{Xiaolong Huang}, \bibinfo{person}{Linjun Yang}, \bibinfo{person}{Rangan Majumder}, {and} \bibinfo{person}{Furu Wei}.} \bibinfo{year}{2023}\natexlab{}.
\newblock \showarticletitle{Improving text embeddings with large language models}.
\newblock \bibinfo{journal}{\emph{arXiv preprint arXiv:2401.00368}} (\bibinfo{year}{2023}).
\newblock


\bibitem[Zhang et~al\mbox{.}(2019)]%
        {zhang2019bertscore}
\bibfield{author}{\bibinfo{person}{Tianyi Zhang}, \bibinfo{person}{Varsha Kishore}, \bibinfo{person}{Felix Wu}, \bibinfo{person}{Kilian~Q Weinberger}, {and} \bibinfo{person}{Yoav Artzi}.} \bibinfo{year}{2019}\natexlab{}.
\newblock \showarticletitle{BERTScore: Evaluating text generation with BERT}. In \bibinfo{booktitle}{\emph{International Conference on Learning Representations}}.
\newblock


\bibitem[Zheng et~al\mbox{.}(2023)]%
        {zheng2023judging}
\bibfield{author}{\bibinfo{person}{Lianmin Zheng}, \bibinfo{person}{Wei-Lin Chiang}, \bibinfo{person}{Ying Sheng}, \bibinfo{person}{Siyuan Zhuang}, \bibinfo{person}{Zhanghao Wu}, \bibinfo{person}{Yonghao Zhuang}, \bibinfo{person}{Zi Lin}, \bibinfo{person}{Zhuohan Li}, \bibinfo{person}{Dacheng Li}, \bibinfo{person}{Eric Xing}, {et~al\mbox{.}}} \bibinfo{year}{2023}\natexlab{}.
\newblock \showarticletitle{Judging llm-as-a-judge with mt-bench and chatbot arena}.
\newblock \bibinfo{journal}{\emph{Advances in Neural Information Processing Systems}}  \bibinfo{volume}{36} (\bibinfo{year}{2023}), \bibinfo{pages}{46595--46623}.
\newblock


\bibitem[Ziems et~al\mbox{.}(2024)]%
        {ziems2024can}
\bibfield{author}{\bibinfo{person}{Caleb Ziems}, \bibinfo{person}{William Held}, \bibinfo{person}{Omar Shaikh}, \bibinfo{person}{Jiaao Chen}, \bibinfo{person}{Zhehao Zhang}, {and} \bibinfo{person}{Diyi Yang}.} \bibinfo{year}{2024}\natexlab{}.
\newblock \bibinfo{title}{Can Large Language Models Transform Computational Social Science?}
\newblock
\showeprint[arxiv]{2305.03514}~[cs.CL]


\end{thebibliography}

\appendix

\section{BioASQ Dataset Examples}
\label{app:dataset_examples}

Examples illustrating the structure of the BioASQ dataset used in our experiments.

\begin{lstlisting}[language=json, caption={Example BioASQ Query-Answer Pairs}, label={app:query_examples}]
[
  {
    "query_id": 0,
    "text": "Is Hirschsprung disease a mendelian or a multifactorial disorder?",
    "metadata": {},
    "gt_answer": "Coding sequence mutations in RET, GDNF, EDNRB, EDN3, and SOX10 are involved in the development of Hirschsprung disease. The majority of these genes was shown to be related to Mendelian syndromic forms of Hirschsprung's disease, whereas the non-Mendelian inheritance of sporadic non-syndromic Hirschsprung disease proved to be complex;"
  },
  {
    "query_id": 1,
    "text": "List signaling molecules (ligands) that interact with the receptor EGFR?",
    "metadata": {},
    "gt_answer": "The 7 known EGFR ligands are: epidermal growth factor (EGF), betacellulin (BTC), epiregulin (EPR), heparin-binding EGF (HB-EGF), transforming growth factor-\alpha [TGF-\alpha], amphiregulin (AREG) and epigen (EPG)."
  }
]
\end{lstlisting}

\begin{lstlisting}[language=json, caption={Example BioASQ Supporting Document Snippets}, label={app:document_examples}]
[
  {
    "title": "",
    "text": "INTRODUCTION: The majority of patients with type 1 diabetes mellitus (T1DM) do not achieve glycemic targets. In addition, treatment with insulin is associated with increased risk for hypoglycemia and weight gain. Accordingly, there is an unmet need for new safe and effective glucose-lowering agents in this population. Sotagliflozin, a dual inhibitor of sodium-glucose co-transporters 1 and 2, has been recently approved for use in patients with T1DM.",
    "metadata": {},
    "doc_id": "33108240"
  },
  {
    "title": "",
    "text": "The World Health Organization is still revising the epidemiology of multi-system inflammatory syndrome in children (MIS-C) and the preliminary case definition, although there is a dearth of robust evidence regarding the clinical presentations, severity, and outcomes. Researchers, epidemiologists, and clinicians are struggling to characterize and describe the disease phenomenon while taking care of the diseased persons at the forefronts.",
    "metadata": {},
    "doc_id": "33110725"
  }
]
\end{lstlisting}

\section{CCRS Evaluation Prompts}
\label{app:prompts}
The specific prompts used to instruct the \texttt{Llama-70B-Instruct} judge model for each CCRS metric calculation.

\begin{promptbox}[Contextual Coherence (CC) Evaluation Prompt] \label{app:prompt_cc}
\small
Evaluate the Contextual Coherence between the generated response and the retrieved context.

\textbf{Retrieved Context (C):} \texttt{\{context\}}

\textbf{Generated Response (r):} \texttt{\{response\}}

\textbf{Task:} Assess the logical consistency and coherence of the response with respect to the provided context. Ensure the response logically follows from and does not contradict the context.

\textbf{Scoring:} Score from 0 to 100, where 0 is completely incoherent or contradictory, and 100 is perfectly coherent and consistent. If no response is generated, the score is 0.

\textbf{Output:} Only print the score and nothing else.
\end{promptbox}

\begin{promptbox}[Question Relevance (QR) Evaluation Prompt] \label{app:prompt_qr}
\small
Evaluate the Question Relevance of the generated response to the user query.

\textbf{User Query (q):} \texttt{\{question\}}

\textbf{Generated Response (r):} \texttt{\{response\}}

\textbf{Task:} Assess how well the response directly addresses the user's query. Consider if the response answers the question asked.

\textbf{Scoring:} Score from 0 to 100, where 0 is completely irrelevant and 100 is perfectly relevant and directly answers the query. If no response is generated, the score is 0.

\textbf{Output:} Only print the score and nothing else.
\end{promptbox}

\begin{promptbox}[Information Density (ID) Evaluation Prompt] \label{app:prompt_id}
\small
Evaluate the Information Density of the generated response.

\textbf{User Query (q):} \texttt{\{question\}}

\textbf{Retrieved Context (C):} \texttt{\{context\}}

\textbf{Generated Response (r):} \texttt{\{response\}}

\textbf{Task:} Assess the balance of conciseness and informativeness in the response, considering both the context and the query. The response should provide necessary information without being overly verbose or unnecessarily brief.

\textbf{Scoring:} Score from 0 to 100, where 0 is either too verbose (contains excessive irrelevant detail) or uninformative (lacks necessary information), and 100 is optimally concise and informative for the query. If no response is generated, the score is 0.

\textbf{Output:} Only print the score and nothing else.
\end{promptbox}

\begin{promptbox}[Answer Correctness (AC) Evaluation Prompt (LLM part)] \label{app:prompt_ac}
\small
Evaluate the Answer Correctness of the generated response.

\textbf{Retrieved Context (C):} \texttt{\{context\}}

\textbf{Generated Response (r):} \texttt{\{response\}}

\textbf{Ground Truth Answer (g):} \texttt{\{ground\_truth\}}

\textbf{Task:} Assess the factual accuracy of the information presented in the response compared to the ground truth answer, considering the provided context. Do not penalize differences in phrasing if the core factual meaning is preserved and accurate according to the ground truth.

\textbf{Scoring:} Score from 0 to 100, where 0 is completely incorrect or contains significant factual errors, and 100 represents perfect factual accuracy (semantically equivalent to the ground truth). If no response is generated, the score is 0.

\textbf{Output:} Only print the score and nothing else.
\end{promptbox}
Note: The final AC score combines this LLM score (weighted 0.3) with an Exact Match check (weighted 0.7).

\begin{promptbox}[Information Recall (IR) Evaluation Prompt] \label{app:prompt_ir}
\small
Evaluate the Information Recall of the generated response.

\textbf{Retrieved Context (C):} \texttt{\{context\}}

\textbf{Generated Response (r):} \texttt{\{response\}}

\textbf{Ground Truth Answer (g):} \texttt{\{ground\_truth\}}

\textbf{Task:} Assess how much of the essential information present in the ground truth answer is captured in the generated response, considering the provided context. Focus on whether key facts or points from the ground truth are included.

\textbf{Scoring:} Score from 0 to 100, where 0 means no essential information from the ground truth is recalled, and 100 means all essential information is fully captured. If no response is generated, the score is 0.

\textbf{Output:} Only print the score and nothing else.
\end{promptbox}

\section{Detailed Metric Property Data}
\label{app:metric_data_details}

\subsection{Observed Metric Bounds}
\label{app:bounds}
Tables showing the frequency and percentage of scores reaching the minimum (0) and maximum (1) bounds for each CCRS metric across the 6 systems (N=4,719).

\begin{table}[H] \centering \small \caption{Observed Bounds (Count/Percentage) of CCRS Components - Part 1 (N=4719)} \label{table:observed_bounds_part1_app}
\begin{tabular}{|c|c|c|c|c|} \hline \textbf{System} & \textbf{Bound} & \textbf{CC} & \textbf{QR} & \textbf{ID} \\ \hline
\multirow{2}{*}{A: M+BM25} & 0 & 186 / 3.9\% & 61 / 1.3\% & 14 / 0.3\% \\  & 1 & 779 / 16.5\% & 3698 / 78.4\% & 188 / 4.0\% \\ \hline
\multirow{2}{*}{C: L8+BM25} & 0 & 1070 / 22.7\% & 229 / 4.8\% & 177 / 3.7\% \\ & 1 & 630 / 13.3\% & 3064 / 64.9\% & 136 / 2.9\% \\ \hline
\multirow{2}{*}{E: L3.2+BM25} & 0 & 892 / 18.9\% & 151 / 3.2\% & 29 / 0.6\% \\ & 1 & 368 / 7.8\% & 3104 / 65.8\% & 182 / 3.9\% \\ \hline
\multirow{2}{*}{B: M+E5} & 0 & 186 / 3.9\% & 43 / 0.9\% & 16 / 0.3\% \\ & 1 & 724 / 15.3\% & 4031 / 85.4\% & 132 / 2.8\% \\ \hline
\multirow{2}{*}{D: L8+E5} & 0 & 995 / 21.1\% & 139 / 2.9\% & 82 / 1.7\% \\ & 1 & 618 / 13.1\% & 3361 / 71.2\% & 132 / 2.8\% \\ \hline
\multirow{2}{*}{F: L3.2+E5} & 0 & 819 / 17.4\% & 100 / 2.1\% & 17 / 0.4\% \\ & 1 & 419 / 8.9\% & 3381 / 71.6\% & 147 / 3.1\% \\ \hline
\end{tabular} \end{table}

\begin{table}[H] \centering \small \caption{Observed Bounds (Count/Percentage) of CCRS Components - Part 2 (N=4719)} \label{table:observed_bounds_part2_app}
\begin{tabular}{|c|c|c|c|} \hline \textbf{System} & \textbf{Bound} & \textbf{AC} & \textbf{IR} \\ \hline
\multirow{2}{*}{A: M+BM25} & 0 & 377 / 8.0\% & 610 / 12.9\% \\ & 1 & 0 / 0.0\% & 254 / 5.4\% \\ \hline
\multirow{2}{*}{C: L8+BM25} & 0 & 614 / 13.0\% & 1015 / 21.5\% \\ & 1 & 0 / 0.0\% & 146 / 3.1\% \\ \hline
\multirow{2}{*}{E: L3.2+BM25} & 0 & 592 / 12.5\% & 1007 / 21.3\% \\ & 1 & 0 / 0.0\% & 102 / 2.2\% \\ \hline
\multirow{2}{*}{B: M+E5} & 0 & 321 / 6.8\% & 483 / 10.2\% \\ & 1 & 0 / 0.0\% & 228 / 4.8\% \\ \hline
\multirow{2}{*}{D: L8+E5} & 0 & 530 / 11.2\% & 887 / 18.8\% \\ & 1 & 0 / 0.0\% & 138 / 2.9\% \\ \hline
\multirow{2}{*}{F: L3.2+E5} & 0 & 562 / 11.9\% & 895 / 19.0\% \\ & 1 & 0 / 0.0\% & 112 / 2.4\% \\ \hline
\end{tabular} \end{table}

\subsection{Detailed CCRS Validity Analysis (Per System Correlations)}
\label{app:validity_details}
Correlation matrices (Pearson, Spearman, Kendall) for each of the six RAG systems, illustrating system-specific relationships between CCRS metrics.

\begin{table}[H] \centering \caption{Correlations for System A (Mistral+BM25)} \label{tab:bm25_mistral_corr_app} \small
\begin{tabular}{@{}lrrrrr@{}} \toprule & CC & QR & ID & AC & IR \\ \midrule \multicolumn{6}{l}{\textbf{Pearson}} \\
CC & 1.000 & 0.211 & 0.280 & 0.218 & 0.343 \\ QR & 0.211 & 1.000 & 0.451 & 0.469 & 0.496 \\ ID & 0.280 & 0.451 & 1.000 & 0.299 & 0.299 \\ AC & 0.218 & 0.469 & 0.299 & 1.000 & 0.822 \\ IR & 0.343 & 0.496 & 0.299 & 0.822 & 1.000 \\ \midrule \multicolumn{6}{l}{\textbf{Spearman}} \\
CC & 1.000 & 0.284 & 0.324 & 0.300 & 0.362 \\ QR & 0.284 & 1.000 & 0.458 & 0.421 & 0.495 \\ ID & 0.324 & 0.458 & 1.000 & 0.370 & 0.368 \\ AC & 0.300 & 0.421 & 0.370 & 1.000 & 0.809 \\ IR & 0.362 & 0.495 & 0.368 & 0.809 & 1.000 \\ \midrule \multicolumn{6}{l}{\textbf{Kendall}} \\
CC & 1.000 & 0.255 & 0.297 & 0.255 & 0.303 \\ QR & 0.255 & 1.000 & 0.416 & 0.362 & 0.425 \\ ID & 0.297 & 0.416 & 1.000 & 0.312 & 0.310 \\ AC & 0.255 & 0.362 & 0.312 & 1.000 & 0.736 \\ IR & 0.303 & 0.425 & 0.310 & 0.736 & 1.000 \\ \bottomrule \end{tabular} \end{table}

\begin{table}[H] \centering \caption{Correlations for System B (Mistral+E5)} \label{tab:e5_mistral_corr_app} \small
\begin{tabular}{@{}lrrrrr@{}} \toprule & CC & QR & ID & AC & IR \\ \midrule \multicolumn{6}{l}{\textbf{Pearson}} \\
CC & 1.000 & 0.205 & 0.321 & 0.226 & 0.377 \\ QR & 0.205 & 1.000 & 0.437 & 0.413 & 0.411 \\ ID & 0.321 & 0.437 & 1.000 & 0.253 & 0.259 \\ AC & 0.226 & 0.413 & 0.253 & 1.000 & 0.814 \\ IR & 0.377 & 0.411 & 0.259 & 0.814 & 1.000 \\ \midrule \multicolumn{6}{l}{\textbf{Spearman}} \\
CC & 1.000 & 0.294 & 0.324 & 0.324 & 0.402 \\ QR & 0.294 & 1.000 & 0.391 & 0.354 & 0.403 \\ ID & 0.324 & 0.391 & 1.000 & 0.281 & 0.292 \\ AC & 0.324 & 0.354 & 0.281 & 1.000 & 0.785 \\ IR & 0.402 & 0.403 & 0.292 & 0.785 & 1.000 \\ \midrule \multicolumn{6}{l}{\textbf{Kendall}} \\
CC & 1.000 & 0.269 & 0.299 & 0.278 & 0.341 \\ QR & 0.269 & 1.000 & 0.359 & 0.305 & 0.350 \\ ID & 0.299 & 0.359 & 1.000 & 0.238 & 0.247 \\ AC & 0.278 & 0.305 & 0.238 & 1.000 & 0.708 \\ IR & 0.341 & 0.350 & 0.247 & 0.708 & 1.000 \\ \bottomrule \end{tabular} \end{table}

\begin{table}[H] \centering \caption{Correlations for System C (Llama8B+BM25)} \label{tab:bm25_Llama3_corr_app} \small
\begin{tabular}{@{}lrrrrr@{}} \toprule & CC & QR & ID & AC & IR \\ \midrule \multicolumn{6}{l}{\textbf{Pearson}} \\
CC & 1.000 & 0.365 & 0.514 & 0.237 & 0.528 \\ QR & 0.365 & 1.000 & 0.687 & 0.479 & 0.492 \\ ID & 0.514 & 0.687 & 1.000 & 0.420 & 0.444 \\ AC & 0.237 & 0.479 & 0.420 & 1.000 & 0.734 \\ IR & 0.528 & 0.492 & 0.444 & 0.734 & 1.000 \\ \midrule \multicolumn{6}{l}{\textbf{Spearman}} \\
CC & 1.000 & 0.354 & 0.402 & 0.234 & 0.514 \\ QR & 0.354 & 1.000 & 0.611 & 0.362 & 0.527 \\ ID & 0.402 & 0.611 & 1.000 & 0.339 & 0.466 \\ AC & 0.234 & 0.362 & 0.339 & 1.000 & 0.716 \\ IR & 0.514 & 0.527 & 0.466 & 0.716 & 1.000 \\ \midrule \multicolumn{6}{l}{\textbf{Kendall}} \\
CC & 1.000 & 0.301 & 0.328 & 0.203 & 0.425 \\ QR & 0.301 & 1.000 & 0.538 & 0.304 & 0.445 \\ ID & 0.328 & 0.538 & 1.000 & 0.275 & 0.374 \\ AC & 0.203 & 0.304 & 0.275 & 1.000 & 0.642 \\ IR & 0.425 & 0.445 & 0.374 & 0.642 & 1.000 \\ \bottomrule \end{tabular} \end{table}

\begin{table}[H] \centering \caption{Correlations for System D (Llama8B+E5)} \label{tab:e5_Llama3_corr_app} \small
\begin{tabular}{@{}lrrrrr@{}} \toprule & CC & QR & ID & AC & IR \\ \midrule \multicolumn{6}{l}{\textbf{Pearson}} \\
CC & 1.000 & 0.334 & 0.478 & 0.195 & 0.511 \\ QR & 0.334 & 1.000 & 0.585 & 0.363 & 0.436 \\ ID & 0.478 & 0.585 & 1.000 & 0.281 & 0.362 \\ AC & 0.195 & 0.363 & 0.281 & 1.000 & 0.713 \\ IR & 0.511 & 0.436 & 0.362 & 0.713 & 1.000 \\ \midrule \multicolumn{6}{l}{\textbf{Spearman}} \\
CC & 1.000 & 0.387 & 0.367 & 0.215 & 0.511 \\ QR & 0.387 & 1.000 & 0.552 & 0.223 & 0.459 \\ ID & 0.367 & 0.552 & 1.000 & 0.207 & 0.363 \\ AC & 0.215 & 0.223 & 0.207 & 1.000 & 0.684 \\ IR & 0.511 & 0.459 & 0.363 & 0.684 & 1.000 \\ \midrule \multicolumn{6}{l}{\textbf{Kendall}} \\
CC & 1.000 & 0.335 & 0.298 & 0.189 & 0.423 \\ QR & 0.335 & 1.000 & 0.488 & 0.189 & 0.390 \\ ID & 0.298 & 0.488 & 1.000 & 0.167 & 0.290 \\ AC & 0.189 & 0.189 & 0.167 & 1.000 & 0.617 \\ IR & 0.423 & 0.390 & 0.290 & 0.617 & 1.000 \\ \bottomrule \end{tabular} \end{table}

\begin{table}[H] \centering \caption{Correlations for System E (Llama3.2+BM25)} \label{tab:bm25_Llama32_corr_app} \small
\begin{tabular}{@{}lrrrrr@{}} \toprule & CC & QR & ID & AC & IR \\ \midrule \multicolumn{6}{l}{\textbf{Pearson}} \\
CC & 1.000 & 0.225 & 0.356 & 0.159 & 0.462 \\ QR & 0.225 & 1.000 & 0.361 & 0.315 & 0.421 \\ ID & 0.356 & 0.361 & 1.000 & 0.315 & 0.337 \\ AC & 0.159 & 0.315 & 0.315 & 1.000 & 0.701 \\ IR & 0.462 & 0.421 & 0.337 & 0.701 & 1.000 \\ \midrule \multicolumn{6}{l}{\textbf{Spearman}} \\
CC & 1.000 & 0.279 & 0.366 & 0.221 & 0.490 \\ QR & 0.279 & 1.000 & 0.474 & 0.267 & 0.464 \\ ID & 0.366 & 0.474 & 1.000 & 0.307 & 0.387 \\ AC & 0.221 & 0.267 & 0.307 & 1.000 & 0.672 \\ IR & 0.490 & 0.464 & 0.387 & 0.672 & 1.000 \\ \midrule \multicolumn{6}{l}{\textbf{Kendall}} \\
CC & 1.000 & 0.238 & 0.300 & 0.195 & 0.406 \\ QR & 0.238 & 1.000 & 0.412 & 0.227 & 0.392 \\ ID & 0.300 & 0.412 & 1.000 & 0.247 & 0.307 \\ AC & 0.195 & 0.227 & 0.247 & 1.000 & 0.612 \\ IR & 0.406 & 0.392 & 0.307 & 0.612 & 1.000 \\ \bottomrule \end{tabular} \end{table}

\begin{table}[H] \centering \caption{Correlations for System F (Llama3.2+E5)} \label{tab:e5_Llama32_corr_app} \small
\begin{tabular}{@{}lrrrrr@{}} \toprule & CC & QR & ID & AC & IR \\ \midrule \multicolumn{6}{l}{\textbf{Pearson}} \\
CC & 1.000 & 0.220 & 0.354 & 0.170 & 0.477 \\ QR & 0.220 & 1.000 & 0.382 & 0.275 & 0.367 \\ ID & 0.354 & 0.382 & 1.000 & 0.245 & 0.315 \\ AC & 0.170 & 0.275 & 0.245 & 1.000 & 0.726 \\ IR & 0.477 & 0.367 & 0.315 & 0.726 & 1.000 \\ \midrule \multicolumn{6}{l}{\textbf{Spearman}} \\
CC & 1.000 & 0.340 & 0.353 & 0.252 & 0.526 \\ QR & 0.340 & 1.000 & 0.502 & 0.227 & 0.420 \\ ID & 0.353 & 0.502 & 1.000 & 0.226 & 0.344 \\ AC & 0.252 & 0.227 & 0.226 & 1.000 & 0.701 \\ IR & 0.526 & 0.420 & 0.344 & 0.701 & 1.000 \\ \midrule \multicolumn{6}{l}{\textbf{Kendall}} \\
CC & 1.000 & 0.296 & 0.289 & 0.221 & 0.433 \\ QR & 0.296 & 1.000 & 0.438 & 0.193 & 0.357 \\ ID & 0.289 & 0.438 & 1.000 & 0.179 & 0.270 \\ AC & 0.221 & 0.193 & 0.179 & 1.000 & 0.630 \\ IR & 0.433 & 0.357 & 0.270 & 0.630 & 1.000 \\ \bottomrule \end{tabular} \end{table}

\subsection{Empirical Tie Probabilities}
\label{app:ties}
The probability that two randomly selected responses receive the same score for a given metric, calculated empirically for each system.

\begin{table}[H] \small \centering \caption{Empirical Tie Probabilities (\%) for CCRS Metrics Across Systems} \label{tab:empirical-tie-probs_app}
\begin{tabular}{@{}lcccccc@{}}
\toprule Metric & A & B & C & D & E & F \\ \midrule
CC & 33.9 & 40.7 & 22.6 & 24.0 & 23.0 & 24.1 \\
QR & 62.5 & 73.5 & 45.9 & 53.7 & 46.7 & 54.2 \\
ID & 31.1 & 33.1 & 21.1 & 22.1 & 21.4 & 21.5 \\
AC & 16.3 & 16.2 & 16.6 & 16.5 & 16.0 & 15.6 \\
IR & 17.0 & 18.2 & 15.4 & 14.9 & 15.4 & 14.8 \\ \bottomrule
\end{tabular} \end{table}

\subsection{Population Level Statistics}
\label{app:pop_stats}
Detailed descriptive statistics (central tendency, variability, shape) for each CCRS metric across the six systems.

\begin{table}[H] \small \centering \caption{Central Tendency Measures Across Systems and Metrics (\%)} \label{fig:central_tendency_app}
\begin{tabular}{|l|l|r|r|r|r|} \hline \textbf{System} & \textbf{Metric} & \textbf{Mean} & \textbf{GeoMean*} & \textbf{Median} & \textbf{Midhinge} \\ \hline
A: M+BM25 & AC & 21.10 & 0.00 & 24.00 & 23.25 \\
 & CC & 87.95 & 0.00 & 95.00 & 92.50 \\ & ID & 86.81 & 0.00 & 85.00 & 90.00 \\ & IR & 54.96 & 0.00 & 60.00 & 56.50 \\ & QR & 92.83 & 0.00 & 100.00 & 100.00 \\ \hline
B: M+E5 & AC & 21.81 & 0.00 & 24.00 & 23.25 \\
 & CC & 88.61 & 0.00 & 95.00 & 95.00 \\ & ID & 86.90 & 0.00 & 85.00 & 90.00 \\ & IR & 58.28 & 0.00 & 60.00 & 60.00 \\ & QR & 95.40 & 0.00 & 100.00 & 100.00 \\ \hline
C: L8+BM25 & AC & 18.54 & 0.00 & 24.00 & 18.00 \\
 & CC & 67.89 & 0.00 & 85.00 & 67.50 \\ & ID & 79.09 & 0.00 & 85.00 & 87.50 \\ & IR & 43.00 & 0.00 & 40.00 & 43.33 \\ & QR & 87.07 & 0.00 & 100.00 & 90.00 \\ \hline
D: L8+E5 & AC & 19.24 & 0.00 & 24.00 & 18.75 \\
 & CC & 69.29 & 0.00 & 95.00 & 77.50 \\ & ID & 80.41 & 0.00 & 85.00 & 87.50 \\ & IR & 45.44 & 0.00 & 50.00 & 50.00 \\ & QR & 90.55 & 0.00 & 100.00 & 95.00 \\ \hline
E: L3.2+BM25 & AC & 18.41 & 0.00 & 24.00 & 18.00 \\
 & CC & 68.88 & 0.00 & 80.00 & 77.50 \\ & ID & 80.23 & 0.00 & 85.00 & 87.50 \\ & IR & 42.04 & 0.00 & 40.00 & 40.00 \\ & QR & 88.28 & 0.00 & 100.00 & 90.00 \\ \hline
F: L3.2+E5 & AC & 18.76 & 0.00 & 24.00 & 18.75 \\
 & CC & 70.88 & 0.00 & 85.00 & 87.50 \\ & ID & 80.43 & 0.00 & 85.00 & 87.50 \\ & IR & 44.79 & 0.00 & 40.00 & 50.00 \\ & QR & 90.83 & 0.00 & 100.00 & 95.00 \\ \hline
\multicolumn{6}{l}{\footnotesize *Geometric Mean is 0 due to presence of 0 scores.}
\end{tabular} \end{table}

\begin{table}[H] \small \centering \caption{Variability Measures Across Systems and Metrics} \label{fig:variability_app}
\begin{tabular}{|l|l|r|r|r|r|r|} \hline \textbf{System} & \textbf{Metric} & \textbf{Var} & \textbf{Min} & \textbf{Max} & \textbf{Range} & \textbf{IQR} \\ \hline
A: M+BM25 & AC & 75.72 & 0.0 & 30.7 & 30.7 & 10.5 \\
 & CC & 424.25 & 0.0 & 100.0 & 100.0 & 5.0 \\ & ID & 123.58 & 0.0 & 100.0 & 100.0 & 10.0 \\ & IR & 918.67 & 0.0 & 100.0 & 100.0 & 47.0 \\ & QR & 340.18 & 0.0 & 100.0 & 100.0 & 0.0 \\ \hline
B: M+E5 & AC & 69.38 & 0.0 & 30.7 & 30.7 & 10.5 \\
 & CC & 423.80 & 0.0 & 100.0 & 100.0 & 0.0 \\ & ID & 123.40 & 0.0 & 100.0 & 100.0 & 10.0 \\ & IR & 834.55 & 0.0 & 100.0 & 100.0 & 40.0 \\ & QR & 227.31 & 0.0 & 100.0 & 100.0 & 0.0 \\ \hline
C: L8+BM25 & AC & 93.68 & 0.0 & 30.7 & 30.7 & 12.0 \\
 & CC & 1527.75 & 0.0 & 100.0 & 100.0 & 55.0 \\ & ID & 502.50 & 0.0 & 100.0 & 100.0 & 15.0 \\ & IR & 973.87 & 0.0 & 100.0 & 100.0 & 46.7 \\ & QR & 633.71 & 0.0 & 100.0 & 100.0 & 20.0 \\ \hline
D: L8+E5 & AC & 89.46 & 0.0 & 30.7 & 30.7 & 13.5 \\
 & CC & 1494.39 & 0.0 & 100.0 & 100.0 & 35.0 \\ & ID & 392.57 & 0.0 & 100.0 & 100.0 & 15.0 \\ & IR & 970.01 & 0.0 & 100.0 & 100.0 & 60.0 \\ & QR & 440.64 & 0.0 & 100.0 & 100.0 & 10.0 \\ \hline
E: L3.2+BM25 & AC & 93.49 & 0.0 & 30.7 & 30.7 & 12.0 \\
 & CC & 1330.38 & 0.0 & 100.0 & 100.0 & 35.0 \\ & ID & 335.04 & 0.0 & 100.0 & 100.0 & 15.0 \\ & IR & 944.44 & 0.0 & 100.0 & 100.0 & 40.0 \\ & QR & 537.50 & 0.0 & 100.0 & 100.0 & 20.0 \\ \hline
F: L3.2+E5 & AC & 92.92 & 0.0 & 30.7 & 30.7 & 13.5 \\
 & CC & 1298.15 & 0.0 & 100.0 & 100.0 & 15.0 \\ & ID & 321.37 & 0.0 & 100.0 & 100.0 & 15.0 \\ & IR & 952.22 & 0.0 & 100.0 & 100.0 & 60.0 \\ & QR & 408.58 & 0.0 & 100.0 & 100.0 & 10.0 \\ \hline
\end{tabular} \end{table}

\begin{table}[H] \small \centering \caption{Shape Measures Across Systems and Metrics} \label{fig:shapemeasures_app}
\begin{tabular}{|l|l|r|r|} \hline \textbf{System} & \textbf{Metric} & \textbf{Skewness} & \textbf{Kurtosis} \\ \hline
A: M+BM25 & AC & -1.211 & 0.486 \\ & CC & -3.425 & 11.379 \\ & ID & -3.686 & 20.748 \\ & IR & -0.503 & -0.882 \\ & QR & -3.278 & 11.045 \\ \hline
B: M+E5 & AC & -1.387 & 1.047 \\ & CC & -3.537 & 11.916 \\ & ID & -3.949 & 22.791 \\ & IR & -0.695 & -0.565 \\ & QR & -4.363 & 20.605 \\ \hline
C: L8+BM25 & AC & -0.766 & -0.672 \\ & CC & -1.030 & -0.744 \\ & ID & -2.221 & 4.479 \\ & IR & 0.020 & -1.260 \\ & QR & -2.412 & 5.112 \\ \hline
D: L8+E5 & AC & -0.872 & -0.460 \\ & CC & -1.092 & -0.615 \\ & ID & -2.262 & 5.052 \\ & IR & -0.078 & -1.271 \\ & QR & -3.009 & 9.181 \\ \hline
E: L3.2+BM25 & AC & -0.719 & -0.738 \\ & CC & -1.165 & -0.358 \\ & ID & -2.071 & 4.284 \\ & IR & 0.050 & -1.262 \\ & QR & -2.527 & 5.965 \\ \hline
F: L3.2+E5 & AC & -0.783 & -0.674 \\ & CC & -1.256 & -0.140 \\ & ID & -2.056 & 4.160 \\ & IR & -0.071 & -1.281 \\ & QR & -2.968 & 9.049 \\ \hline \end{tabular} \end{table}

\section{RAGChecker Comparison Details}
\label{app:ragchecker_details}
Supporting data for the comparison between CCRS and RAGChecker (Precision, Recall, Faithfulness).

\begin{table}[H] \centering \small \caption{Observed Bounds (Count/Percentage) for RAGChecker Metrics (N=4719)} \label{table:observed_bounds_ragchecker_app}
\begin{tabular}{|c|c|c|c|c|} \hline \textbf{System} & \textbf{Bound} & \textbf{Precision} & \textbf{Recall} & \textbf{Faithfulness} \\ \hline
\multirow{2}{*}{A: M+BM25} & 0 & 568 / 12.0\% & 767 / 16.2\% & 209 / 4.4\% \\ & 1 & 1471 / 31.2\% & 1003 / 21.2\% & 3158 / 66.9\% \\ \hline
\multirow{2}{*}{B: M+E5} & 0 & 500 / 10.6\% & 587 / 12.4\% & 128 / 2.7\% \\ & 1 & 1475 / 31.3\% & 1075 / 22.8\% & 3469 / 73.5\% \\ \hline
\multirow{2}{*}{C: L8+BM25} & 0 & 971 / 20.6\% & 783 / 16.6\% & 564 / 11.9\% \\ & 1 & 1710 / 36.2\% & 1037 / 22.0\% & 3175 / 67.3\% \\ \hline
\multirow{2}{*}{D: L8+E5} & 0 & 839 / 17.8\% & 640 / 13.6\% & 454 / 9.6\% \\ & 1 & 1788 / 37.9\% & 1062 / 22.5\% & 3386 / 71.8\% \\ \hline
\multirow{2}{*}{E: L3.2+BM25} & 0 & 947 / 20.1\% & 860 / 18.2\% & 505 / 10.7\% \\ & 1 & 1788 / 37.9\% & 960 / 20.3\% & 3062 / 64.9\% \\ \hline
\multirow{2}{*}{F: L3.2+E5} & 0 & 857 / 18.2\% & 689 / 14.6\% & 391 / 8.3\% \\ & 1 & 1773 / 37.6\% & 1018 / 21.6\% & 3353 / 71.1\% \\ \hline \end{tabular} \end{table}

\begin{table}[H] \small \centering \caption{Empirical Tie Probabilities (\%) for RAGChecker Metrics} \label{tab:empirical-tie-probs-all_app}
\begin{tabular}{@{}lccc@{}} \toprule System & Precision & Recall & Faithfulness \\ \midrule
A: M+BM25 & 13.0 & 9.3 & 45.5 \\ B: M+E5 & 12.6 & 8.9 & 54.4 \\ C: L8+BM25 & 18.5 & 9.6 & 47.0 \\ D: L8+E5 & 18.7 & 9.1 & 52.6 \\ E: L3.2+BM25 & 19.5 & 9.4 & 43.6 \\ F: L3.2+E5 & 18.5 & 8.9 & 51.4 \\ \bottomrule \end{tabular} \end{table}

\begin{table}[H] \small \centering \caption{Population Statistics for RAGChecker Metrics (F/P/R, Scores 0-100)} \label{tab:pop_stats_ragchecker_app}
\begin{tabular}{|l|l|r|r|r|r|r|} \hline \textbf{System} & \textbf{M} & \textbf{Mean} & \textbf{Median} & \textbf{Var} & \textbf{Skew} & \textbf{Kurt} \\ \hline
A: M+BM25 & F & 87.3 & 100.0 & 6.1 & -2.34 & 4.93 \\ & P & 57.2 & 55.6 & 12.9 & -0.15 & -1.35 \\ & R & 49.9 & 50.0 & 12.3 & 0.08 & -1.28 \\ \hline
B: M+E5 & F & 91.3 & 100.0 & 4.2 & -3.08 & 9.74 \\ & P & 58.0 & 57.1 & 12.5 & -0.17 & -1.32 \\ & R & 53.7 & 50.0 & 11.6 & -0.05 & -1.22 \\ \hline
C: L8+BM25 & F & 82.5 & 100.0 & 11.0 & -1.82 & 1.76 \\ & P & 55.8 & 53.3 & 15.9 & -0.17 & -1.54 \\ & R & 50.3 & 50.0 & 12.6 & 0.06 & -1.31 \\ \hline
D: L8+E5 & F & 85.8 & 100.0 & 9.2 & -2.20 & 3.32 \\ & P & 58.2 & 60.0 & 15.3 & -0.26 & -1.48 \\ & R & 53.2 & 50.0 & 11.9 & -0.05 & -1.25 \\ \hline
E: L3.2+BM25 & F & 81.9 & 100.0 & 10.5 & -1.76 & 1.64 \\ & P & 57.2 & 60.0 & 15.8 & -0.22 & -1.52 \\ & R & 48.0 & 46.7 & 12.7 & 0.14 & -1.30 \\ \hline
F: L3.2+E5 & F & 86.3 & 100.0 & 8.5 & -2.25 & 3.70 \\ & P & 58.0 & 60.0 & 15.2 & -0.25 & -1.47 \\ & R & 51.2 & 50.0 & 12.1 & 0.04 & -1.27 \\ \hline \end{tabular} \end{table}

\section{Detailed Statistical Test Results}
\label{app:stats_details}
Complete pairwise comparison results from Tukey's HSD tests ($\alpha = 0.05$, B=10,000 permutations) for each CCRS metric. Diff = (Mean1 - Mean2) in percentage points. Significant p-values (<0.05) are bolded.

\begin{table}[H] \centering \caption{Contextual Coherence (CC) Pairwise Comparisons} \label{tab:app_stats_cc_full_final} \small
\begin{tabular}{@{}lrrrc@{}} \toprule Comparison & Mean Diff & P-value & Significant? \\ \midrule A vs B & -0.66 & 0.8236 & No \\ A vs C & +20.06 & \textbf{$<$0.0001} & Yes \\ A vs D & +18.66 & \textbf{$<$0.0001} & Yes \\ A vs E & +19.07 & \textbf{$<$0.0001} & Yes \\ A vs F & +17.07 & \textbf{$<$0.0001} & Yes \\ B vs C & +20.72 & \textbf{$<$0.0001} & Yes \\ B vs D & +19.32 & \textbf{$<$0.0001} & Yes \\ B vs E & +19.73 & \textbf{$<$0.0001} & Yes \\ B vs F & +17.73 & \textbf{$<$0.0001} & Yes \\ C vs D & -1.40 & 0.1017 & No \\ C vs E & -0.99 & 0.4446 & No \\ C vs F & -2.99 & \textbf{$<$0.0001} & Yes \\ D vs E & +0.41 & 0.9736 & No \\ D vs F & -1.59 & \textbf{0.0417} & Yes \\ E vs F & -2.00 & \textbf{0.0036} & Yes \\ \bottomrule \end{tabular} \end{table}

\begin{table}[H] \centering \caption{Question Relevance (QR) Pairwise Comparisons} \label{tab:app_stats_qr_full_final} \small
\begin{tabular}{@{}lrrrc@{}} \toprule Comparison & Mean Diff & P-value & Significant? \\ \midrule A vs B & -2.57 & \textbf{$<$0.0001} & Yes \\ A vs C & +5.76 & \textbf{$<$0.0001} & Yes \\ A vs D & +2.28 & \textbf{$<$0.0001} & Yes \\ A vs E & +4.55 & \textbf{$<$0.0001} & Yes \\ A vs F & +2.00 & \textbf{$<$0.0001} & Yes \\ B vs C & +8.33 & \textbf{$<$0.0001} & Yes \\ B vs D & +4.85 & \textbf{$<$0.0001} & Yes \\ B vs E & +7.12 & \textbf{$<$0.0001} & Yes \\ B vs F & +4.57 & \textbf{$<$0.0001} & Yes \\ C vs D & -3.48 & \textbf{$<$0.0001} & Yes \\ C vs E & -1.21 & \textbf{0.0037} & Yes \\ C vs F & -3.76 & \textbf{$<$0.0001} & Yes \\ D vs E & +2.27 & \textbf{$<$0.0001} & Yes \\ D vs F & -0.28 & 0.9553 & No \\ E vs F & -2.55 & \textbf{$<$0.0001} & Yes \\ \bottomrule \end{tabular} \end{table}

\begin{table}[H] \centering \caption{Information Density (ID) Pairwise Comparisons} \label{tab:app_stats_id_full_final} \small
\begin{tabular}{@{}lrrrc@{}} \toprule Comparison & Mean Diff & P-value & Significant? \\ \midrule A vs B & -0.09 & 0.9998 & No \\ A vs C & +7.72 & \textbf{$<$0.0001} & Yes \\ A vs D & +6.40 & \textbf{$<$0.0001} & Yes \\ A vs E & +6.58 & \textbf{$<$0.0001} & Yes \\ A vs F & +6.38 & \textbf{$<$0.0001} & Yes \\ B vs C & +7.81 & \textbf{$<$0.0001} & Yes \\ B vs D & +6.49 & \textbf{$<$0.0001} & Yes \\ B vs E & +6.67 & \textbf{$<$0.0001} & Yes \\ B vs F & +6.47 & \textbf{$<$0.0001} & Yes \\ C vs D & -1.32 & \textbf{0.0004} & Yes \\ C vs E & -1.14 & \textbf{0.0032} & Yes \\ C vs F & -1.34 & \textbf{0.0004} & Yes \\ D vs E & +0.18 & 0.9932 & No \\ D vs F & -0.02 & 1.0000 & No \\ E vs F & -0.20 & 0.9891 & No \\ \bottomrule \end{tabular} \end{table}

\begin{table}[H] \centering \caption{Answer Correctness (AC) Pairwise Comparisons} \label{tab:app_stats_ac_full_final} \small
\begin{tabular}{@{}lrrrc@{}} \toprule Comparison & Mean Diff & P-value & Significant? \\ \midrule A vs B & -0.71 & \textbf{$<$0.0001} & Yes \\ A vs C & +2.56 & \textbf{$<$0.0001} & Yes \\ A vs D & +1.86 & \textbf{$<$0.0001} & Yes \\ A vs E & +2.69 & \textbf{$<$0.0001} & Yes \\ A vs F & +2.34 & \textbf{$<$0.0001} & Yes \\ B vs C & +3.27 & \textbf{$<$0.0001} & Yes \\ B vs D & +2.57 & \textbf{$<$0.0001} & Yes \\ B vs E & +3.40 & \textbf{$<$0.0001} & Yes \\ B vs F & +3.06 & \textbf{$<$0.0001} & Yes \\ C vs D & -0.70 & \textbf{$<$0.0001} & Yes \\ C vs E & +0.13 & 0.9025 & No \\ C vs F & -0.22 & 0.5122 & No \\ D vs E & +0.83 & \textbf{$<$0.0001} & Yes \\ D vs F & +0.48 & \textbf{0.0020} & Yes \\ E vs F & -0.35 & 0.0620 & No \\ \bottomrule \end{tabular} \end{table}

\begin{table}[H] \centering \caption{Information Recall (IR) Pairwise Comparisons} \label{tab:app_stats_ir_full_final} \small
\begin{tabular}{@{}lrrrc@{}} \toprule Comparison & Mean Diff & P-value & Significant? \\ \midrule A vs B & -3.32 & \textbf{$<$0.0001} & Yes \\ A vs C & +11.96 & \textbf{$<$0.0001} & Yes \\ A vs D & +9.52 & \textbf{$<$0.0001} & Yes \\ A vs E & +12.92 & \textbf{$<$0.0001} & Yes \\ A vs F & +10.17 & \textbf{$<$0.0001} & Yes \\ B vs C & +15.28 & \textbf{$<$0.0001} & Yes \\ B vs D & +12.84 & \textbf{$<$0.0001} & Yes \\ B vs E & +16.24 & \textbf{$<$0.0001} & Yes \\ B vs F & +13.49 & \textbf{$<$0.0001} & Yes \\ C vs D & -2.44 & \textbf{$<$0.0001} & Yes \\ C vs E & +0.96 & 0.2123 & No \\ C vs F & -1.78 & \textbf{0.0006} & Yes \\ D vs E & +3.39 & \textbf{$<$0.0001} & Yes \\ D vs F & +0.65 & 0.6418 & No \\ E vs F & -2.74 & \textbf{$<$0.0001} & Yes \\ \bottomrule \end{tabular} \end{table}

\section{Statistical Testing Implementation Code}
\label{app:code}
The Python code used for performing the Tukey's HSD with randomization and the Holm-Bonferroni correction across hypotheses.

\begin{lstlisting}[language=Python, caption={Statistical Testing Implementation for RAG Hypotheses}, label={code:stats_appendix_final}]
import numpy as np
from scipy import stats
import json
from typing import Dict, List, Tuple, Any
from tqdm import tqdm
import argparse # Added for file arguments

def load_system_data(files: Dict[str, str]) -> Dict[str, np.ndarray]:
    """Load data for all systems and extract metrics."""
    print("\nLoading system data...")
    data_systems = {}
    # Ensure metrics match the order used in analysis consistently
    metrics = ['Contextual_Coherence', 'Question_Relevance', 'Information_Density',
               'Answer_Correctness', 'Information_Recall']

    for sys_label, file_path in files.items():
        print(f"Loading system {sys_label} from {file_path}")
        try:
            with open(file_path, 'r') as f:
                data = json.load(f)
                # Check if 'ccrs_results' key exists
                if 'ccrs_results' not in data:
                    print(f"Error: 'ccrs_results' key not found in {file_path}")
                    return None

                results_list = data['ccrs_results']
                if not isinstance(results_list, list):
                    print(f"Error: 'ccrs_results' is not a list in {file_path}")
                    return None

                processed_data = []
                for i, m in enumerate(results_list):
                    if not isinstance(m, dict) or 'metrics' not in m:
                        print(f"Warning: Invalid item format at index {i} in {file_path}. Skipping.")
                        continue
                    metric_values = []
                    valid_item = True
                    for metric in metrics:
                        if metric not in m['metrics']:
                            print(f"Warning: Metric '{metric}' not found for item {i} in {file_path}. Skipping item.")
                            valid_item = False
                            break
                        # Ensure score is numeric and handle potential None or non-numeric types
                        score = m['metrics'][metric]
                        if isinstance(score, (int, float)):
                            metric_values.append(score / 100.0) # Normalize
                        else:
                            print(f"Warning: Non-numeric score '{score}' for metric '{metric}' item {i} in {file_path}. Skipping item.")
                            valid_item = False
                            break
                    if valid_item:
                        processed_data.append(metric_values)

                if not processed_data: # If no valid data was processed
                     print(f"Error: No valid data processed from {file_path}")
                     return None

                data_systems[sys_label] = np.array(processed_data)

            print(f"System {sys_label} loaded: shape {data_systems[sys_label].shape}")

        except FileNotFoundError:
            print(f"Error: File not found {file_path}")
            return None
        except json.JSONDecodeError:
             print(f"Error: Could not decode JSON from {file_path}")
             return None
        except Exception as e:
             print(f"An unexpected error occurred loading {file_path}: {e}")
             return None

    # Final checks after loading all files
    if len(data_systems) != len(files):
        print("Error: Could not load all specified system files.")
        return None
    if any(arr.size == 0 for arr in data_systems.values()):
        print("Error: At least one system resulted in no valid data after processing.")
        return None


    # Check for shape consistency (number of samples and metrics)
    first_shape = next(iter(data_systems.values())).shape
    if len(first_shape) != 2:
        print("Error: Loaded data arrays do not have 2 dimensions (samples, metrics).")
        return None
    num_samples = first_shape[0]
    num_metrics_loaded = first_shape[1]

    if not all(arr.shape[0] == num_samples for arr in data_systems.values()):
         print("Error: Loaded data arrays have inconsistent number of samples (queries).")
         return None
    if num_metrics_loaded != len(metrics):
         print(f"Error: Expected {len(metrics)} metrics, but loaded {num_metrics_loaded}.")
         return None

    print(f"Successfully loaded data for {len(data_systems)} systems with {num_samples} samples each.")
    return data_systems


def tukeys_hsd_randomization(X: np.ndarray, pairs: List[Tuple[int, int]],
                           B: int = 10000,
                           one_sided: bool = False,
                           expected_directions: Dict[str, int] = None) -> Tuple[Dict[str, float],
                             Dict[str, float],
                             np.ndarray]:
    """Perform Tukey's HSD with randomization for multiple pairs."""
    n = X.shape[0] # Number of samples
    m = X.shape[1] # Number of systems

    if n == 0 or m < 2 or not pairs:
        print("Warning: Invalid input for randomization test.")
        means = np.array([]) if m == 0 else np.nanmean(X, axis=0)
        return {}, {}, means

    valid_pairs = [(i, j) for i, j in pairs if i < m and j < m]
    if not valid_pairs:
        print("Warning: No valid pairs provided for comparison.")
        return {}, {}, np.nanmean(X, axis=0)
    pair_keys = [f"{i},{j}" for i, j in valid_pairs]

    p_values = {key: 0.0 for key in pair_keys}
    observed_means = np.nanmean(X, axis=0)
    observed_diffs = {f"{i},{j}": observed_means[i] - observed_means[j] for i, j in valid_pairs}

    # print("Observed means:", np.round(observed_means * 100, 3))
    # print("Observed differences:", {k: np.round(v * 100, 3) for k, v in observed_diffs.items() if not np.isnan(v)})
    # print(f"Running {B} permutations...")

    X_perm = np.copy(X)
    count_greater_equal = {key: 0 for key in pair_keys}
    nan_diff_keys = {k for k, v in observed_diffs.items() if np.isnan(v)}
    if nan_diff_keys: print(f"Warning: NaN observed difference for pairs: {nan_diff_keys}.")

    for b in tqdm(range(B), desc="Permutations", ncols=100, leave=False, mininterval=1.0):
        # Efficient permutation: shuffle indices for each row
        for t in range(n):
            row_data = X[t, :]
            non_nan_indices = ~np.isnan(row_data)
            if not np.any(non_nan_indices): continue # Skip rows with all NaNs
            permuted_non_nan = np.random.permutation(row_data[non_nan_indices])
            row_perm = np.full(m, np.nan)
            row_perm[non_nan_indices] = permuted_non_nan
            X_perm[t, :] = row_perm

        means_star = np.nanmean(X_perm, axis=0)
        valid_means_star = means_star[~np.isnan(means_star)]
        if len(valid_means_star) < 2: continue # Need at least 2 non-NaN means
        q_star = np.nanmax(means_star) - np.nanmin(means_star)
        if np.isnan(q_star): continue

        for pair_key in pair_keys:
             if pair_key in nan_diff_keys: continue
             # Use tolerance for floating point comparison
             if q_star >= abs(observed_diffs[pair_key]) - 1e-9:
                 count_greater_equal[pair_key] += 1

    # Finalize p-values
    for pair_key in pair_keys:
        if pair_key in nan_diff_keys:
            p_values[pair_key] = np.nan
        else:
            # Use (count)/B as p-value estimate
            p_values[pair_key] = count_greater_equal[pair_key] / B

    # Adjust for one-sided tests
    if one_sided:
        if expected_directions is None: raise ValueError("Expected directions needed for one-sided test.")
        # print("Adjusting for one-sided test...")
        for pair_key in list(p_values.keys()):
            if np.isnan(p_values[pair_key]): continue

            i_str, j_str = pair_key.split(',')
            i, j = int(i_str), int(j_str)
            lookup_key = f"{i},{j}"
            expected_dir = expected_directions.get(lookup_key)
            if expected_dir is None:
                 # Try reverse pair if direction not defined for i,j
                 reverse_lookup_key = f"{j},{i}"
                 expected_dir = expected_directions.get(reverse_lookup_key)
                 if expected_dir is not None: expected_dir *= -1 # Invert direction
                 else: raise KeyError(f"Direction for pair {lookup_key} or {reverse_lookup_key} not found.")

            observed_diff = observed_diffs[pair_key]
            if np.isnan(observed_diff): continue

            current_p = p_values[pair_key]
            # Check if observed difference matches expected direction (use tolerance)
            is_correct_direction = (expected_dir > 0 and observed_diff > 1e-9) or \
                                   (expected_dir < 0 and observed_diff < -1e-9)
            is_zero_diff = abs(observed_diff) < 1e-9

            if is_correct_direction:
                # If direction is correct, p-value is halved
                p_values[pair_key] = current_p / 2.0
            elif expected_dir != 0 and is_zero_diff:
                 # If difference is effectively zero for a directional test
                 p_values[pair_key] = 0.5
            elif expected_dir == 0:
                 # Keep two-sided p-value if no direction expected (should not happen here)
                 pass
            else: # Observed difference is in the opposite direction
                 # p-value becomes 1 - (two-sided p / 2)
                 p_values[pair_key] = 1.0 - (current_p / 2.0)
                 p_values[pair_key] = min(p_values[pair_key], 1.0) # Cap at 1.0

    # print("Final p-values:", {k: (f"{v:.6f}" if not np.isnan(v) else "NaN") for k, v in p_values.items()})
    return p_values, observed_diffs, observed_means


def apply_holm_bonferroni(p_values: List[float], names: List[str],
                         alpha: float = 0.05) -> Tuple[Dict[str, bool], Dict[str, float]]:
    """Apply Holm-Bonferroni correction across hypotheses."""
    n = len(p_values)
    if n == 0: return {}, {}

    # Handle potential NaNs by treating them as non-significant (p=1.0) for sorting
    p_values_nonan = [p if not np.isnan(p) else 1.0 for p in p_values]
    indices = range(n)
    # Sort by non-NaN p-value, keep original p-value, index, name
    sorted_pairs = sorted(zip(p_values_nonan, p_values, indices, names))

    reject = {name: False for name in names}
    adjusted_p_values = {name: 1.0 for name in names} # Initialize adjusted p-values

    print("\nHolm-Bonferroni correction across hypotheses:")
    significant_found = False
    for i, (p_sort, p_orig, idx, name) in enumerate(sorted_pairs):
        # Use original p-value (which might be NaN) for reporting
        raw_p_report = p_orig if not np.isnan(p_orig) else 1.0

        # Holm compares raw p-value with adjusted alpha
        holm_alpha = alpha / (n - i)

        print(f" {i+1}. Hyp: {name}: raw p={raw_p_report:.6f}, Compare with alpha_adj={holm_alpha:.6f}")

        # Use p_sort (non-NaN version) for the comparison logic
        if p_sort <= holm_alpha:
            print(f"   -> Reject H0 for {name}")
            reject[name] = True
            significant_found = True
        else:
            print(f"   -> Fail to reject H0 for {name} (and subsequent hypotheses)")
            # Once we fail to reject, stop rejecting for subsequent hypotheses
            break # Optimization: no need to check further hypotheses

    # Calculate final adjusted p-values ensuring monotonicity
    # Adjusted p-value for hypothesis i is max(p_j * (n - j + 1)) for j <= i (where p_j is sorted raw p)
    last_p_adj = 0.0
    for i, (p_sort, p_orig, idx, name) in enumerate(sorted_pairs):
         holm_p = min(1.0, p_sort * (n - i)) # Adjusted p for this step
         last_p_adj = max(last_p_adj, holm_p) # Enforce monotonicity
         adjusted_p_values[name] = last_p_adj if not np.isnan(p_orig) else 1.0 # Assign adjusted p, keep NaN as 1.0

    if not significant_found:
        print(" No hypotheses rejected.")

    return reject, adjusted_p_values


def test_all_hypotheses(data_systems: Dict[str, np.ndarray],
                       alpha: float = 0.05) -> Dict[str, Any]:
    """Run all hypothesis tests with correct multiple testing correction."""
    metric_names = ['CC', 'QR', 'ID', 'AC', 'IR']
    h2_metrics = ['QR', 'IR'] # Metrics relevant for H2
    hypothesis_names = ['H1', 'H2', 'H3']
    system_labels = ['A', 'B', 'C', 'D', 'E', 'F']
    sys_idx = {label: i for i, label in enumerate(system_labels)}

    # Define comparison pairs indices based on system_labels A=0, B=1, ... F=5
    h1_pairs = [(sys_idx['A'], sys_idx['C']), (sys_idx['A'], sys_idx['E']),
                (sys_idx['B'], sys_idx['D']), (sys_idx['B'], sys_idx['F'])]
    h2_pairs = [(sys_idx['D'], sys_idx['C']), (sys_idx['F'], sys_idx['E'])]
    h3_pairs = [(sys_idx['C'], sys_idx['E']), (sys_idx['D'], sys_idx['F'])]

    # Expected directions for one-sided tests (H1: Mistral > Llama, H2: E5 > BM25)
    h1_directions = {f"{i},{j}": 1 for i,j in h1_pairs} # Assuming mean(A/B) > mean(C/D/E/F)
    h2_directions = {f"{i},{j}": 1 for i,j in h2_pairs} # Assuming mean(D/F) > mean(C/E)

    # Storage for aggregated p-values per hypothesis
    h1_pvals_all_metrics_pairs = []
    h2_pvals_all_metrics_pairs = []
    h3_min_p_by_metric = {}
    results_by_metric = {}
    # Combine data for easier slicing: shape (n_samples, n_metrics, n_systems)
    all_system_data_array = np.stack([data_systems[lbl] for lbl in system_labels], axis=2)

    for metric_idx, metric_name in enumerate(metric_names):
        print(f"\n===== Testing metric: {metric_name} =====")
        # Extract data for the current metric: shape (n_samples, n_systems)
        metric_data = all_system_data_array[:, metric_idx, :]

        # --- H1: Mistral Superiority ---
        print("\n--- H1 (Mistral > Llama) ---")
        h1_pvals_pairs, h1_diffs, _ = tukeys_hsd_randomization(metric_data, h1_pairs, one_sided=True, expected_directions=h1_directions)
        # Collect valid p-values for H1 aggregation
        h1_pvals_all_metrics_pairs.extend([p for p in h1_pvals_pairs.values() if p is not None and not np.isnan(p)])

        # --- H2: E5 Advantage (Llama only) ---
        h2_pvals_pairs_metric = {}
        h2_diffs_metric = {}
        if metric_name in h2_metrics:
            print(f"\n--- H2 (E5 > BM25 for Llama) ---")
            h2_pvals_pairs_metric, h2_diffs_metric, _ = tukeys_hsd_randomization(metric_data, h2_pairs, one_sided=True, expected_directions=h2_directions)
            # Collect valid p-values for H2 aggregation
            h2_pvals_all_metrics_pairs.extend([p for p in h2_pvals_pairs_metric.values() if p is not None and not np.isnan(p)])
        else: # Initialize placeholders if metric not relevant for H2
            h2_pvals_pairs_metric = {f"{i},{j}": np.nan for i,j in h2_pairs}
            h2_diffs_metric = {f"{i},{j}": np.nan for i,j in h2_pairs}

        # --- H3: Llama Model Differences ---
        print(f"\n--- H3 (Llama 8B vs 3.2B) ---")
        # Two-sided test for H3
        h3_pvals_pairs, h3_diffs, means = tukeys_hsd_randomization(metric_data, h3_pairs, one_sided=False)
        # Find the minimum non-NaN p-value for this metric for H3 aggregation
        valid_h3_pvals = [p for p in h3_pvals_pairs.values() if p is not None and not np.isnan(p)]
        h3_min_p_by_metric[metric_name] = min(valid_h3_pvals) if valid_h3_pvals else 1.0

        # Store detailed results per metric
        results_by_metric[metric_name] = {
            'h1_pairs_results': {f"{system_labels[i]}_vs_{system_labels[j]}": {'p_value': h1_pvals_pairs.get(f"{i},{j}", np.nan), 'difference': h1_diffs.get(f"{i},{j}", np.nan)} for i,j in h1_pairs},
            'h2_pairs_results': {f"{system_labels[i]}_vs_{system_labels[j]}": {'p_value': h2_pvals_pairs_metric.get(f"{i},{j}", np.nan), 'difference': h2_diffs_metric.get(f"{i},{j}", np.nan)} for i,j in h2_pairs} if metric_name in h2_metrics else None,
            'h3_pairs_results': {f"{system_labels[i]}_vs_{system_labels[j]}": {'p_value': h3_pvals_pairs.get(f"{i},{j}", np.nan), 'difference': h3_diffs.get(f"{i},{j}", np.nan)} for i,j in h3_pairs},
            'means': (means * 100).round(3).tolist() if not np.isnan(means).all() else []
        }
        # print(f"\n{metric_name} Summary: H1 Max(p)={max(h1_pvals_pairs.values() if h1_pvals_pairs else [1.0]):.6f}" +
        #       (f", H2 Max(p)={max(h2_pvals_pairs_metric.values() if h2_pvals_pairs_metric and any(p is not None and not np.isnan(p) for p in h2_pvals_pairs_metric.values()) else [1.0]):.6f}" if metric_name in h2_metrics else "") +
        #       f", H3 Min(p)={h3_min_p_by_metric[metric_name]:.6f}")


    # Aggregate p-values across metrics for each hypothesis based on logic
    # H1/H2 (AND condition): max p-value across all relevant comparisons
    h1_final_p = max(h1_pvals_all_metrics_pairs) if h1_pvals_all_metrics_pairs else 1.0
    h2_final_p = max(h2_pvals_all_metrics_pairs) if h2_pvals_all_metrics_pairs else 1.0
    # H3 (OR condition): min p-value across metrics, then Bonferroni correct
    h3_min_p_across_metrics = min(h3_min_p_by_metric.values()) if h3_min_p_by_metric else 1.0
    h3_final_p_bonferroni = min(1.0, h3_min_p_across_metrics * len(metric_names)) # Bonferroni correction factor = number of metrics

    print("\n===== Aggregated Hypothesis p-values (before Holm-Bonferroni) =====")
    print(f"H1 (Max p across all comparisons): {h1_final_p:.6f}")
    print(f"H2 (Max p across relevant comparisons): {h2_final_p:.6f}")
    print(f"H3 (Bonf-corr Min p across metrics): {h3_final_p_bonferroni:.6f}")

    # Apply Holm-Bonferroni correction across the three final hypothesis p-values
    hypothesis_p_values = [h1_final_p, h2_final_p, h3_final_p_bonferroni]
    significant, final_adjusted_p_values = apply_holm_bonferroni(hypothesis_p_values, hypothesis_names, alpha)

    # Compile final results object
    results = {'hypothesis_results': {}, 'metric_results': results_by_metric}
    for h_idx, h_name in enumerate(hypothesis_names):
         results['hypothesis_results'][h_name] = {
             'significant': significant[h_name],
             'raw_aggregated_p_value': hypothesis_p_values[h_idx],
             'final_adjusted_p_value': final_adjusted_p_values[h_name],
             'conclusion': 'Reject H0' if significant[h_name] else 'Fail to reject H0'
         }
         if h_name == 'H3': # Add extra info for H3
             results['hypothesis_results'][h_name]['min_p_across_metrics_raw'] = h3_min_p_across_metrics
             results['hypothesis_results'][h_name]['metric_min_p_values'] = h3_min_p_by_metric

    print("\n===== Final Conclusions after Holm-Bonferroni Correction =====")
    for h_name, h_result in results['hypothesis_results'].items():
        p_val_desc = 'Max Raw p' if h_name in ['H1', 'H2'] else 'Bonf-corr Min Raw p'
        print(f"\n{h_name}: {p_val_desc}={h_result['raw_aggregated_p_value']:.6f}; Final adj. p={h_result['final_adjusted_p_value']:.6f} -> {h_result['conclusion']}")

    return results


def main():
    parser = argparse.ArgumentParser(description="Run RAG system hypothesis tests.")
    # Add arguments for file paths if needed, otherwise use defaults
    parser.add_argument('--file_a', default='modified_bioasq_bm25_100_0.2_k20_Mistral-7B-results.json')
    parser.add_argument('--file_b', default='modified_bioasq_e5_mistral_100_0.2_k20_Mistral-7B-results.json')
    parser.add_argument('--file_c', default='modified_bioasq_bm25_100_0.2_k20_Llama3-8B-results.json')
    parser.add_argument('--file_d', default='modified_bioasq_e5_mistral_100_0.2_k20_Llama3-8B-results.json')
    parser.add_argument('--file_e', default='modified_bioasq_bm25_100_0.2_k20_Llama3.2-3B-results.json')
    parser.add_argument('--file_f', default='modified_bioasq_e5_mistral_100_0.2_k20_Llama3.2-3B-results.json')
    parser.add_argument('--alpha', type=float, default=0.05, help='Significance level alpha.')
    parser.add_argument('--output', default='hypothesis_test_results_full.json', help='Output JSON file name.')

    args = parser.parse_args()

    files = {
        'A': args.file_a, 'B': args.file_b, 'C': args.file_c,
        'D': args.file_d, 'E': args.file_e, 'F': args.file_f
    }

    print("Starting hypothesis testing...")
    data_systems = load_system_data(files)
    if data_systems is None:
        print("Failed to load data. Exiting.")
        return

    results = test_all_hypotheses(data_systems, alpha=args.alpha)

    # Helper function for JSON serialization of numpy types
    def convert_numpy(obj):
        if isinstance(obj, np.ndarray): return obj.tolist()
        if isinstance(obj, np.generic): return obj.item()
        if isinstance(obj, (np.bool_,)): return bool(obj)
        if isinstance(obj, dict): return {k: convert_numpy(v) for k, v in obj.items()}
        if isinstance(obj, list): return [convert_numpy(i) for i in obj]
        # Handle NaNs gracefully for JSON output
        if isinstance(obj, float) and np.isnan(obj): return None # Represent NaN as null
        return obj

    try:
        serializable_results = convert_numpy(results)
        with open(args.output, 'w') as f:
            json.dump(serializable_results, f, indent=2)
        print(f"\nComplete results saved to {args.output}")
    except Exception as e:
        print(f"Error saving results to JSON: {e}")

if __name__ == "__main__":
    main()
\end{lstlisting}

\clearpage %

\begin{figure*}[htbp!]
\centering
\small
\begin{subfigure}[b]{0.32\textwidth}
    \centering
    \includegraphics[width=\textwidth]{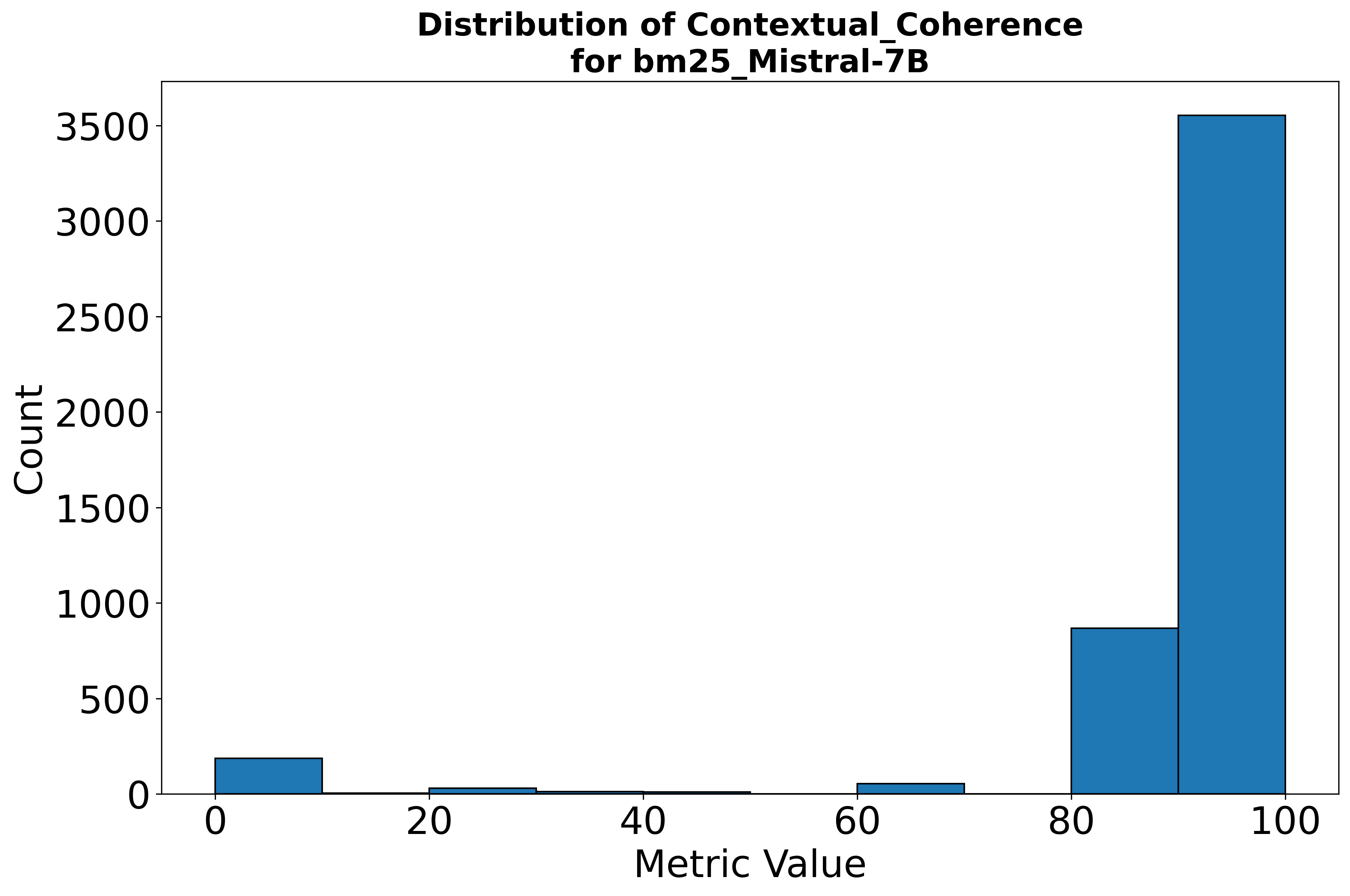}
    \caption{System A (CC)}
\end{subfigure} \hfill
\begin{subfigure}[b]{0.32\textwidth}
    \centering
    \includegraphics[width=\textwidth]{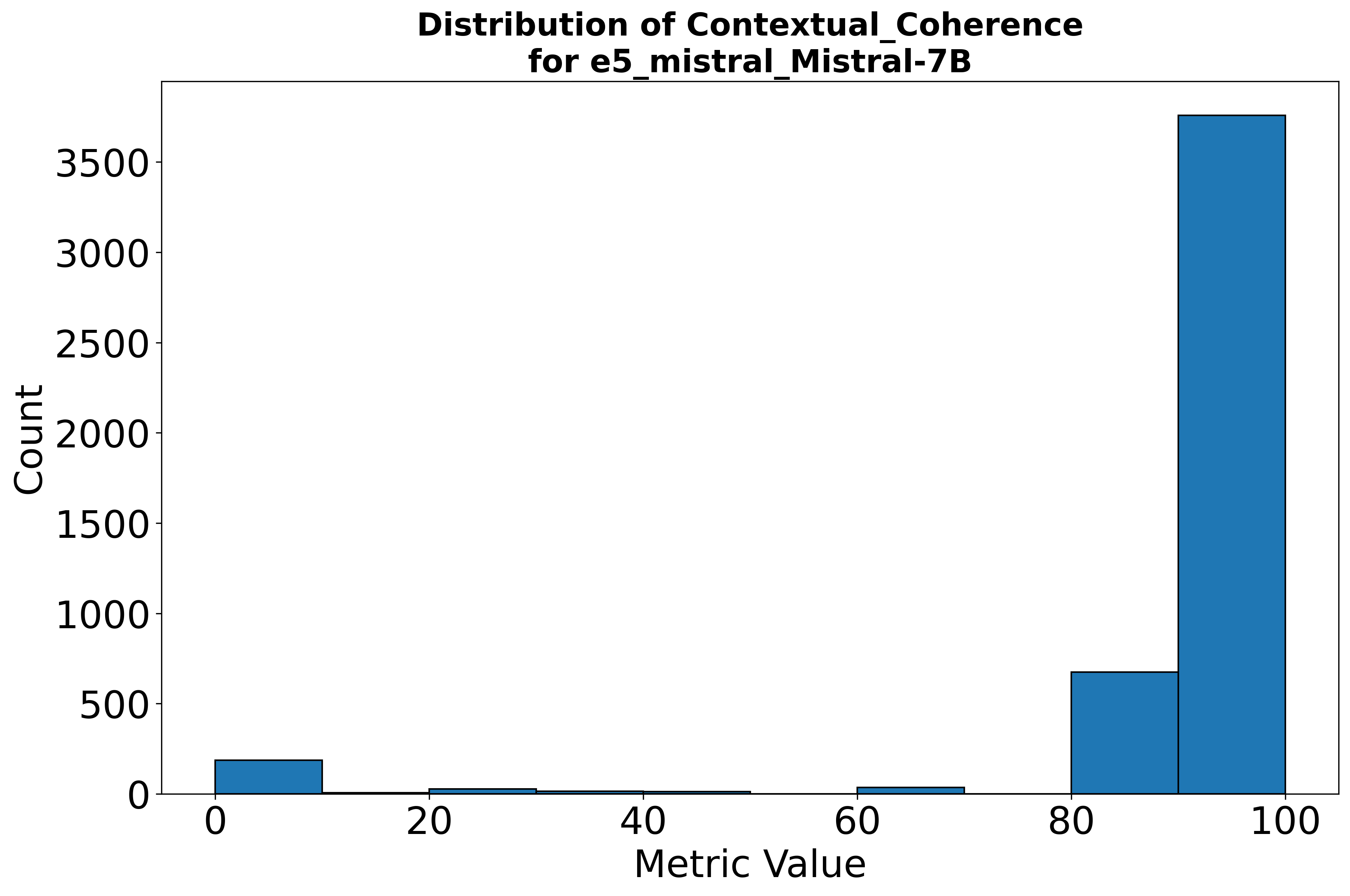}
    \caption{System B (CC)}
\end{subfigure} \hfill
\begin{subfigure}[b]{0.32\textwidth}
    \centering
    \includegraphics[width=\textwidth]{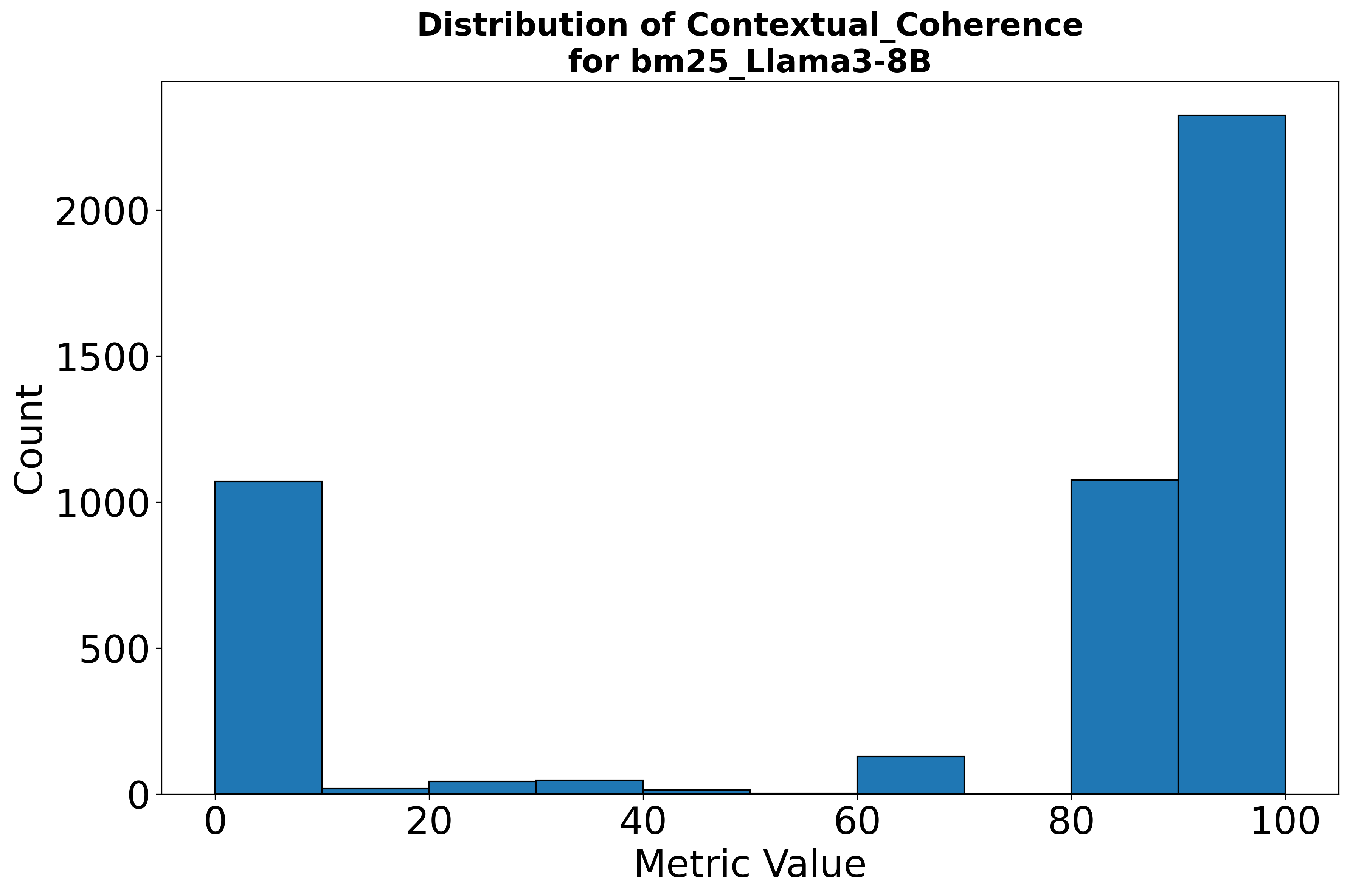}
    \caption{System C (CC)}
\end{subfigure} \\ \vspace{0.5em}
\begin{subfigure}[b]{0.32\textwidth}
    \centering
    \includegraphics[width=\textwidth]{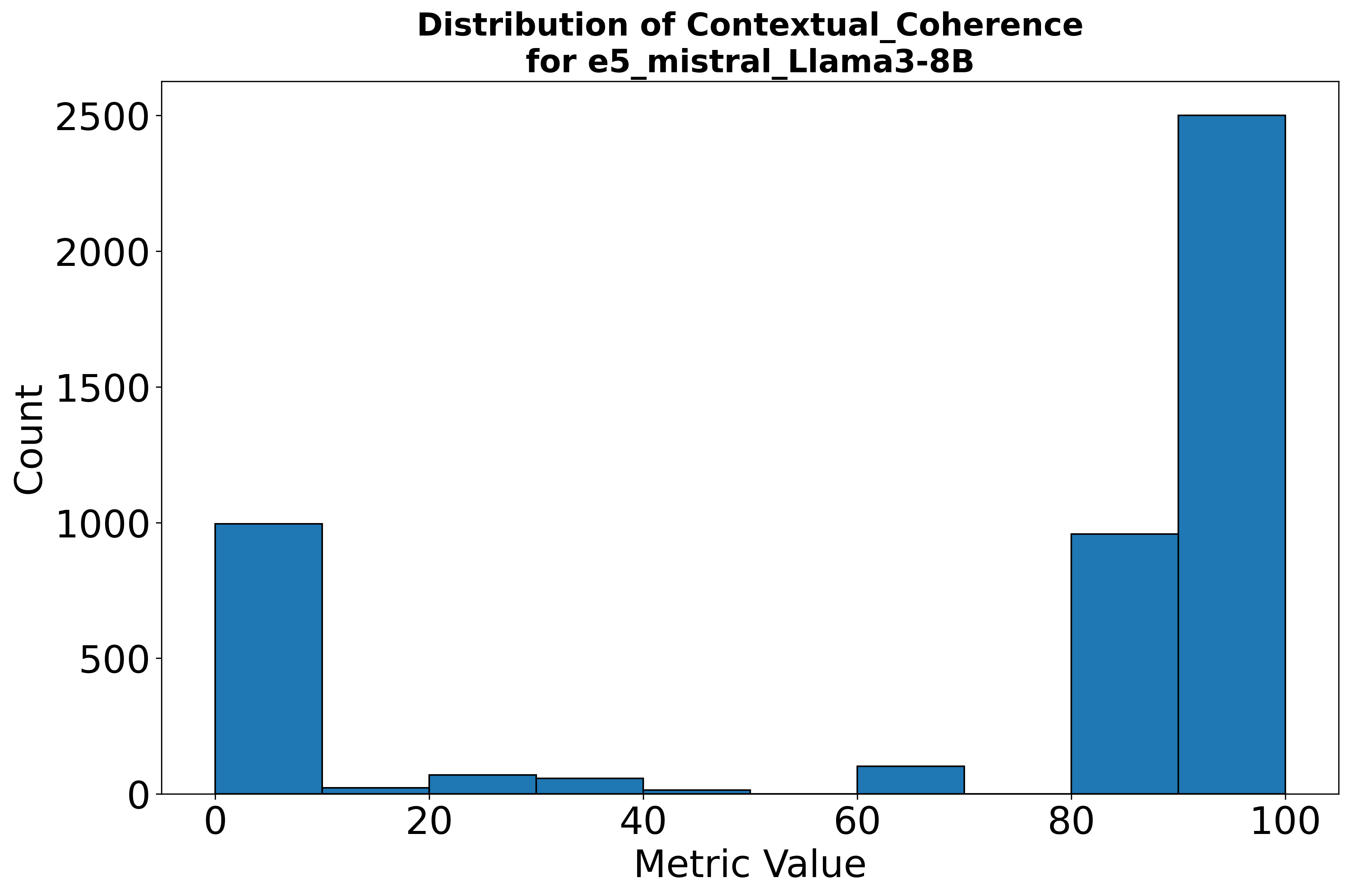}
    \caption{System D (CC)}
\end{subfigure} \hfill
\begin{subfigure}[b]{0.32\textwidth}
    \centering
    \includegraphics[width=\textwidth]{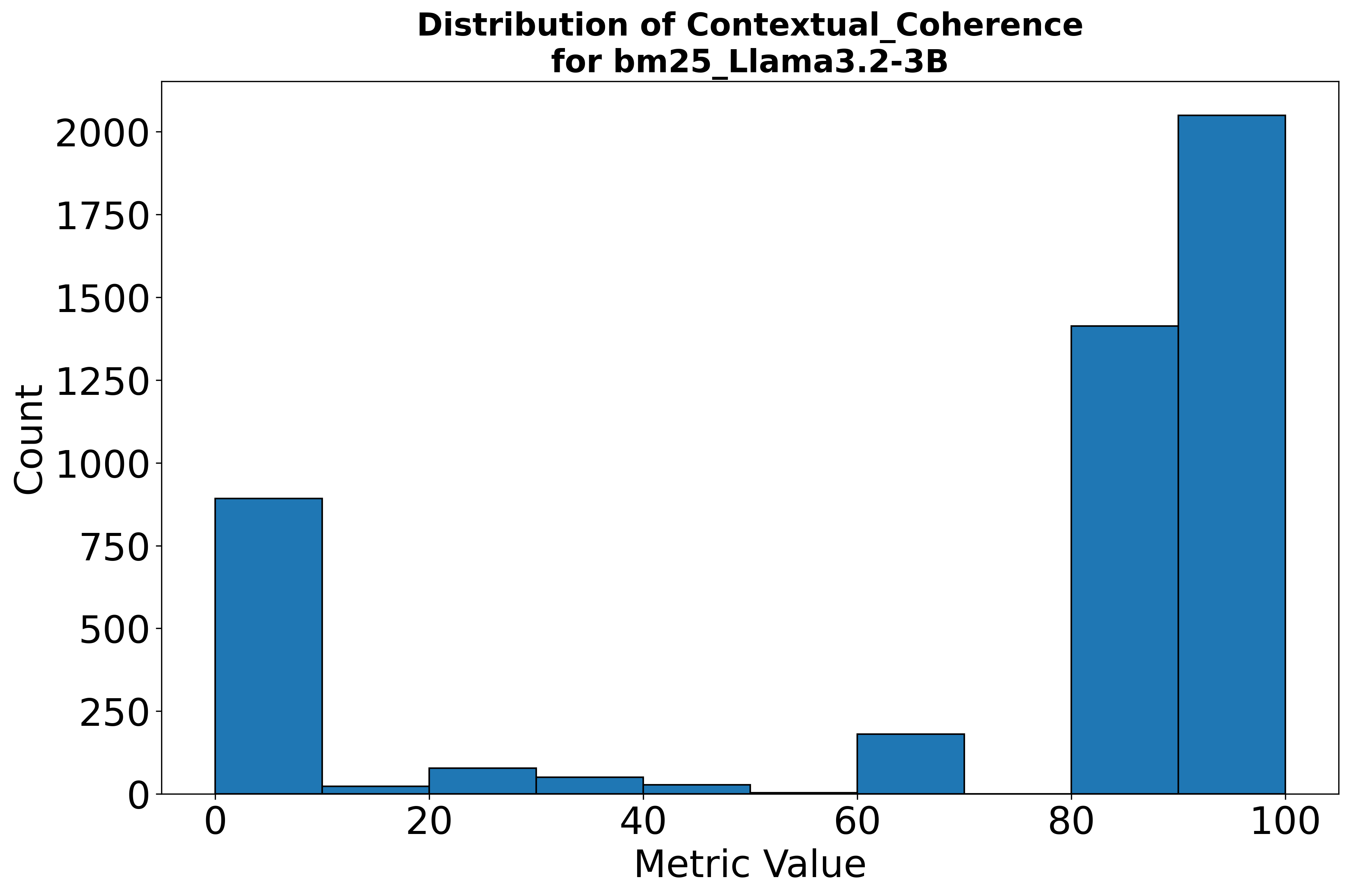}
    \caption{System E (CC)}
\end{subfigure} \hfill
\begin{subfigure}[b]{0.32\textwidth}
    \centering
    \includegraphics[width=\textwidth]{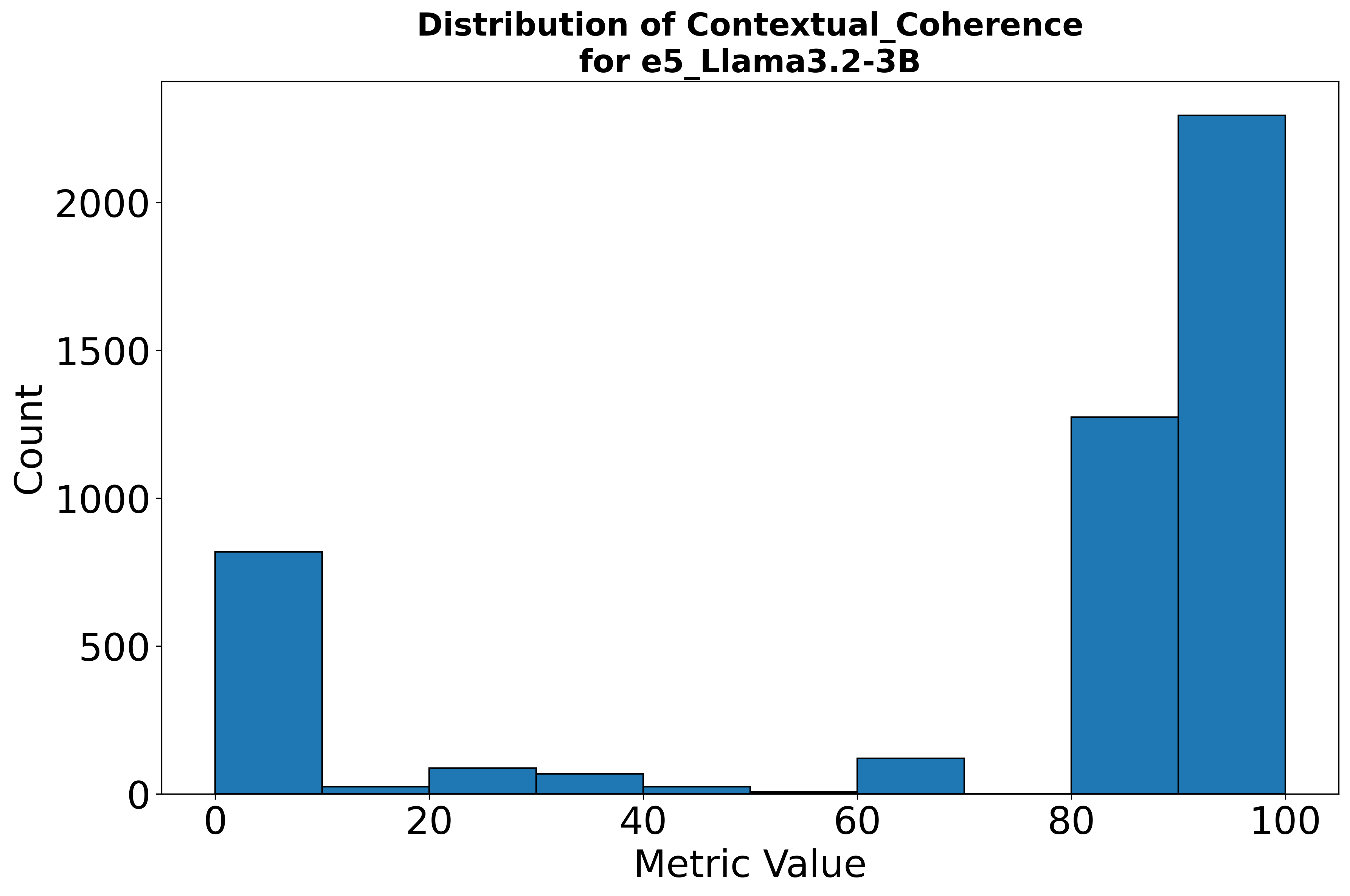}
    \caption{System F (CC)}
\end{subfigure}
\caption{Appendix: Distribution of Contextual Coherence (CC) scores for each RAG system.}
\label{fig:cc_dist}
\end{figure*}

\begin{figure*}[htbp!]
\centering
\small
\begin{subfigure}[b]{0.32\textwidth}
    \centering
    \includegraphics[width=\textwidth]{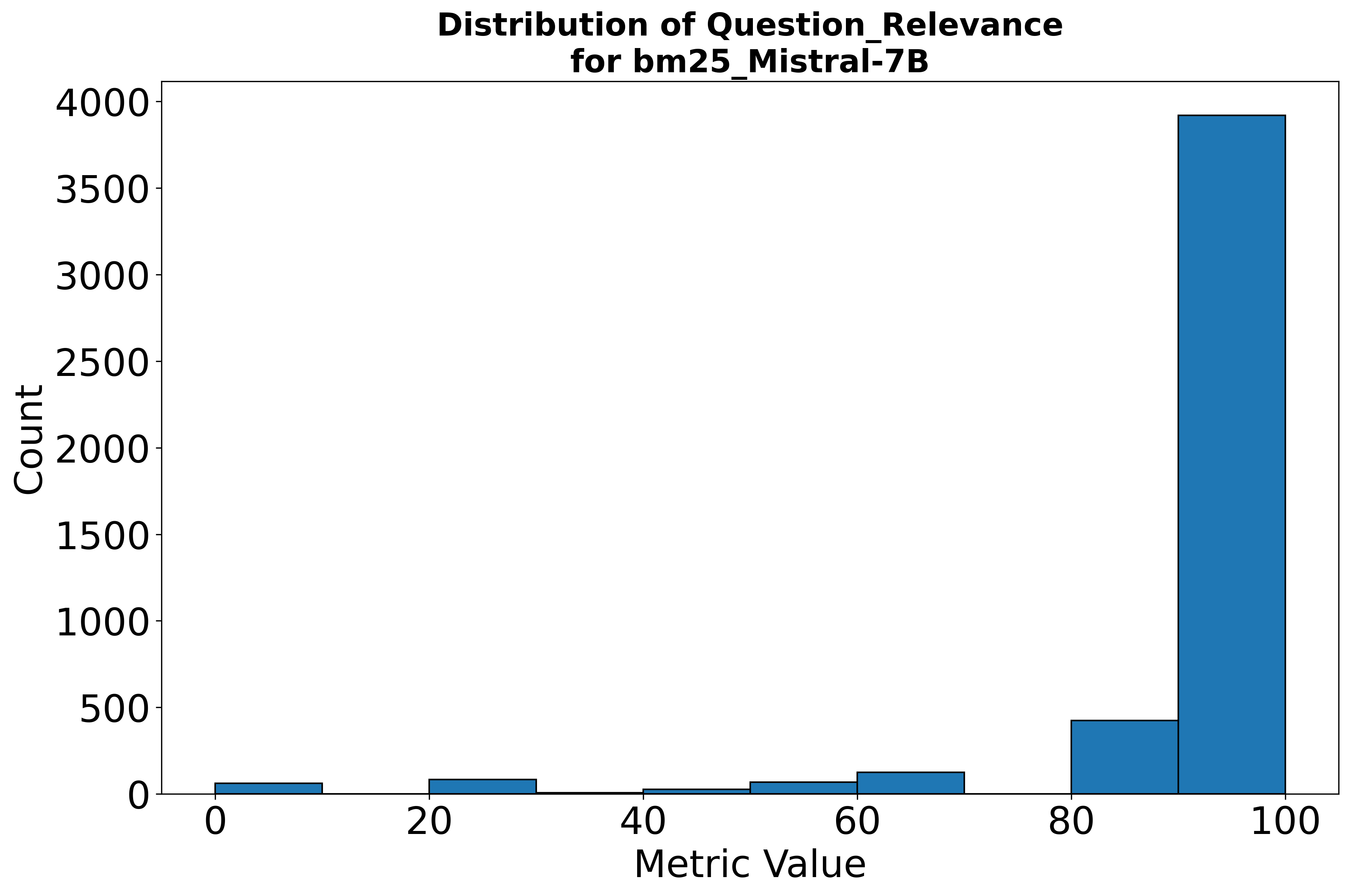}
    \caption{System A (QR)}
\end{subfigure} \hfill
\begin{subfigure}[b]{0.32\textwidth}
    \centering
    \includegraphics[width=\textwidth]{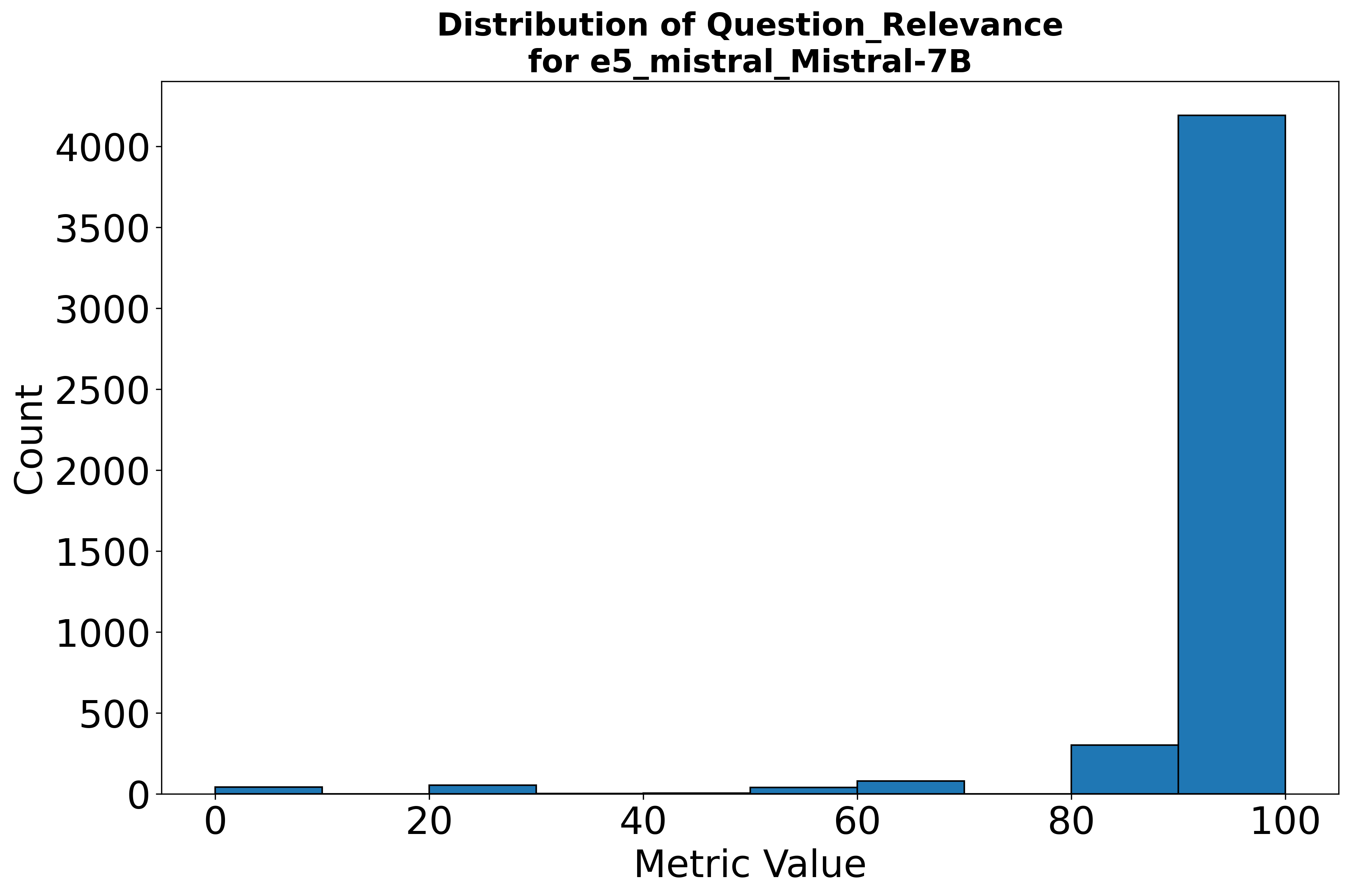}
    \caption{System B (QR)}
\end{subfigure} \hfill
\begin{subfigure}[b]{0.32\textwidth}
    \centering
    \includegraphics[width=\textwidth]{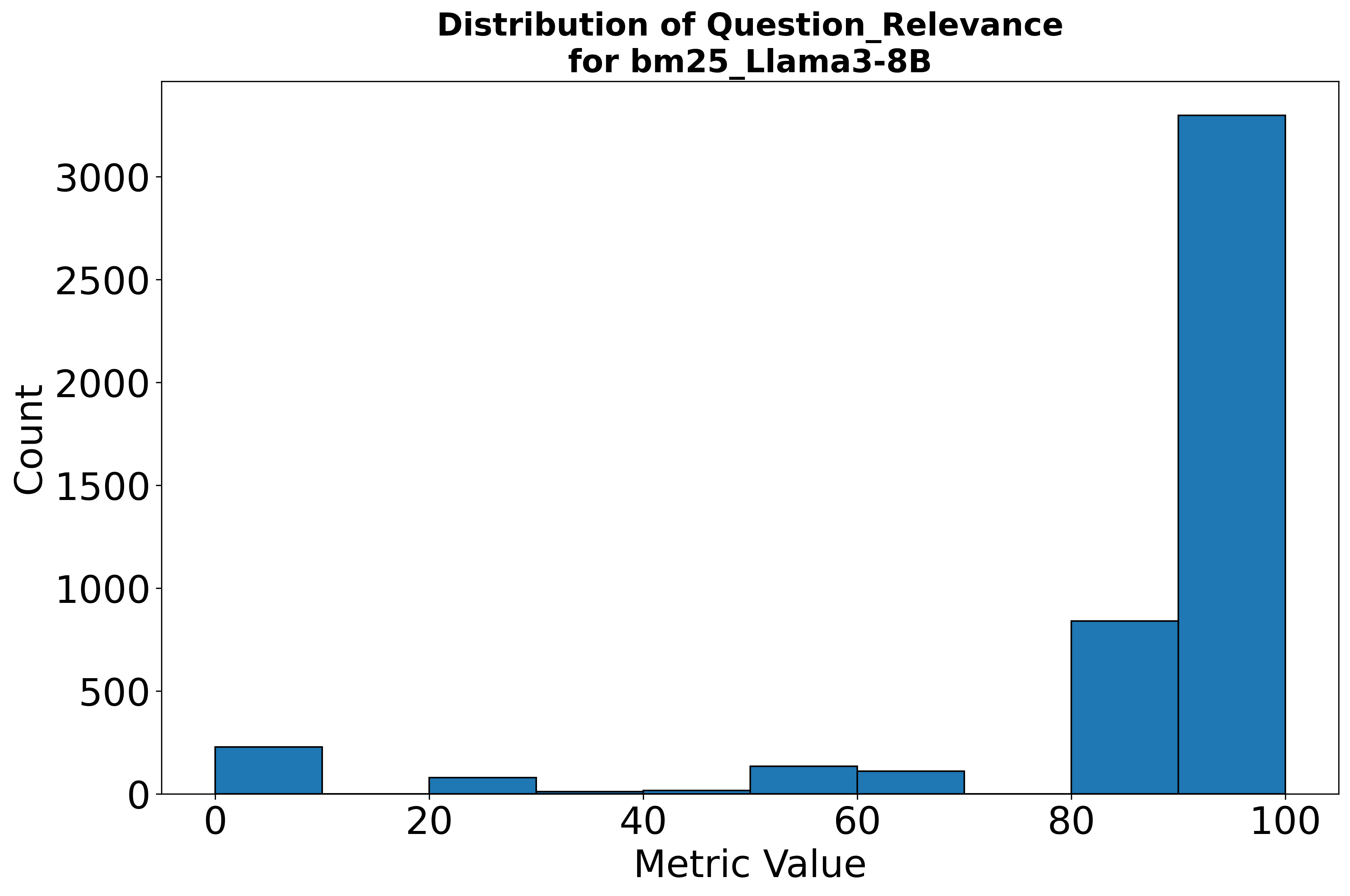}
    \caption{System C (QR)}
\end{subfigure} \\ \vspace{0.5em}
\begin{subfigure}[b]{0.32\textwidth}
    \centering
    \includegraphics[width=\textwidth]{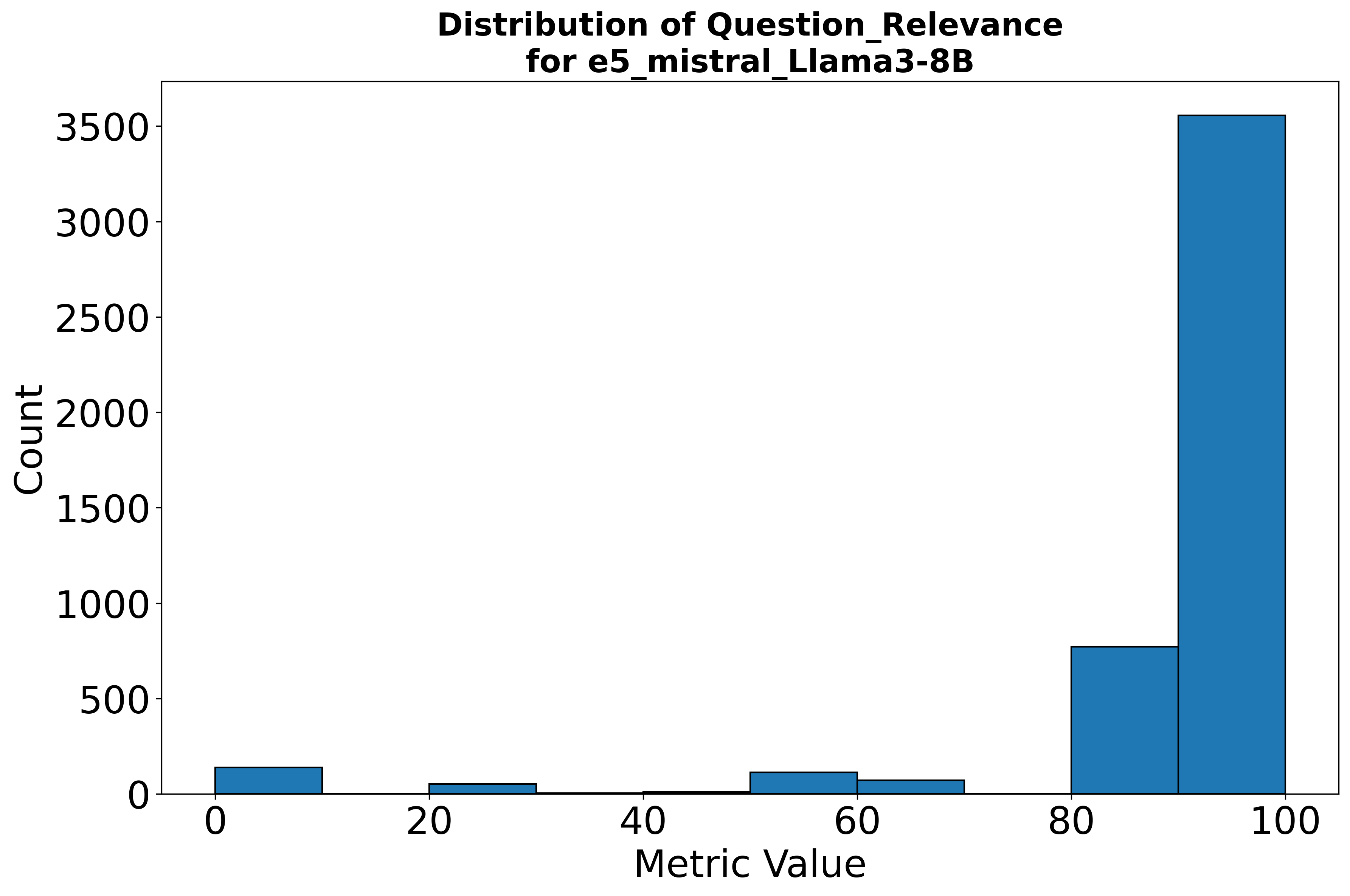}
    \caption{System D (QR)}
\end{subfigure} \hfill
\begin{subfigure}[b]{0.32\textwidth}
    \centering
    \includegraphics[width=\textwidth]{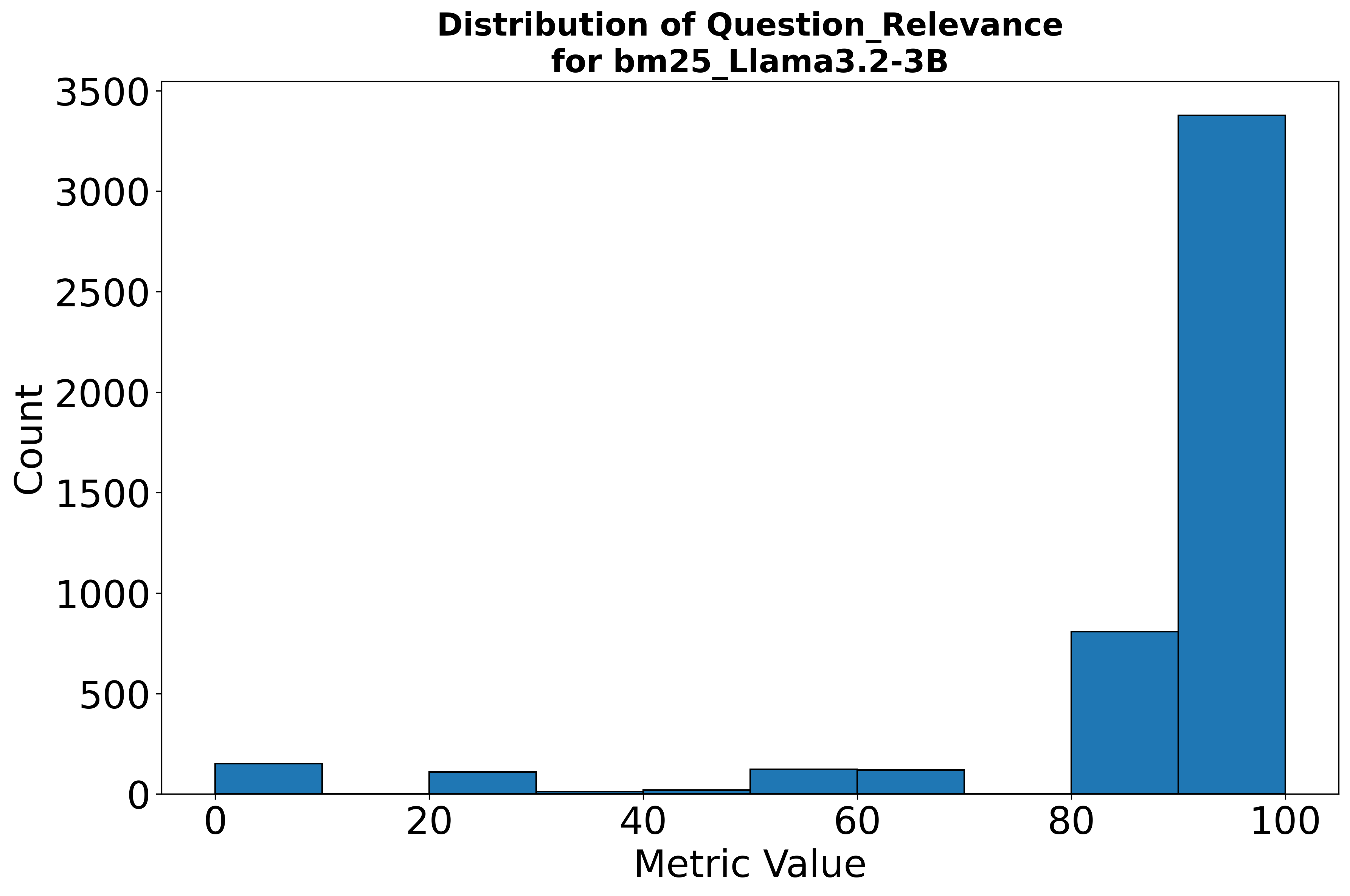}
    \caption{System E (QR)}
\end{subfigure} \hfill
\begin{subfigure}[b]{0.32\textwidth}
    \centering
    \includegraphics[width=\textwidth]{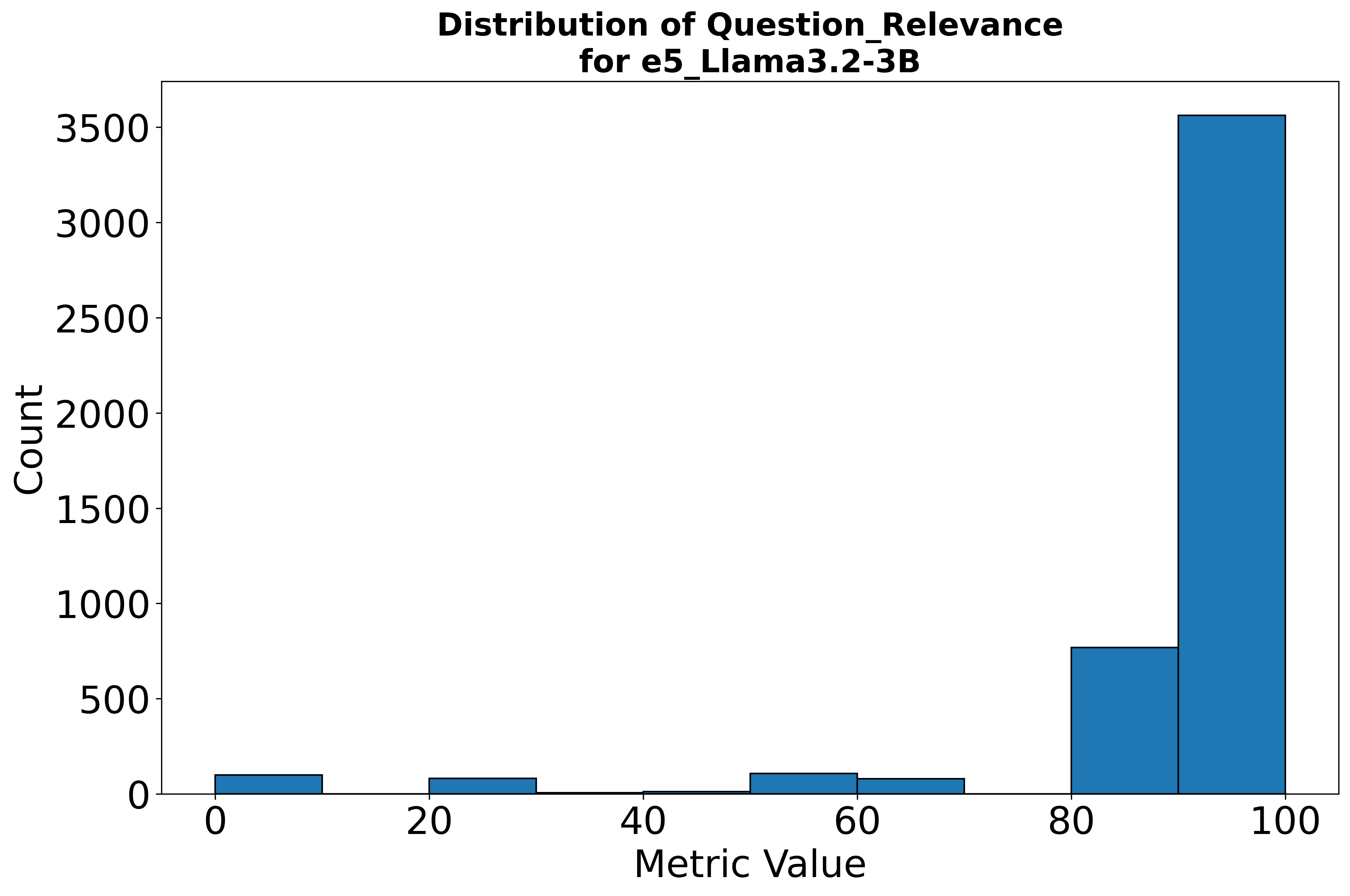}
    \caption{System F (QR)}
\end{subfigure}
\caption{Appendix: Distribution of Question Relevance (QR) scores for each RAG system.}
\label{fig:qr_dist}
\end{figure*}

\begin{figure*}[htbp!]
\centering
\small
\begin{subfigure}[b]{0.32\textwidth}
    \centering
    \includegraphics[width=\textwidth]{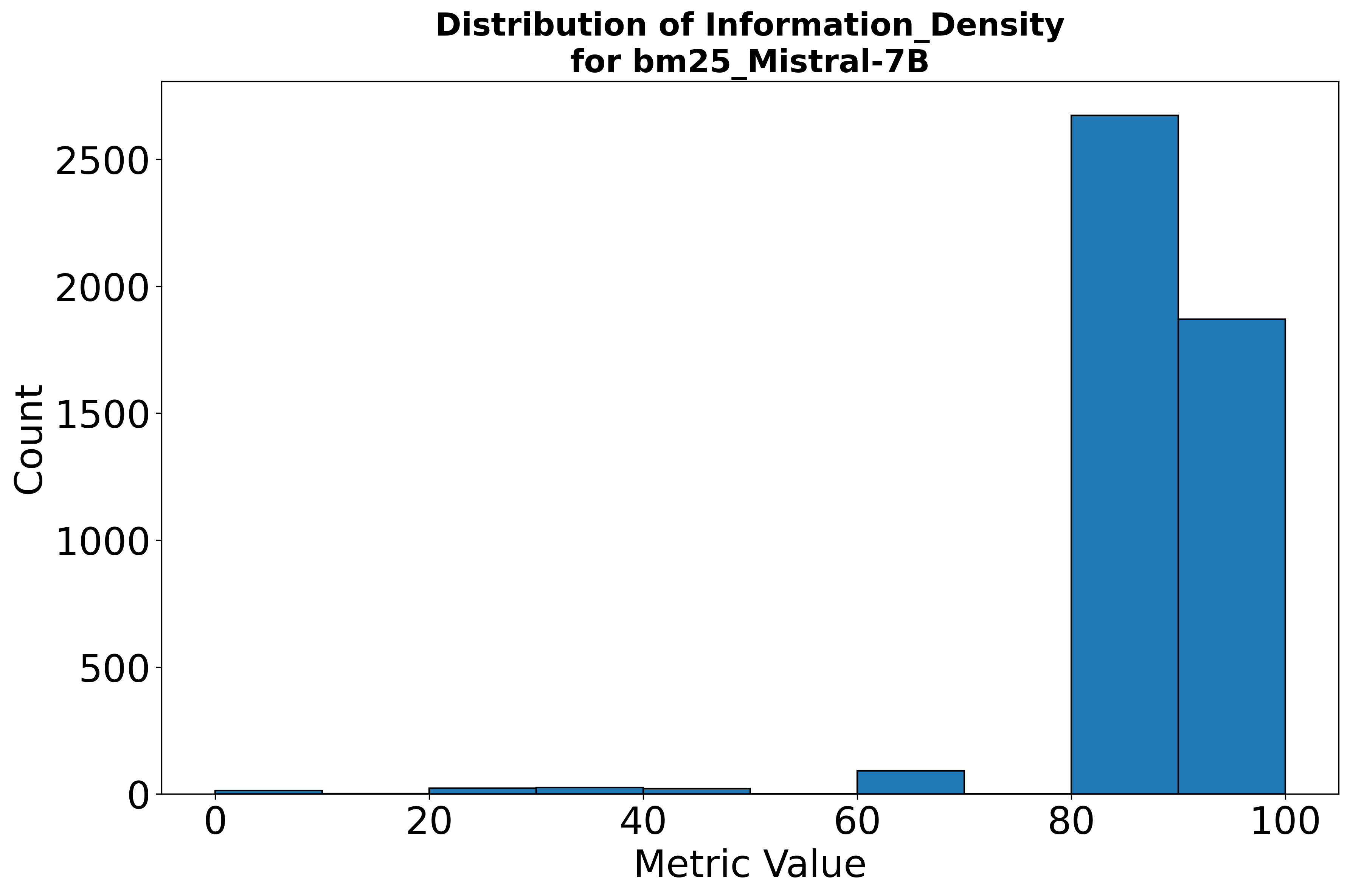}
    \caption{System A (ID)}
\end{subfigure} \hfill
\begin{subfigure}[b]{0.32\textwidth}
    \centering
    \includegraphics[width=\textwidth]{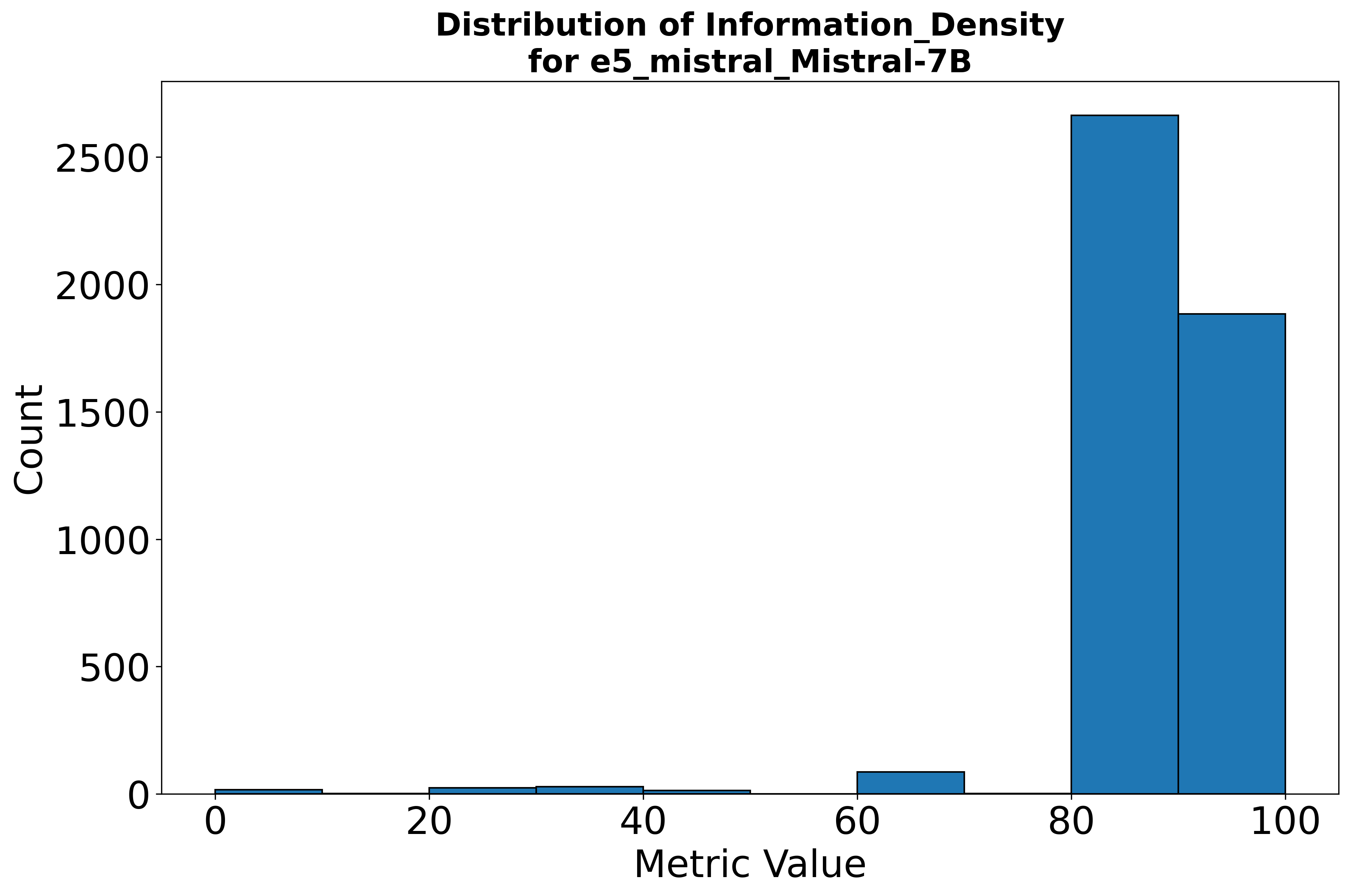}
    \caption{System B (ID)}
\end{subfigure} \hfill
\begin{subfigure}[b]{0.32\textwidth}
    \centering
    \includegraphics[width=\textwidth]{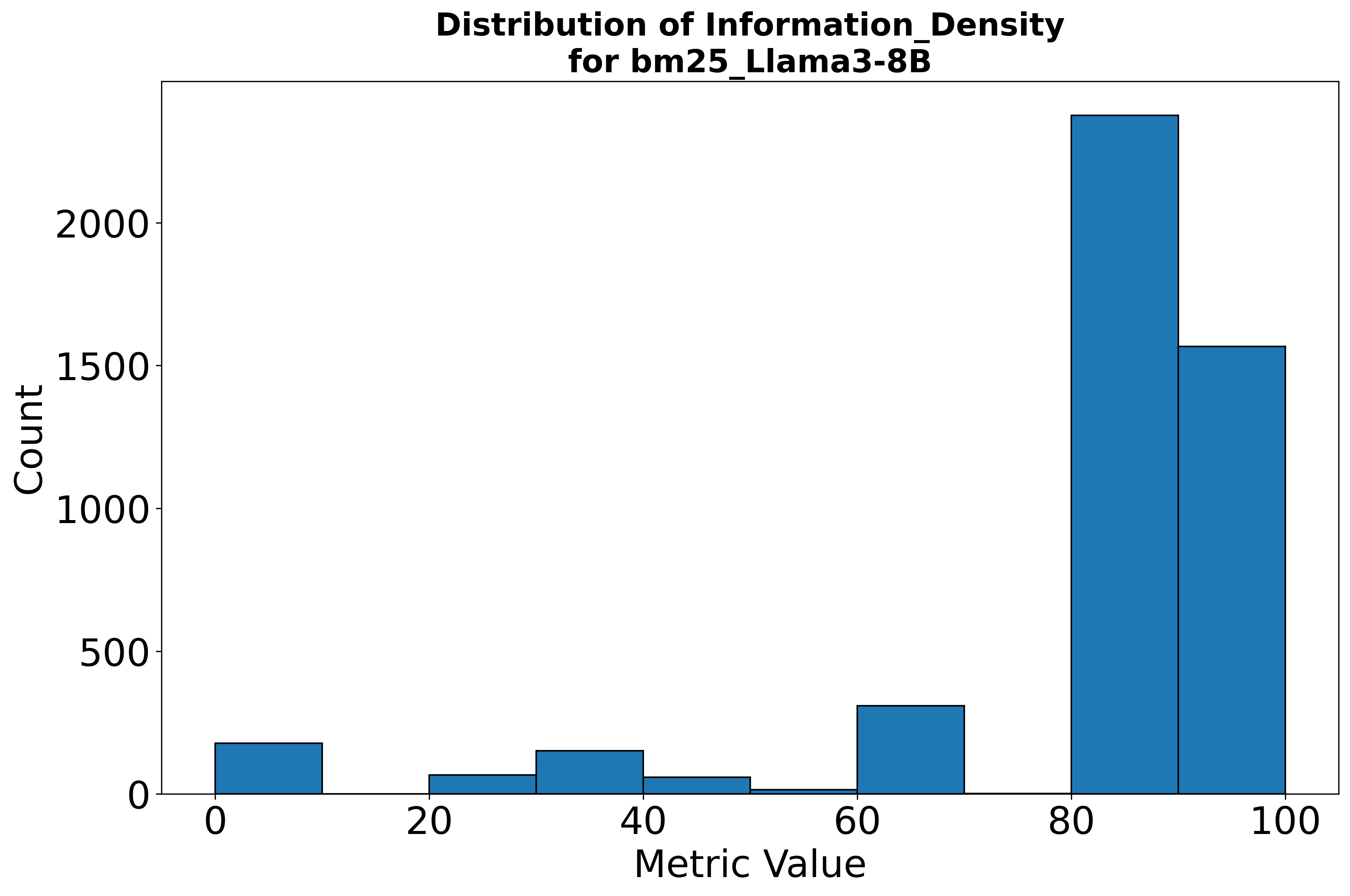}
    \caption{System C (ID)}
\end{subfigure} \\ \vspace{0.5em}
\begin{subfigure}[b]{0.32\textwidth}
    \centering
    \includegraphics[width=\textwidth]{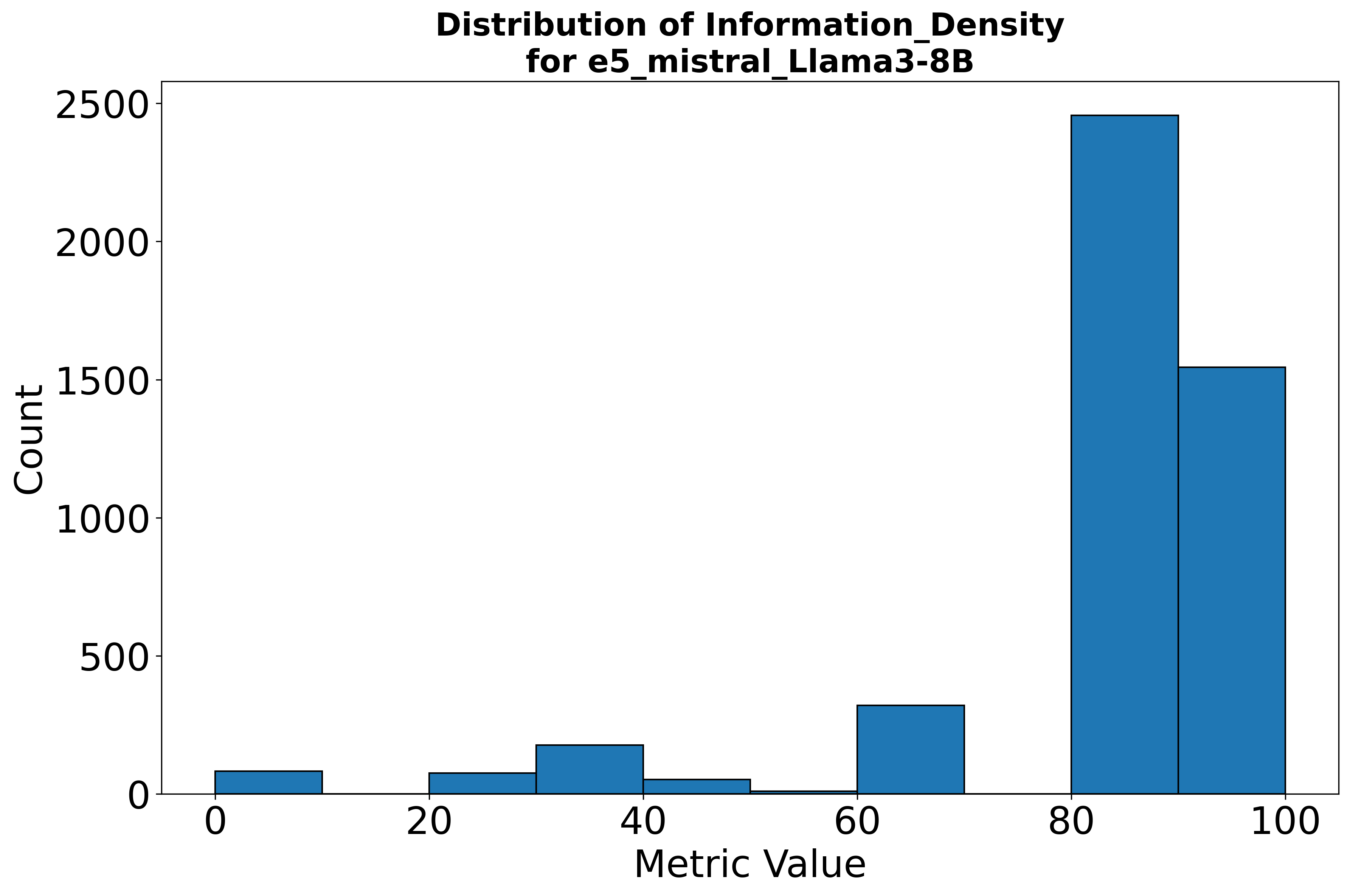}
    \caption{System D (ID)}
\end{subfigure} \hfill
\begin{subfigure}[b]{0.32\textwidth}
    \centering
    \includegraphics[width=\textwidth]{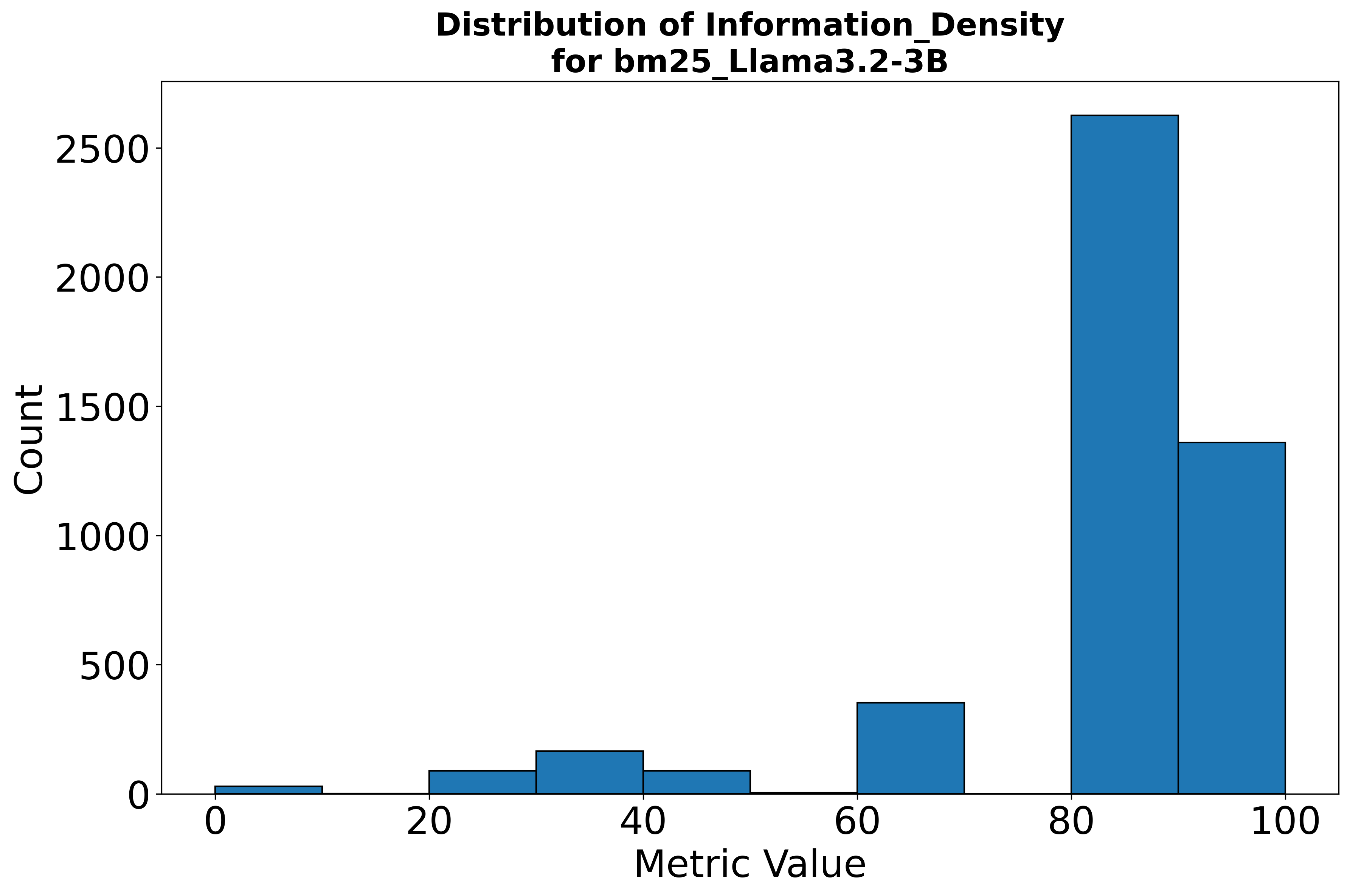}
    \caption{System E (ID)}
\end{subfigure} \hfill
\begin{subfigure}[b]{0.32\textwidth}
    \centering
    \includegraphics[width=\textwidth]{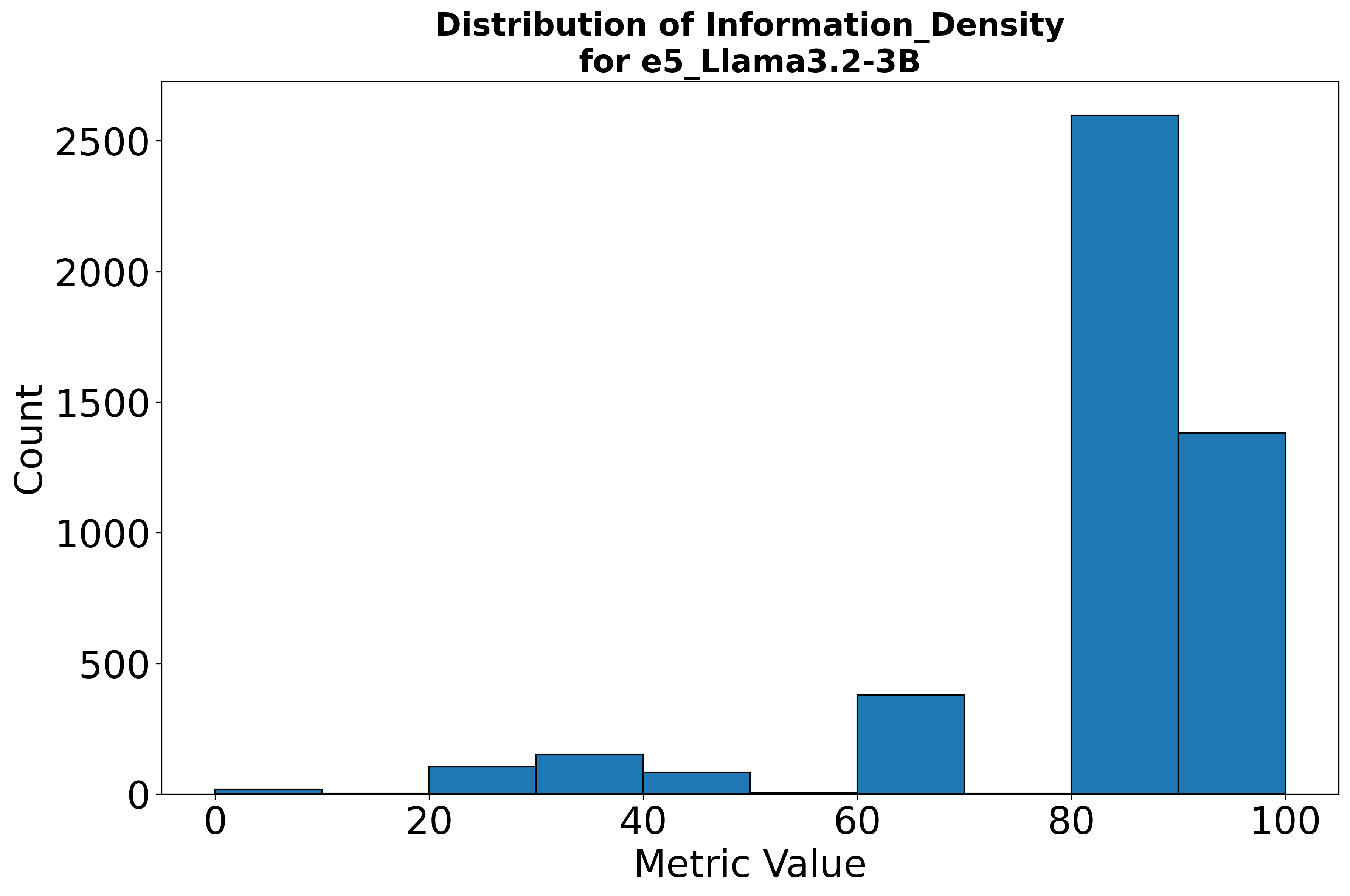}
    \caption{System F (ID)}
\end{subfigure}
\caption{Appendix: Distribution of Information Density (ID) scores for each RAG system.}
\label{fig:id_dist}
\end{figure*}

\begin{figure*}[htbp!]
\centering
\small
\begin{subfigure}[b]{0.32\textwidth}
    \centering
    \includegraphics[width=\textwidth]{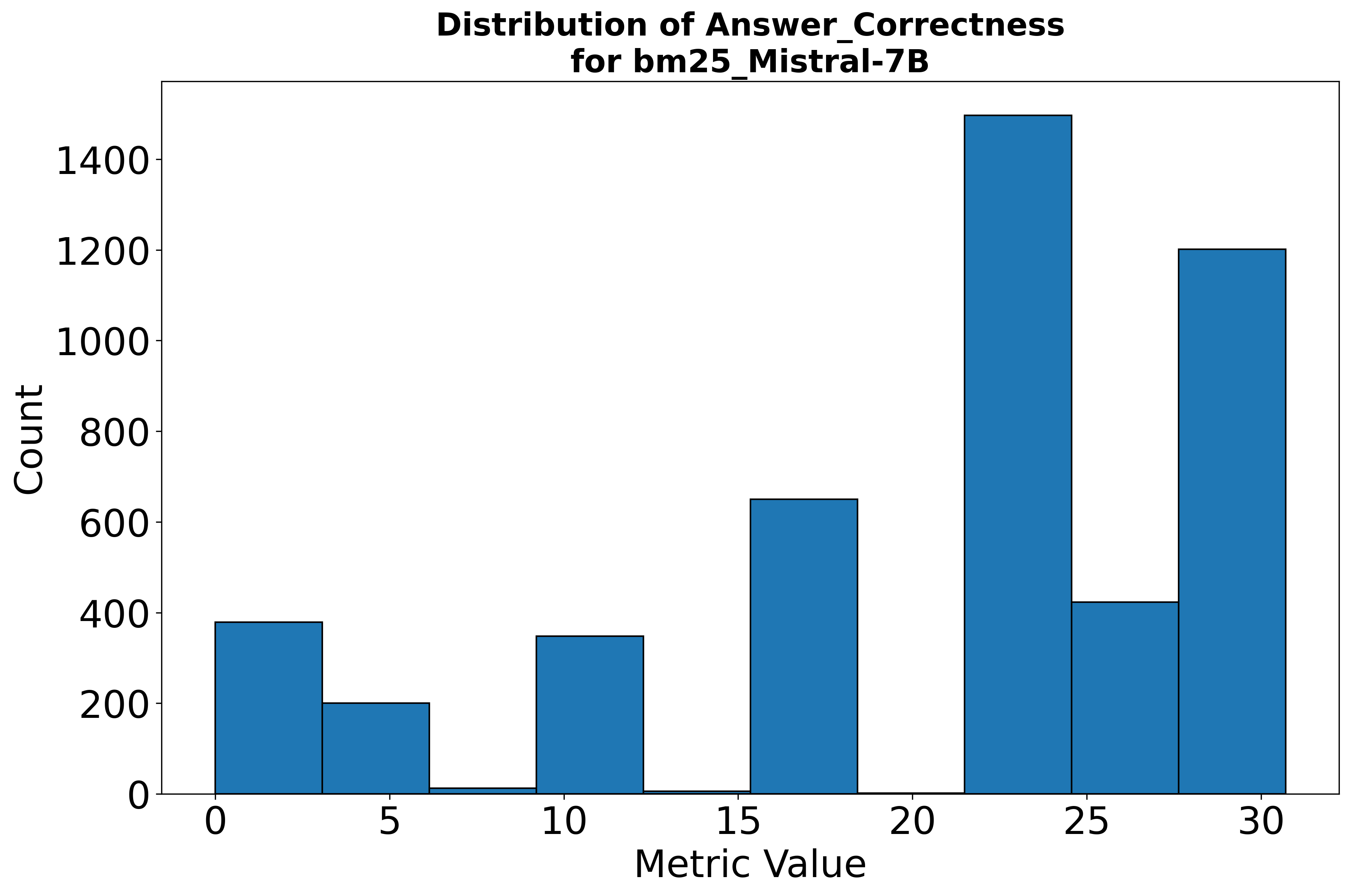}
    \caption{System A (AC)}
\end{subfigure} \hfill
\begin{subfigure}[b]{0.32\textwidth}
    \centering
    \includegraphics[width=\textwidth]{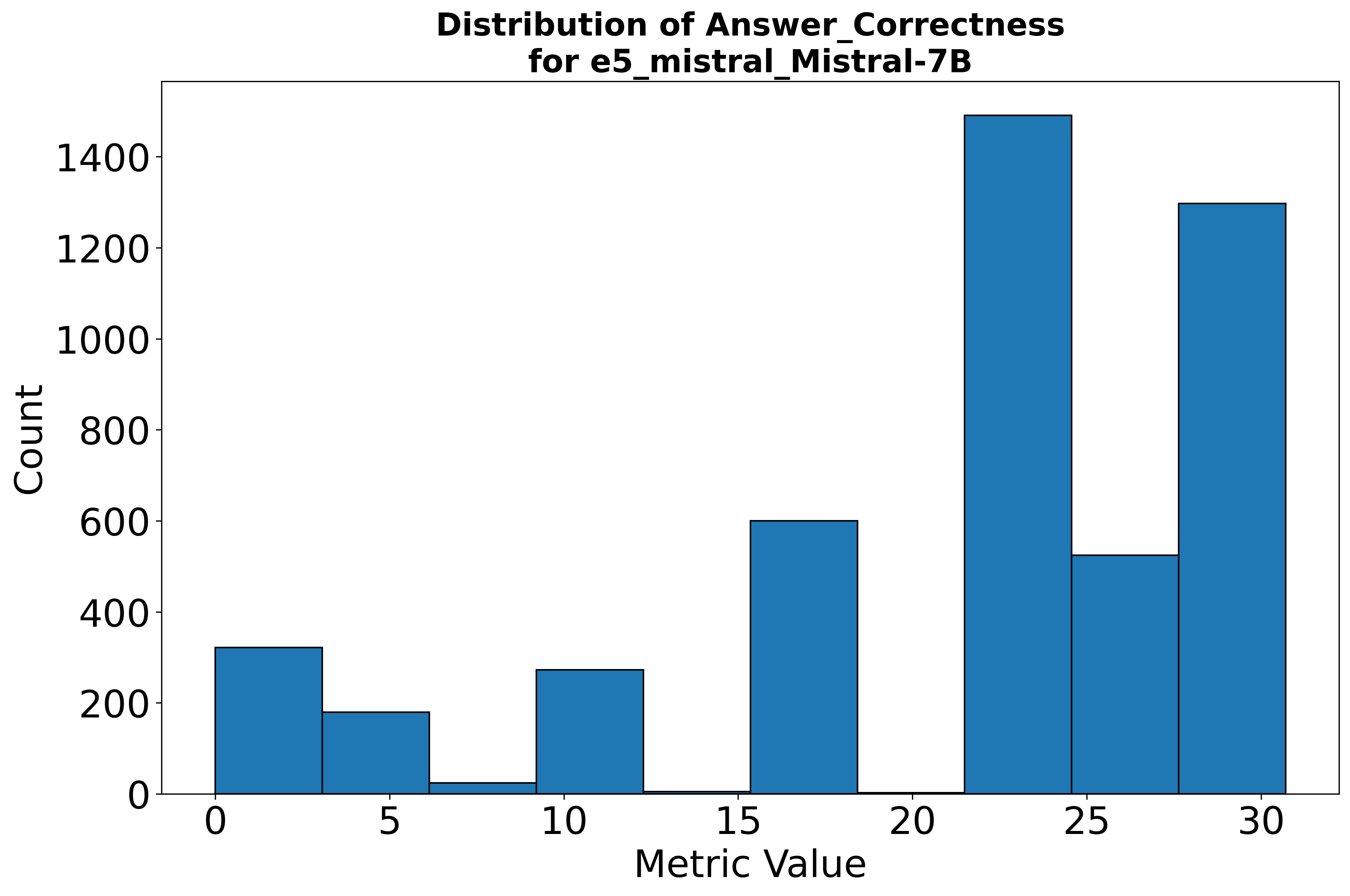}
    \caption{System B (AC)}
\end{subfigure} \hfill
\begin{subfigure}[b]{0.32\textwidth}
    \centering
    \includegraphics[width=\textwidth]{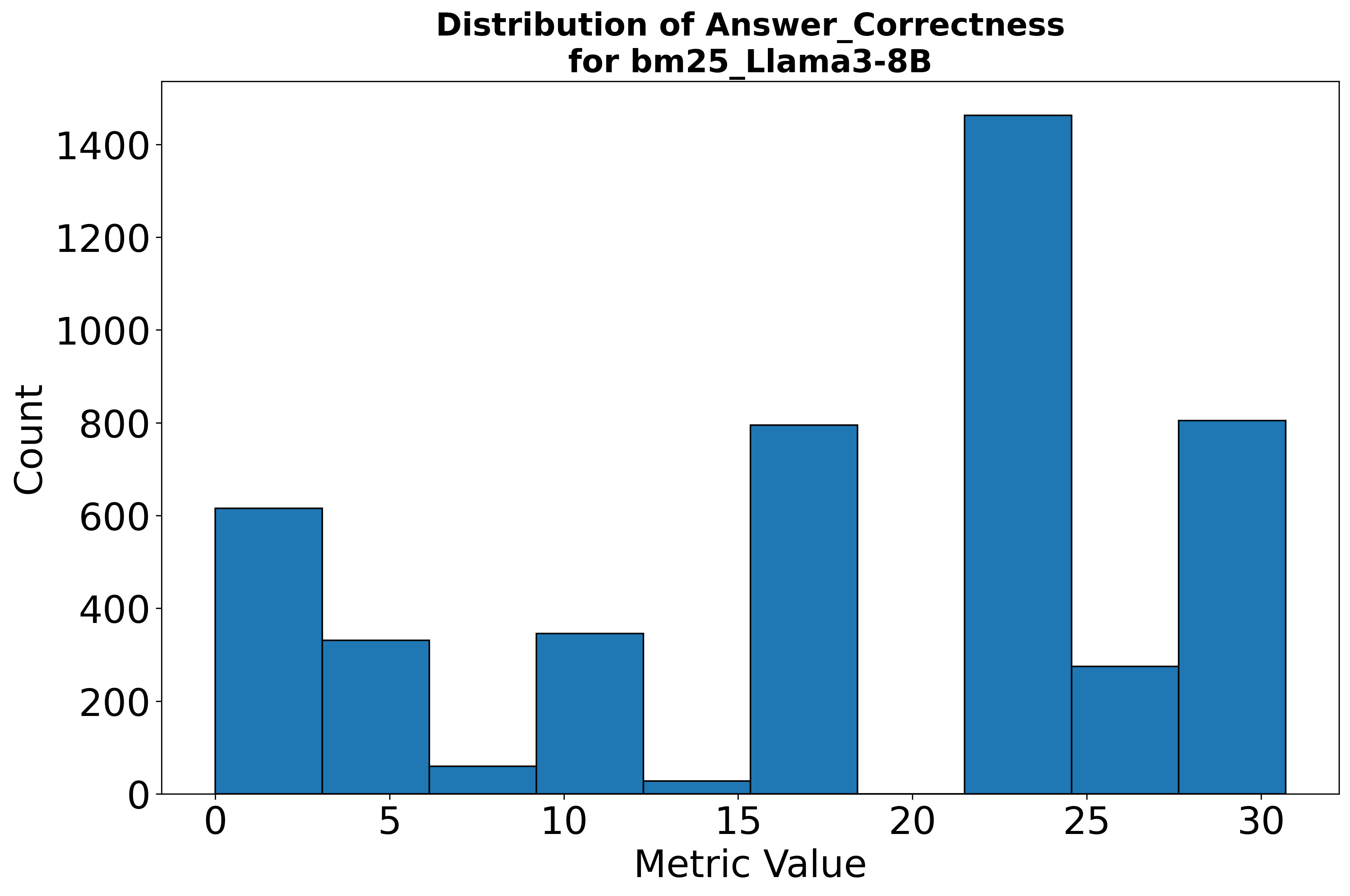}
    \caption{System C (AC)}
\end{subfigure} \\ \vspace{0.5em}
\begin{subfigure}[b]{0.32\textwidth}
    \centering
    \includegraphics[width=\textwidth]{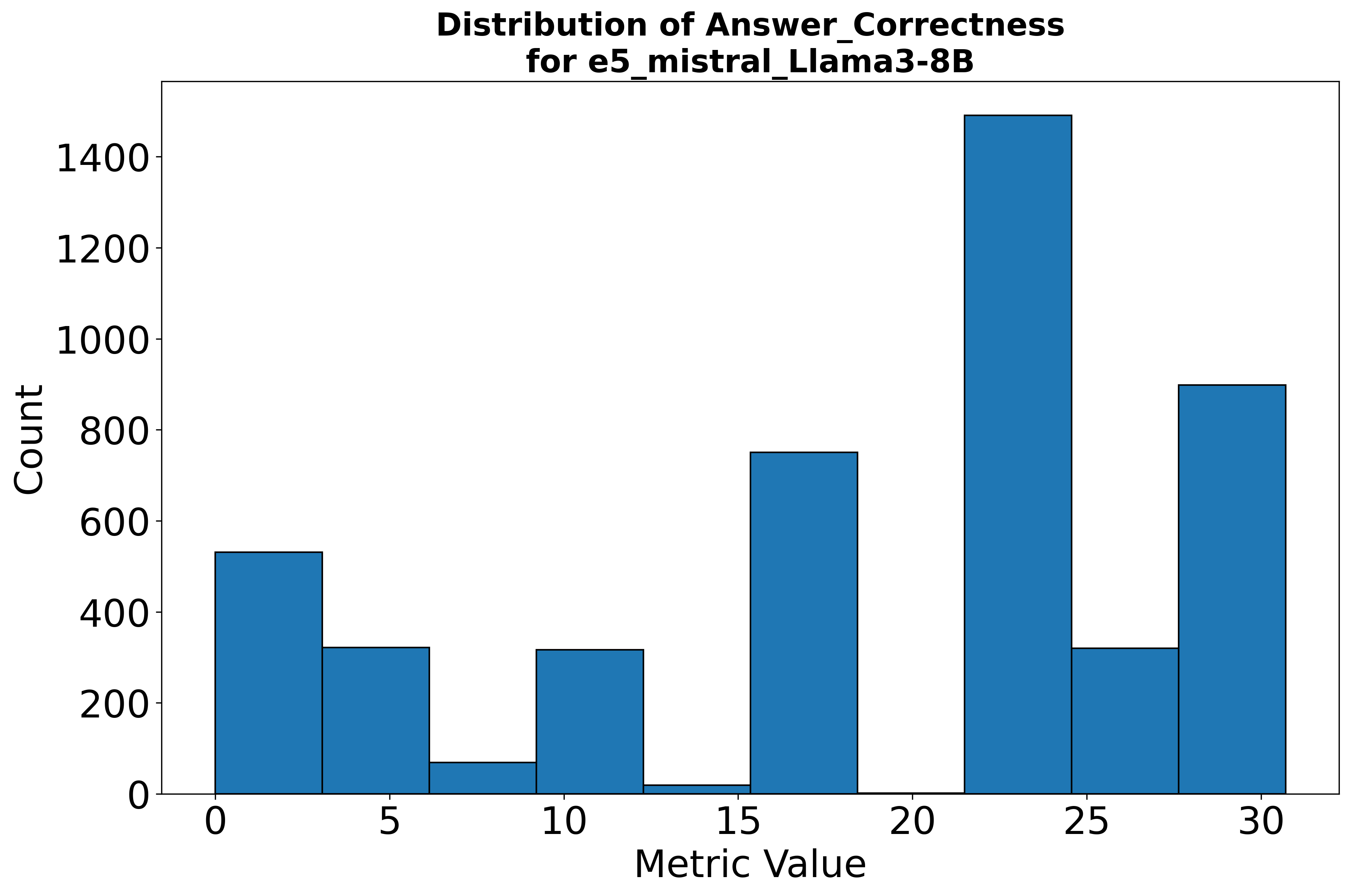}
    \caption{System D (AC)}
\end{subfigure} \hfill
\begin{subfigure}[b]{0.32\textwidth}
    \centering
    \includegraphics[width=\textwidth]{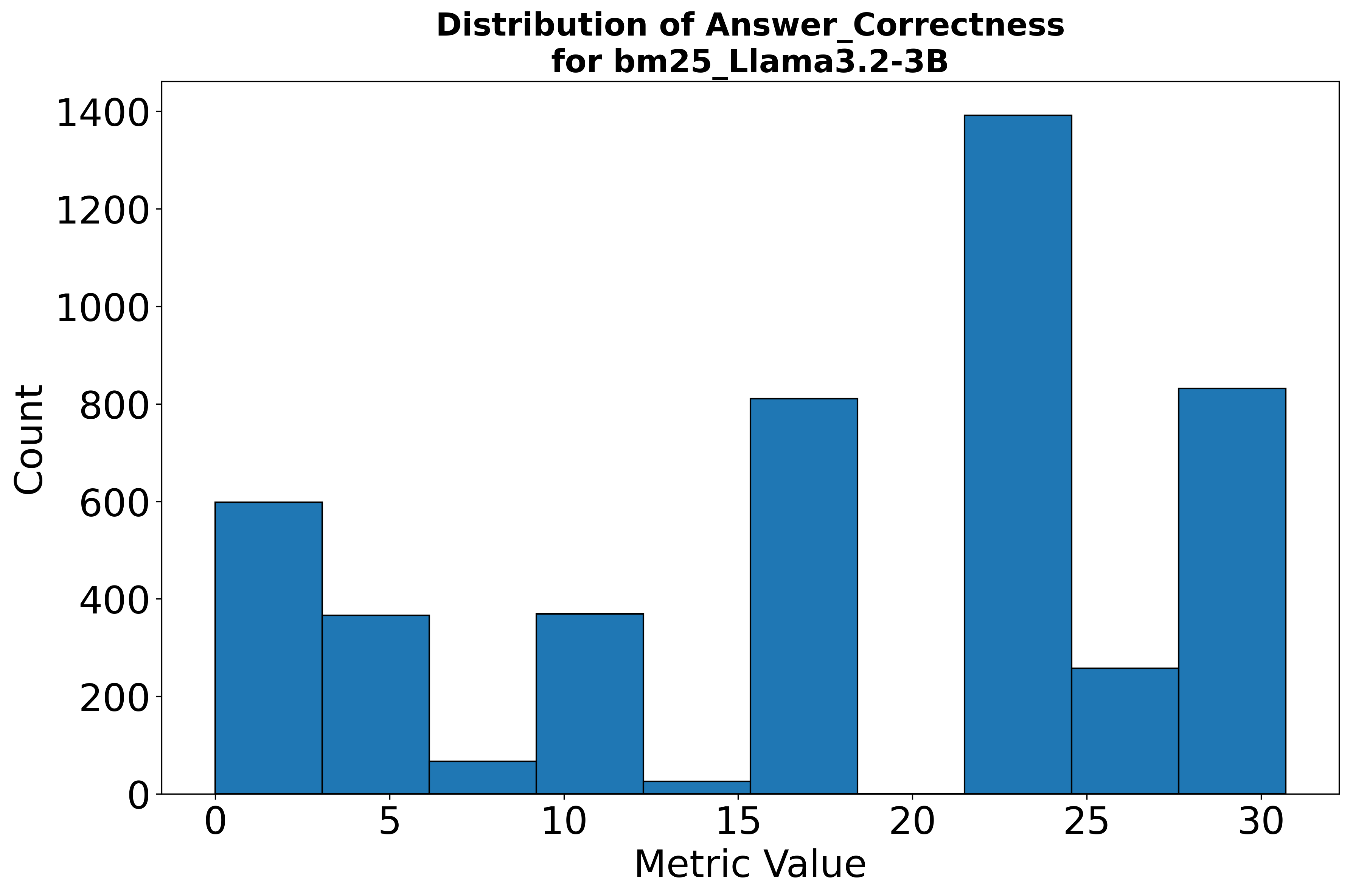}
    \caption{System E (AC)}
\end{subfigure} \hfill
\begin{subfigure}[b]{0.32\textwidth}
    \centering
    \includegraphics[width=\textwidth]{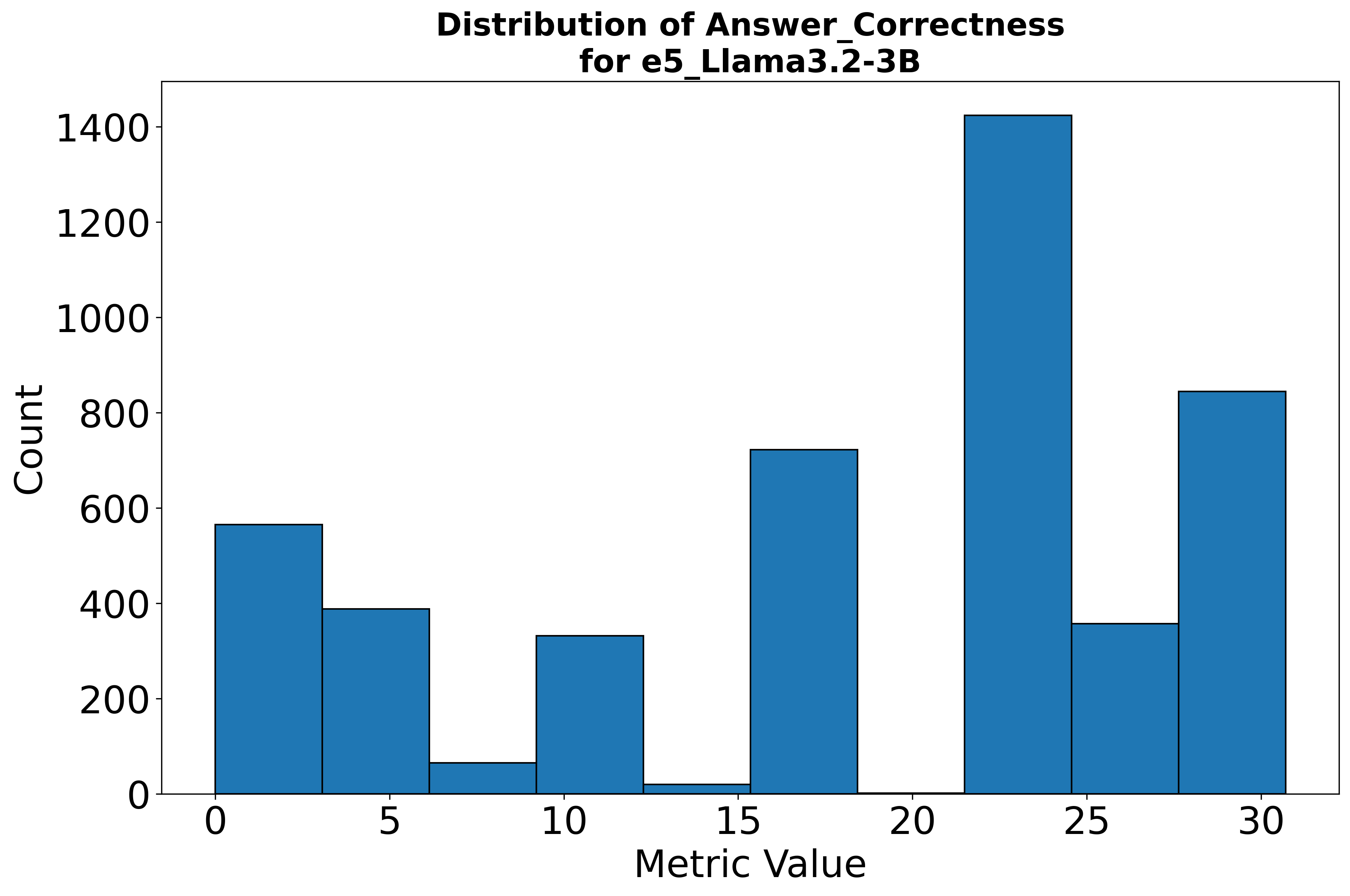}
    \caption{System F (AC)}
\end{subfigure}
\caption{Appendix: Distribution of Answer Correctness (AC) scores for each RAG system.}
\label{fig:ac_dist}
\end{figure*}

\begin{figure*}[htbp!]
\centering
\small
\begin{subfigure}[b]{0.32\textwidth}
    \centering
    \includegraphics[width=\textwidth]{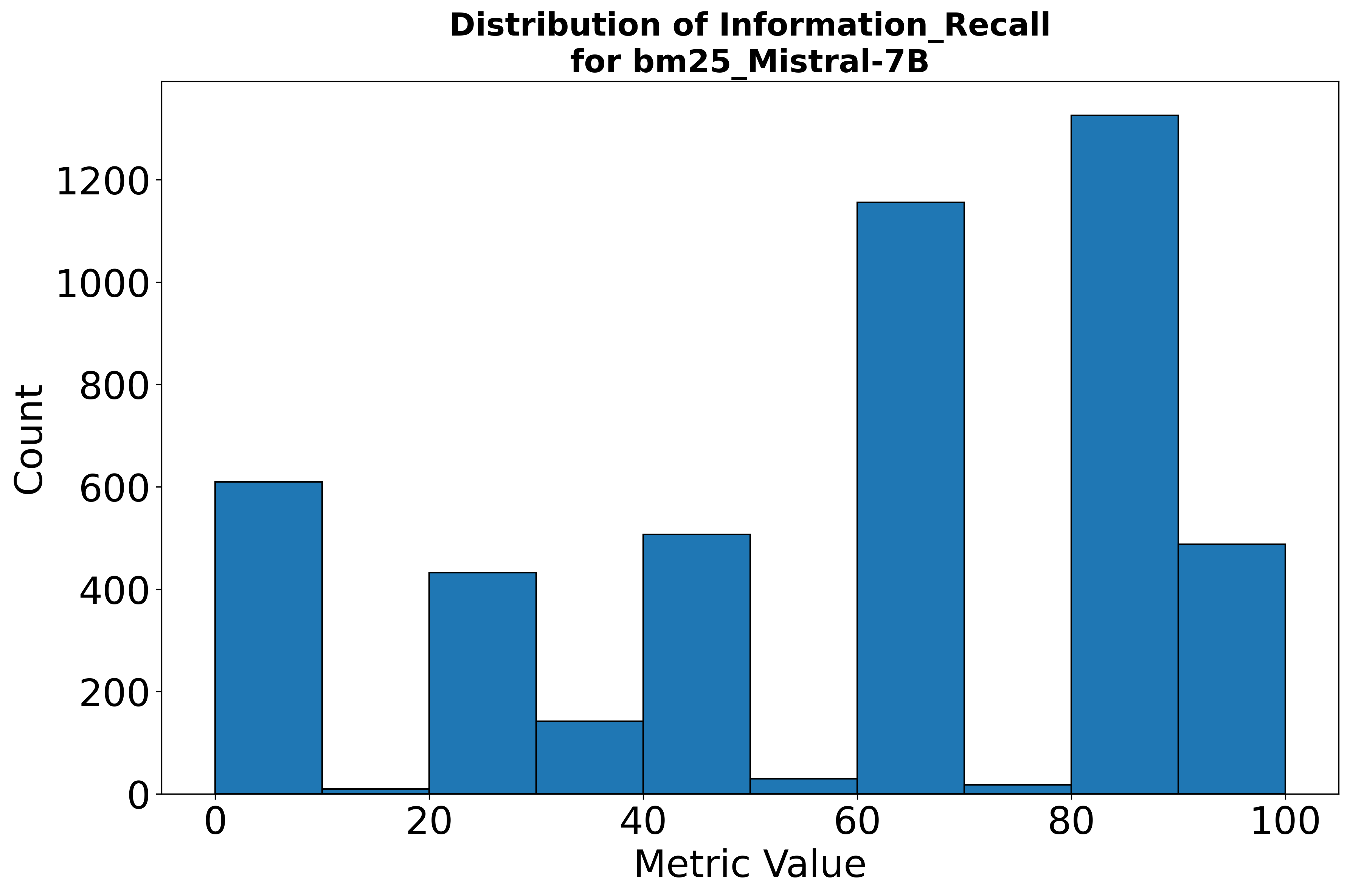}
    \caption{System A (IR)}
\end{subfigure} \hfill
\begin{subfigure}[b]{0.32\textwidth}
    \centering
    \includegraphics[width=\textwidth]{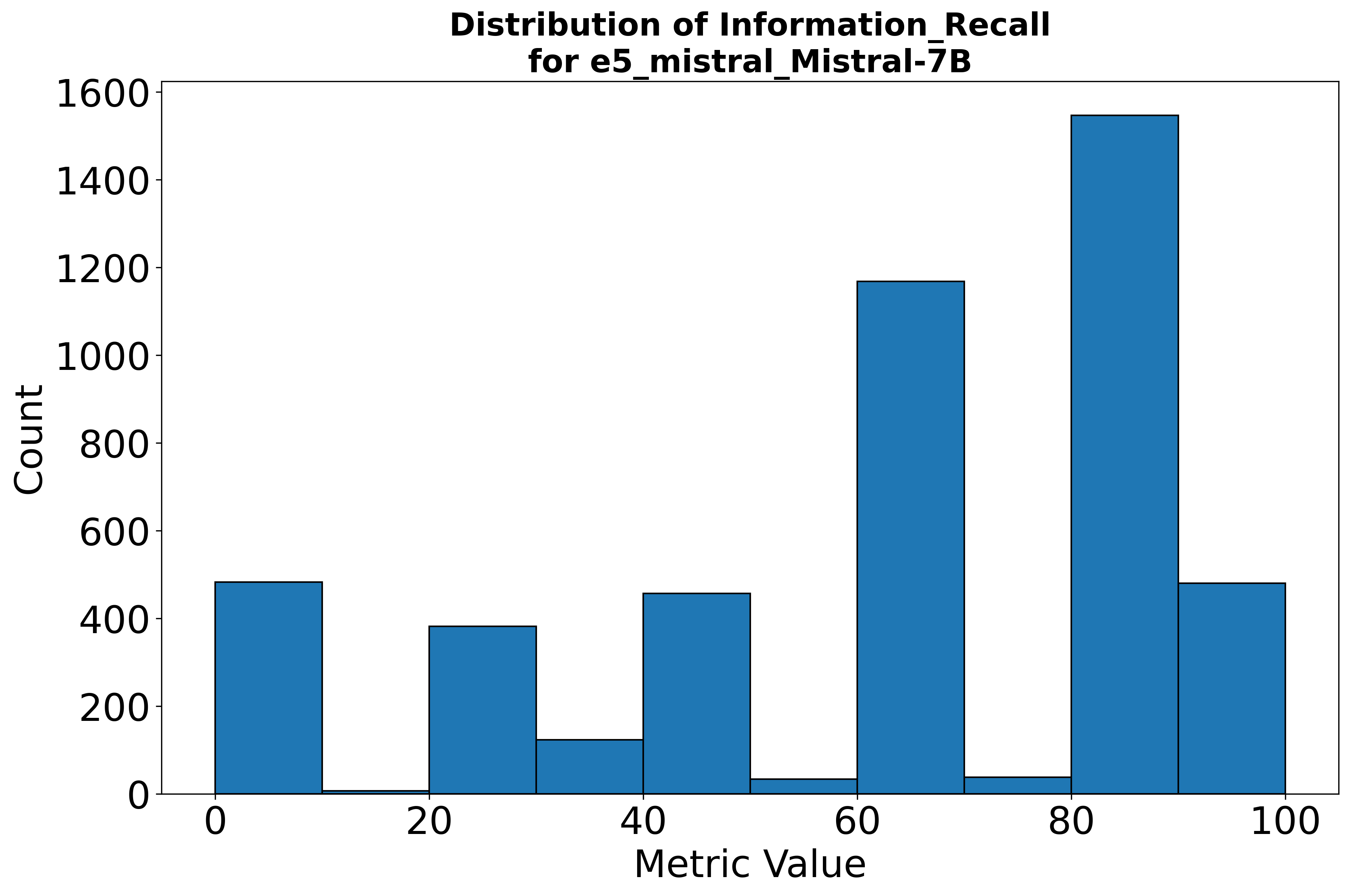}
    \caption{System B (IR)}
\end{subfigure} \hfill
\begin{subfigure}[b]{0.32\textwidth}
    \centering
    \includegraphics[width=\textwidth]{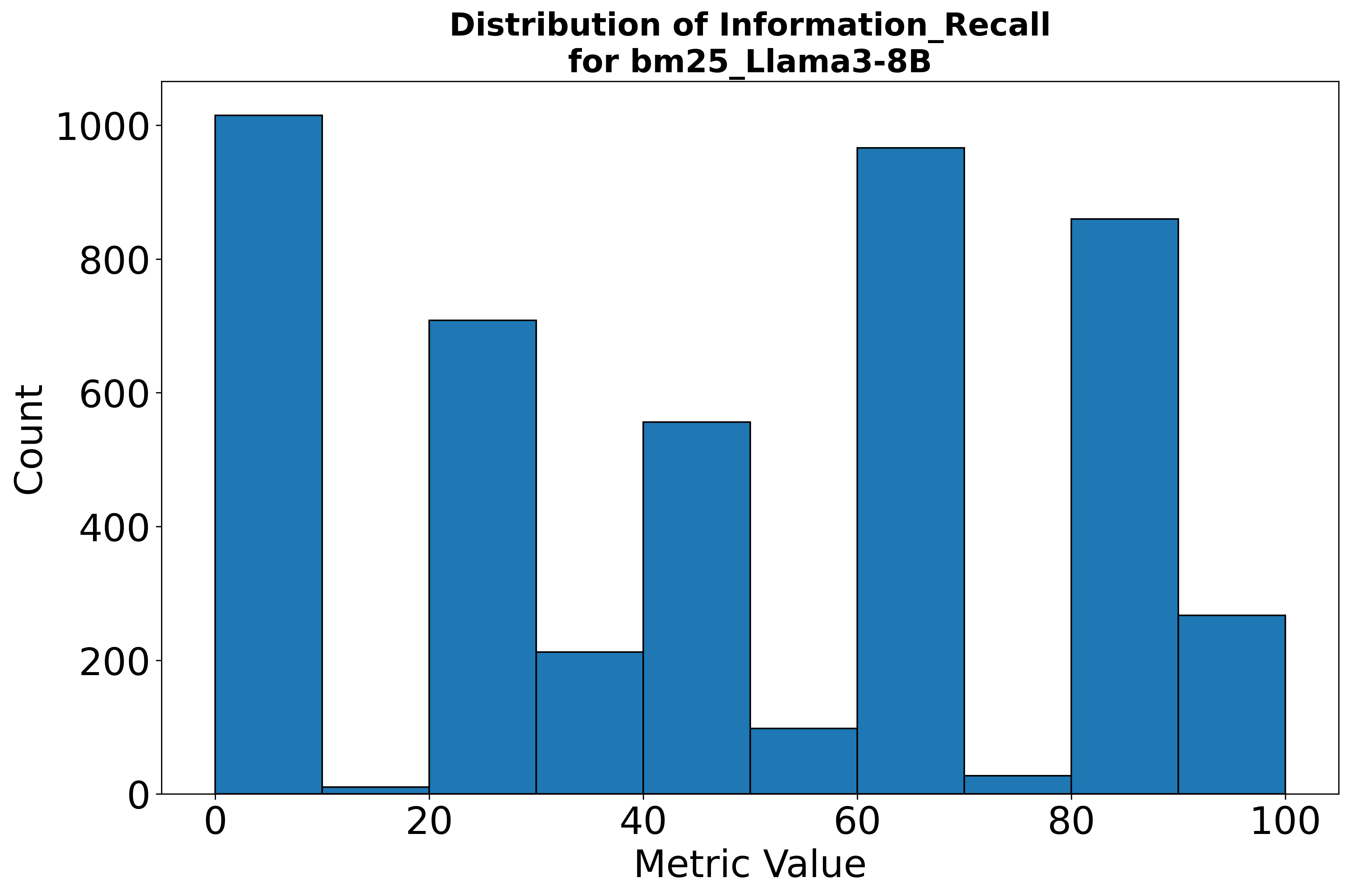}
    \caption{System C (IR)}
\end{subfigure} \\ \vspace{0.5em}
\begin{subfigure}[b]{0.32\textwidth}
    \centering
    \includegraphics[width=\textwidth]{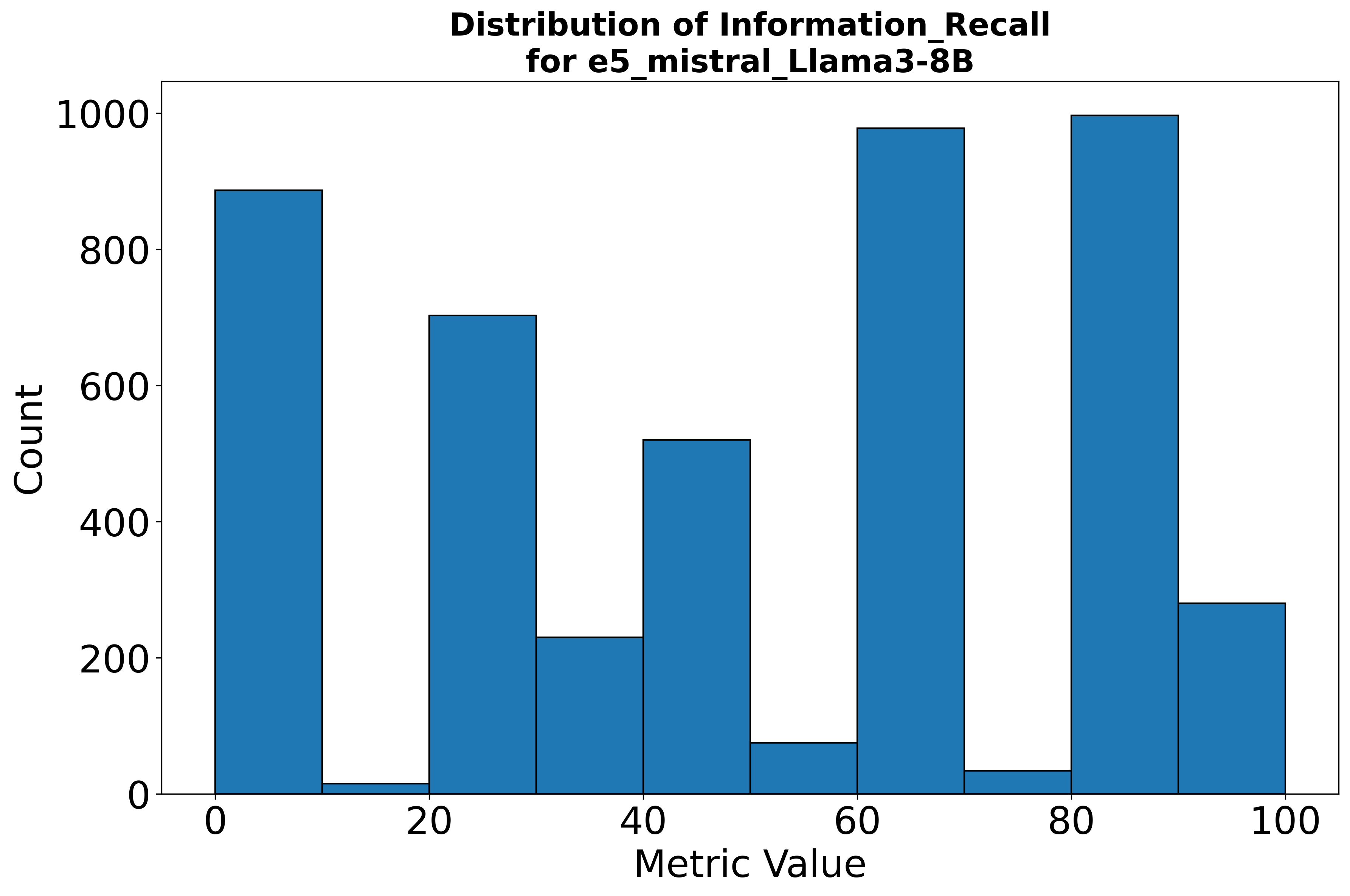}
    \caption{System D (IR)}
\end{subfigure} \hfill
\begin{subfigure}[b]{0.32\textwidth}
    \centering
    \includegraphics[width=\textwidth]{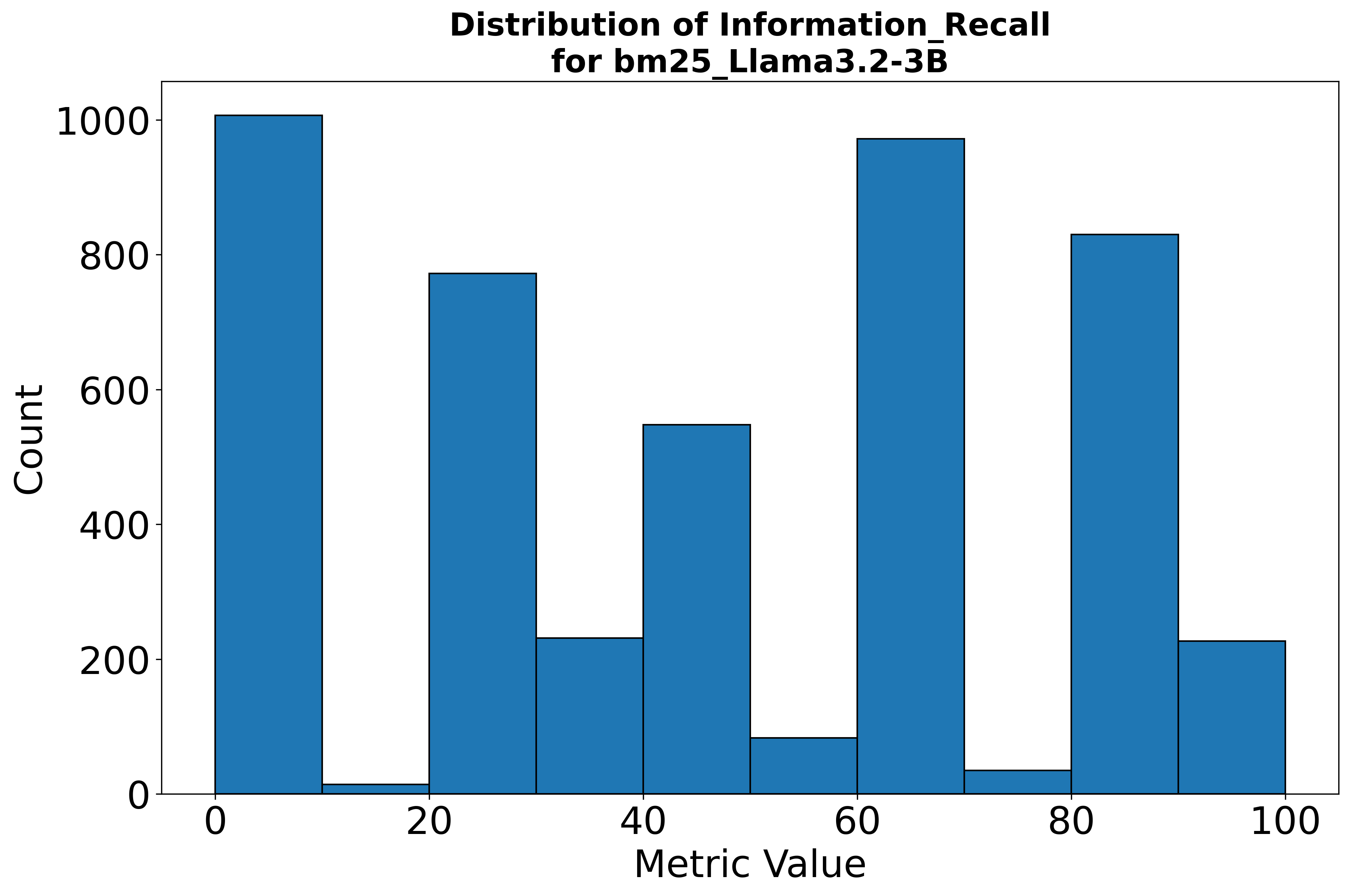}
    \caption{System E (IR)}
\end{subfigure} \hfill
\begin{subfigure}[b]{0.32\textwidth}
    \centering
    \includegraphics[width=\textwidth]{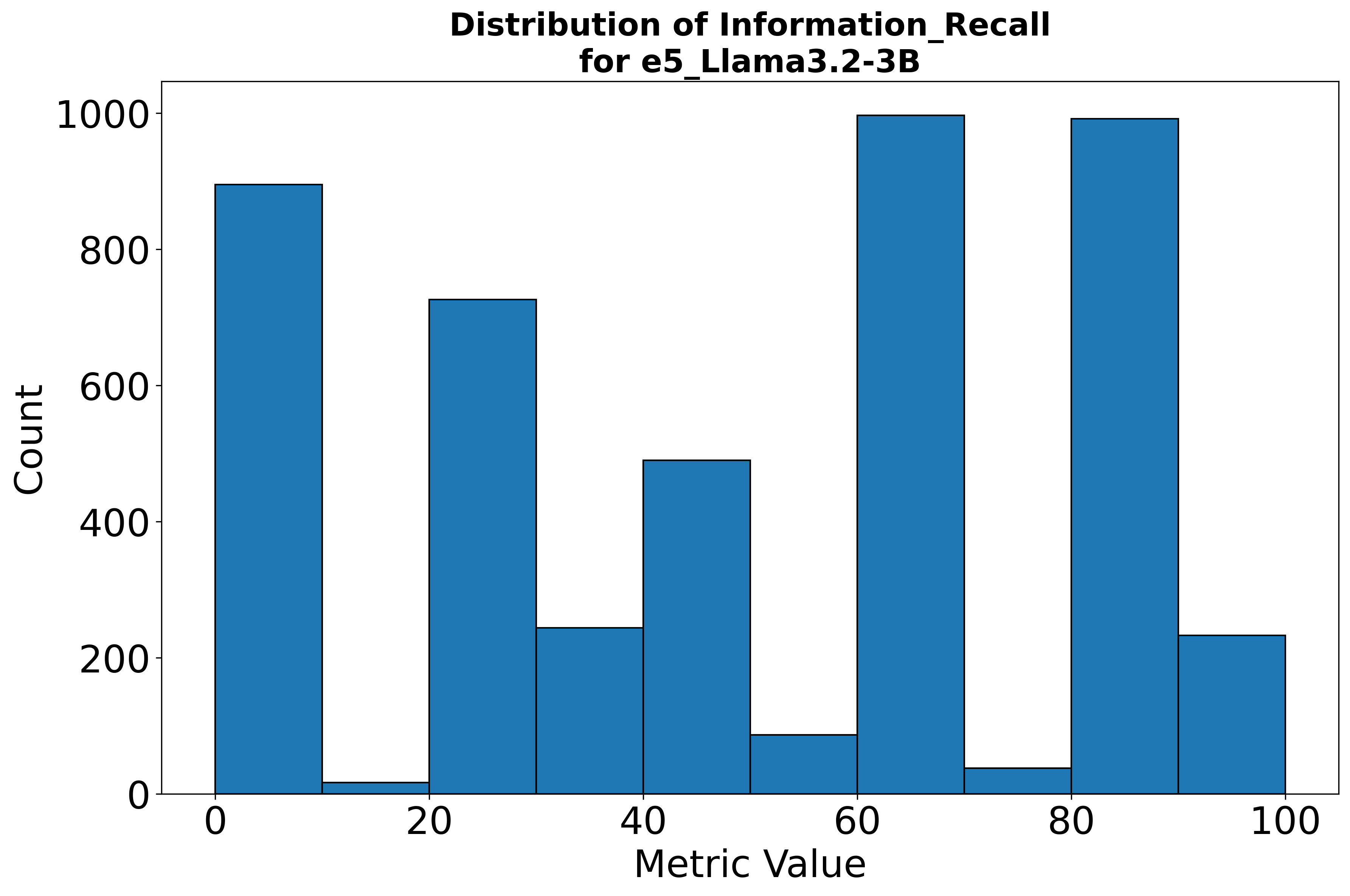}
    \caption{System F (IR)}
\end{subfigure}
\caption{Appendix: Distribution of Information Recall (IR) scores for each RAG system.}
\label{fig:ir_dist}
\end{figure*}

\end{document}